\documentclass{article}

\usepackage[preprint]{neurips_2022}

\usepackage[utf8]{inputenc} % allow utf-8 input
\usepackage[T1]{fontenc}    % use 8-bit T1 fonts
\usepackage{xr-hyper} % x-ref external file -> must load hyperref after xr-hyper
\usepackage{url}            % simple URL typesetting
\usepackage{booktabs}       % professional-quality tables
\usepackage{amsfonts}       % blackboard math symbols
\usepackage{nicefrac}       % compact symbols for 1/2, etc.
\usepackage{microtype}      % microtypography
\usepackage{xcolor}         % colors

\usepackage{bm}

\usepackage[disable]{todonotes}
\usepackage{amsmath, mathtools} %for aligning to = for maths
\usepackage[]{bm}
\usepackage{upgreek} % for the non-italic sigma 
\usepackage{xcolor} % for colouring equations 
\usepackage{pgfplots} %for drawing (3d) plots using latex
    \pgfplotsset{compat=1.16, % stop backward compatability warning
    colormap={whitered}{color(0cm)=(white); color(1cm)=(orange!75!red)}
    }
\usepackage{amssymb}% http://ctan.org/pkg/amssymb
\usepackage{pifont}% http://ctan.org/pkg/pifont
\newcommand{\cmark}{\ding{51}}%
\newcommand{\xmark}{\ding{55}}%
\usepackage{multirow}   % merged rows in table 
\usepackage{subcaption} % stacked subfigures
\usepackage{xurl}
\usepackage{wrapfig}

\usetikzlibrary{shapes.misc}
\tikzset{cross/.style={cross out, draw=black, minimum size=5*(#1-\pgflinewidth), inner sep=0pt, outer sep=0pt},
cross/.default={1pt}}

\usepackage{float}
\floatstyle{boxed}
\newfloat{derivation}{thp}{lop}
\floatname{derivation}{Derivation}
\newsavebox\MBox
\newcommand\Cline[2][red]{{\sbox\MBox{$#2$}%
  \rlap{\usebox\MBox}\color{#1}\rule[-1.2\dp\MBox]{\wd\MBox}{0.5pt}}}
\usepackage{microtype}
\usepackage{graphicx}
\usepackage{hyperref}
\title{Improving the Robustness of Neural Multiplication Units with Reversible Stochasticity}

\author{%
    Bhumika Mistry \& Katayoun Farrahi \& Jonathon Hare \\
    Department of Vision Learning, and Control\\
    Electronics and Computer Science\\
    University of Southampton \\
    \texttt{$\{$bm4g15, k.farrahi, jsh2$\}$@soton.ac.uk} \\
}

\begin{document}

\maketitle

\begin{abstract}
Multilayer Perceptrons struggle to learn certain simple arithmetic tasks. 
Specialist neural modules for arithmetic can outperform classical architectures with gains in extrapolation, interpretability and convergence speeds, but are highly sensitive to the training range. 
In this paper, we show that Neural Multiplication Units (NMUs) are unable to reliably learn tasks as simple as multiplying two inputs when given different training ranges. 
Causes of failure are linked to inductive and input biases which encourage convergence to solutions in undesirable optima. 
A solution, the stochastic NMU (sNMU), is proposed to apply reversible stochasticity, encouraging avoidance of such optima whilst converging to the true solution. 
Empirically, we show that stochasticity provides improved robustness with the potential to improve learned representations of upstream networks for numerical and image tasks. 
\end{abstract}

\section{Introduction}
Machine Learning trains models to generalise to data under the assumption of the data being independent and identically distributed (i.i.d) rather than Out-of-Distribution (OOD)~\citep{Wang-DomainGenSurvey-ijcai2021}. 
Generalisation to OOD data is considered to be extrapolation. 
Extrapolation is desirable as real-world training data can be limited to a small subset of the domain due to time and cost constraints. 
For example, a network to predict the kinematics for a robotic arm may be trained on one configuration, but be tested on an updated configuration outside the training one~\citep{martius2017extrapolation}. 
However, networks still struggle to extrapolate on even simple mathematical tasks~\citep{saxton2018analysing}. 
One way to improve model compatibility with extrapolative data is via inductive biases~\citep{mitchell1980biases}, which help guide the optimisation of parameters using prior assumptions designed into the architecture. 
Structural biases which result in specialists can also help avoid memorisation, a prevalent issue in neural networks~\citep{Zhang2020Identity}. 

Neural Arithmetic Logic Modules (NALMs), are specialist differentiable neural modules which learn systematic generalisations of arithmetic/logic operations \citep{mistry2021primer, trask2018neural, madsen2020neural}. 
NALMs can extrapolate to OOD data while also having interpretable weights. 
The first NALM, introduced by \citet{trask2018neural}, focuses on learning arithmetic operations (i.e. $+$, $-$, $\times$, $\div$, square, and square-root). 
Their Neural Arithmetic Logic Unit (NALU) learns arithmetic operations and selection of relevant inputs in a differentiable manner where the weight values reflect the chosen operation and input. 
Real-world applications have included the NALU as a sub-component in larger end-to-end systems which learns more complex tasks such as analogy composition \citep{wu2020analogical} or scheduling of content-delivery-networks \citep{Zhang2019LivesmartAQ}. 
For the NALM to be a successful sub-component in such networks, three attributes are required. (1) Provide informative gradients for the upstream networks to learn. (2) Be a robust module which can learn regardless of the input as sub-components of such larger networks have limited (if any) control of the distribution/range of their inputs. (3) Learn implicit selection, as there is no guarantee that all input values (i.e. features) are relevant. 
Though recent works improve on some of NALU's shortcomings such as convergence, initialisation, and interpretability \citep{madsen2020neural, schlor2020inalu, heim2020neural}, none have been successful in achieving extrapolation over various input ranges. 

Using stochasticity to improve learning is a practice that has been used by the community for many years. 
Noise can be injected into the input or weights to improve generalisation by implicitly inducing additional regularisation to the cost~\citep{an1996effects}. 
Or instead, noise can added to the gradients to encourage exploration giving the model the opportunity to escape local minimas. 
Such noise is \textit{irreversible}, meaning that the convergence cannot reach zero, but if annealed throughout training then zero convergence can occur~\citep{neelakantan15}. 
In contrast to these methods, our input noise is \textit{fully reversible} allowing for better exploration during training, and the ability to get convergence to minimal loss. 
Stochasticity has also been used to improve model robustness by reducing sensitivity to different inputs. 
For example in adversarial training, the noise is added to the input which the network must \textit{learn} to denoise~\citep{Goodfellow2015}. 
Our approach does not require any learning during the denoising stage and is designed to be automatically reversible no matter the input. 
General purpose neural networks which do not use specialist modules, can also apply data augmentation to perturb the input to teach the model to be more robust~\citep{shorten2019survey}.  
However, as our module learns exact multiplication, augmenting the input will result in the output no longer being a function of multiplication of the original inputs. Therefore, we reverse the effect of input noise in our model at the output to retain the correct input to output relation. 
% This solution takes influence from data augmentation and the reparameterisation trick \citep{kingma2013autoencoding}. 
% % This technique can be interpreted as giving the module an input range which it can solve. 

In this paper, we show that by introducing reversible stochasticity to a multiplication NALM (Section~\ref{sec:sNMU}) it is possible to improve the robustness without compromising on performance even if the NALM is a component in a deep neural network. 
We focus on multiplication as the effect of the operation can be understood by humans, but the space to learn is difficult enough to pose challenges to the optimisation of neural models. 
Multiplication is chosen over other elementary operations as the function's scalar field is more complex to learn than addition and subtraction, and is well defined for the domain of interest unlike division which is undefined when dividing by zero.  
The contributions of this paper are as follows\footnote{Code (MIT license) is available at: \url{https://anonymous.4open.science/r/nalm-robust-nmu-7577}.}: 
\begin{itemize}
    \item In Section~\ref{sec:mlp-vs-nalms} we motivate using specialist arithmetic modules over MLPs, showing gains in extrapolation, convergence speed and interpretability. 
    \item In Section~\ref{sec:single-layer-task} we show how robustness against different training ranges of specialist modules greatly varies using the simplest task for modelling an operation (i.e., calculating the multiplication of two inputs). 
    We relate the robustness issue to local optima inducing biases and in Section~\ref{sec:sNMU} present a novel approach, the sNMU to fix the issue.
    \item We assess the sNMU on arithmetic regression (Section~\ref{sec:arithmetic-ds-task}) and image tasks (static MNIST and sequential MNIST in Section~\ref{sec:mnist-arithmetic}) showing favourable performance compared to networks with pre-solved multiplication. 
\end{itemize}

\section{Overview of Neural Arithmetic Logic Modules}
This section details two NALMs used in our experiments, the Neural Addition Unit (NAU) and the Neural Multiplication Unit (NMU). 
\citet{madsen2020neural} introduce two specialist modules, one for dealing with addition and subtraction (the NAU) and the other for multiplication (the NMU). 
The NAU and NMU have a separate weight matrix and can be stacked (i.e. the output of a NAU is the input for a NMU) to learn compositional arithmetic expressions (e.g. $(x_1 + x_2) \times (x_3 - x_4)$). 
An NAU output element $a_o$ is defined as 
\begin{align}
\textrm{NAU}: a_o &= \sum_{i=1}^{I} \left(W_{i,o}\cdot \mathrm{x}_{i} \right), \label{eq:nau}
\end{align}
where $I$ is the number of inputs. 
An NMU output element $m_o$ is defined as  
\begin{align}
\textrm{NMU}: m_o &= \prod_{i=1}^{I} \left(W_{i,o}\cdot \mathrm{x}_{i} + 1 - W_{i,o} \right) . \label{eq:nmu}
\end{align}

Weights are ideally discrete values, where the NAU is 0, 1, or -1, representing no selection, addition and subtraction, and the NMU is 0 or 1, representing no selection and multiplication. 
To enforce discretisation of weights, regularisation is applied for a given period of training (see Appendix~\ref{app:nau-nmu-reg}) and before applying a sub-unit the weight matrix is clamped to [-1,1] for the NAU or [0,1] for the NMU.

\section{Motivating Specialist Modules over MLPs}\label{sec:mlp-vs-nalms}
We first demonstrate the advantage of using a specialist module, the NMU, over an MLP for multiplication in terms of extrapolation, convergence speed and interpretability for the task of applying the operation to two inputs. 
A single hidden layer MLP network with a width of either 1 or 100 is learnt to check both extremes. 
A ReLU activation is used and we keep bias terms to avoid constraining expressiveness. 

\textbf{Setup:} Given two inputs $x_1$ and $x_2$, output the value for $x_1 \times x_2$. 
% This uses the same setup as Section~\ref{subsec:smt-setup}. 
Networks are trained on a range $\mathcal{U}$[1,2) and tested on an extrapolation range $\mathcal{U}$[2,6). 
Results show the average error and best iteration step with a 95\% confidence interval. 
For experiment details see Single Module Task in Appendix~\ref{app:exp-parameters}. 

\begin{figure}[b]
\vskip 0.1in
\centering
% NOTE - the trim and clip remove the key showing the bins
\includegraphics[trim={0cm 2.8cm 0cm 0cm}, clip, width=0.8\textwidth]{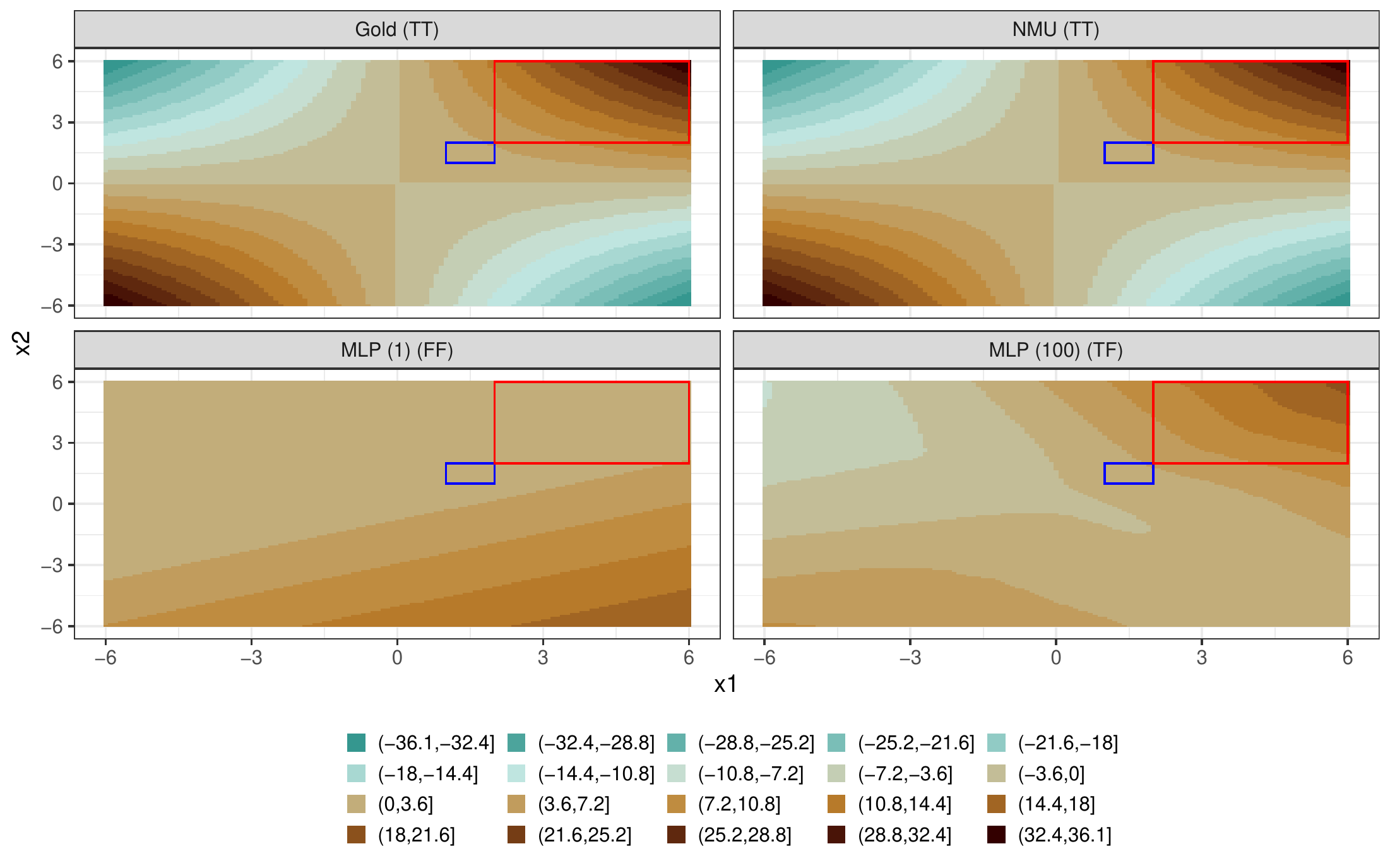}
\caption{20 bin surface plots for the multiplication operation, comparing the golden solution (top left) to a learnt NMU (top right) and MLPs of different widths (bottom row). The letters in the brackets are True (T)/False (F), representing if the minimum loss threshold (1e-5) for the interpolation and extrapolation range have been met respectively. The blue and red squares represent the interpolation (training) and extrapolation (test) ranges respectively.}
\label{fig:surface-mul}
\vskip -0.1in
\end{figure}

\textbf{Results:} Table~\ref{tab:mlp-vs-nalm} shows MLPs are unable to find any solutions for multiplication. 
Comparing a width of 1 and 100: 
(1) Increasing width reduces both interpolation and extrapolation mean square error (MSE) at the cost of increased training iterations and parameters, but extrapolative solutions are not learnt.  % careful with this one
(2) Extremely wide networks encourage memorisation over the interpolation range which is unable to hold over the extrapolation range. 
(3) Robustness over different initialisations improves with increased width. 
In contrast, the NMU, which requires learning only two parameters (which is three less parameters than the MLP of width one), has success on all initialisations, is robust, and requires the least iterations to be solved. 
Comparing the learnt 2D surface plots (Figure~\ref{fig:surface-mul}) of the MLPs and the NMU better visualises points (1) and (2) from above. 

\begin{table*}[t]
\centering
\caption{Comparison of learning multiplication using a MLP or NMU for multiplying two inputs. The extrapolation and interpolation error correspond to the iteration with the lowest validation (interpolation) error. For the MSE columns, values in brackets represent the percentage of successful seeds which meet the minimum extrapolation criteria of 1e-5. Solved iteration represents the iteration which successful models get solved at. 95\% confidence intervals are given.}
\label{tab:mlp-vs-nalm}
\vskip 0.1in
\scalebox{0.8}{
\begin{tabular}{lllllll}
\toprule
\textbf{Model} & \textbf{\#Params} & \textbf{Interpolation MSE; (\%)} & \textbf{Extrapolation MSE; (\%)} & \textbf{Solved Iteration} & \textbf{Iterations}\\\midrule
MLP (1)      & 5       & 2.3E-01 $\pm$ 7.7E-02  (0) & 1.9E+02 $\pm$2.9E01 (0)  &- & 5E+04\\
MLP (100)    & 401     & 3.4E-07 $\pm$ 4.7E-08  (100) & 3.7E+01 $\pm$ 9.5E-01 (0) &- & 2E+06\\
\textbf{NMU} & \textbf{2} & \textbf{4.6E-14 $\pm$ 0 (100)}    & \textbf{9.9E-13 $\pm$ 0 (100)} &\textbf{1.0E+04 $\pm$ 2.8E+02 } & 5E+04\\\bottomrule   
\end{tabular}
}
\vskip -0.1in
\end{table*}

\section{Numerical Experiment Setups and Evaluation Metrics}\label{sec:exp-setup-and-eval}
Before learning larger systems, we must first verify if NALMs can work as an independent unit. 
As NALMs expect unnormalised inputs, we use two synthetic arithmetic regression tasks to test robustness against different training ranges.  
This section describes the experimental setup for our `Single Module Task' and the `Arithmetic Dataset Task'~\citep{madsen2020neural}. 
We also detail the three evaluation metrics: \textit{success rate, speed of convergence} and \textit{sparsity error} along with the interpolation and extrapolation ranges. 
Summaries of parameters and hardware/runtimes are found in Appendices~\ref{app:exp-parameters} and \ref{app:hardware-and-timings}. 

\subsection{Single Module Task}\label{subsec:smt-setup}
\textbf{Motivation.} To determine the best module to further explore, we evaluate if the simplest possible multiplication task can be learnt over various training ranges. 
% Hyperparameters are not tuned as this experiment is for observing problems in existing models.

\textbf{Setup.} Given two inputs $x_1$ and $x_2$ (as floats), use a single module to calculate the value for $x_1 \times x_2$. 
We test a variety of existing multiplication NALMs including the: NALU~\citep{trask2018neural}, $\mathrm{NAC}_{\bullet}$~\citep{trask2018neural}, NMU~\citep{madsen2020neural}, iNALU~\citep{schlor2020inalu}, G-NALU~\citep{rana2019exploring}, NPU, and Real NPU~\citep{heim2020neural}. 
A detailed explanation of each module can be found in~\citet{mistry2021primer} and therefore is omitted here. 

\subsection{Arithmetic Dataset Task}\label{subsec:adt-setup}
\textbf{Motivation:} This non-trivial synthetic task assesses if a multiplication module can pass useful gradients to learn the upstream layer. 
We first focus on learning linear units before moving onto more complex image based tasks where the solution of upstream layers is non-trivial.

\textbf{Setup.} This task requires learning a two-layer NALM for performing addition (in the first layer) followed by multiplication (in the second layer). 
The input vector, for a single unbatched data sample, consists of a 100 floating point numbers drawn from a uniform distribution with an upper and lower bound. 
The addition module must learn to select and sum two different overlapping subsets of this vector. 
Each subset is created from a continuous slice of the input between a lower and upper index. 
The proportion of a subset in comparison to the input and overlap ratio between the two subsets are set as 0.25 and 0.5 respectively. 
The subsets selected remain consistent for a batch but can vary for different seeds. 
Each seed uses a different model initialisation and a different training dataset for a range. 
The multiplication module must select both output values of the addition module and multiply them to give the predicted output value. 
We use \citet{madsen2020neural}'s hyperparameters as these values are selected from parameter sweeps. 

\begin{table}[]
\centering
\caption{Interpolation (train/validation) and extrapolation (test) ranges used. Data (as floats) is drawn from a Uniform distribution with the range values as the lower and upper bounds.}
\vspace{1em}
\label{tab:SLTR-ranges}
\begin{tabular}{llllll}
\toprule
\textbf{Interpolation} & {[}-20, -10) & {[}-2, -1) & {[}-1.2, -1.1) & {[}-0.2, -0.1) & {[}-2, 2)                  \\
\textbf{Extrapolation} & {[}-40, -20) & {[}-6, -2) & {[}-6.1, -1.2) & {[}-2, -0.2)   & {[}{[}-6, -2), {[}2, 6){]} \\\midrule
\textbf{Interpolation} & {[}0.1, 0.2) & {[}1, 2) & {[}1.1, 1.2) & {[}10, 20) &  \\
\textbf{Extrapolation} & {[}0.2, 2)   & {[}2, 6) & {[}1.2, 6)   & {[}20, 40) & 
\\\bottomrule
\end{tabular}
\end{table}

\begin{figure*}[bt]
% \vskip 0.1in
\centering
\includegraphics[width=0.9\textwidth]{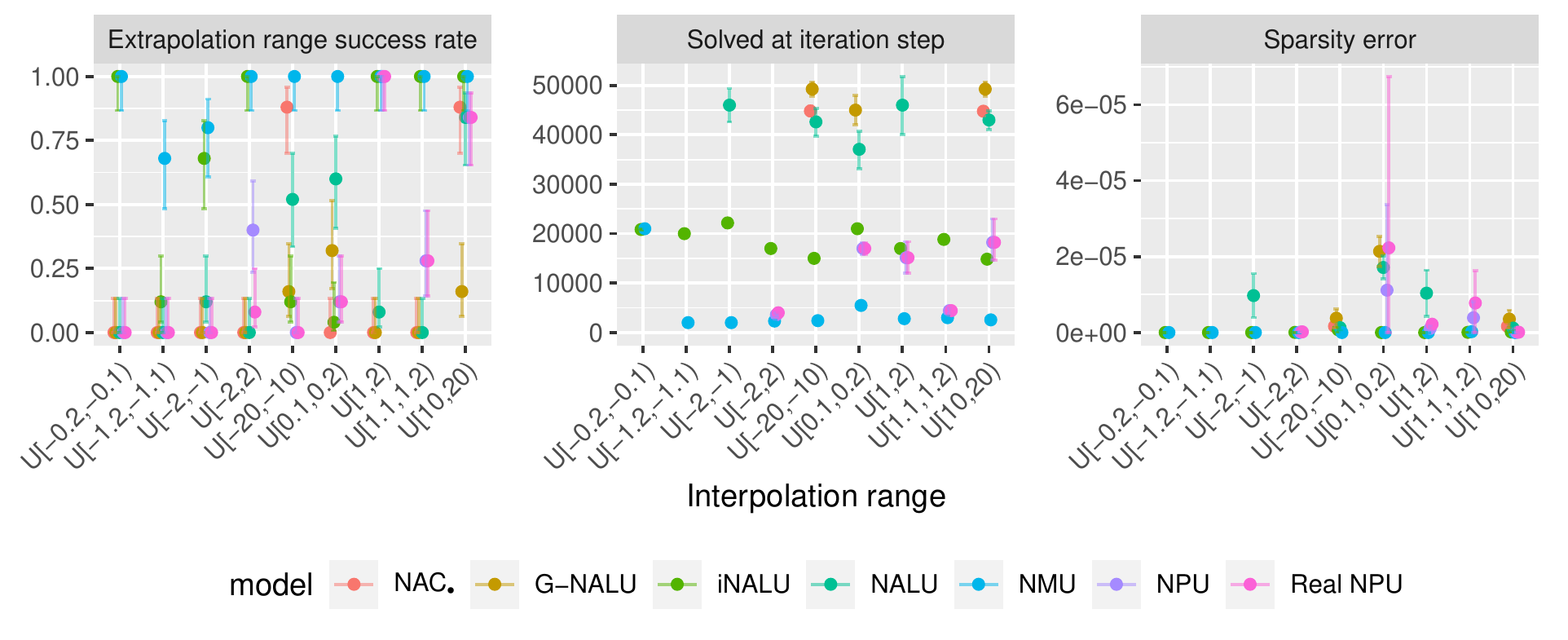}
\caption{Single Module Task for multiplication.}
\label{fig:slt-range-mul}
% \vskip -0.1in
\end{figure*}

\begin{figure}[!b]
\centering
\includegraphics[width=0.5\textwidth]{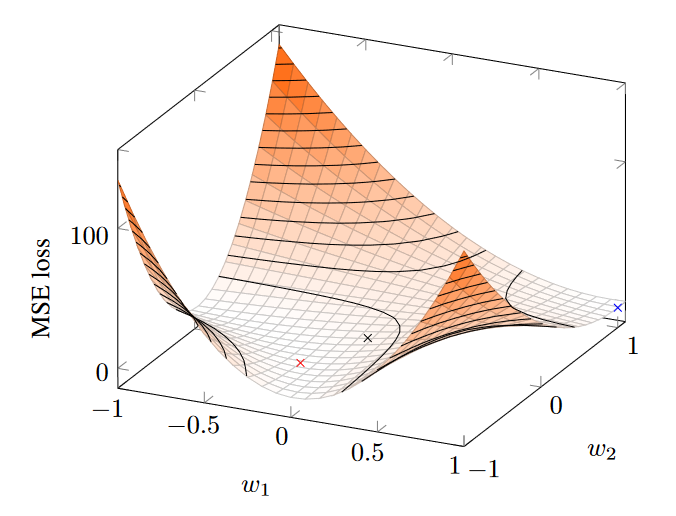}
% \centering
% \begin{tikzpicture}[scale=0.7]
%     \begin{axis}[domain=-1:1,y domain=-1:1.1,
%     xlabel=$w_1$,
%     ylabel=$w_2$,
%     zlabel=MSE loss,
%     colormap name=whitered,]
%     \addplot3[surf] {(-2*-1.8 - ((-2*x +1-x)*(-1.8*y +1-y)))^2};
%     \addplot3[ contour gnuplot = {number=20, labels={false},
%       draw color = black}, samples = 21,] {(-2*-1.8 - ((-2*x +1-x)*(-1.8*y +1-y)))^2};
%     \draw (-0.1667,-0.5) node[cross,red] {};    % local min W=[-1/6,-0.5]; pass for interp, fail on extrap
%     \draw (1,1) node[cross,blue] {}; % global min sol W=[1,1]
%     \draw (0,0) node[cross,black] {}; % global min sol W=[1,1]
%     \end{axis}
    
% \end{tikzpicture}
\caption{Static Loss Landscape with batch size of 1 for NMU weights in a Single Module Task for learning `$-2 \times -1.8$'. Ideally the weights should converge to the global minima (1,1) (blue cross) which is the extrapolative solution. However, an alternate minima at (-$\frac{1}{6}$,-0.5) (red cross) exists which solves $-2\times -1.8$ but will not extrapolate. Furthermore, since the weights for this minima are $< 0.5$ the model will stop at (0, 0) (black cross) due to weight clipping and regularisation.} \label{fig:3d-loss}
\end{figure}

\subsection{Evaluation Metrics}
To quantitatively evaluate the modules, we use \citet{maep-madsen-johansen-2019}'s evaluation scheme. 
We assess the NALMs over multiple ranges. 
Interpolation (training/validation) and extrapolation (test) ranges are presented in Table~\ref{tab:SLTR-ranges} and are chosen based on previous work~\citep{madsen2020neural} which requires the interpolation and extrapolation ranges not to overlap in order to test out-of-distribution performance. 
Early stopping is applied using a validation dataset sampled from the interpolation range. 

The three evaluation metrics are: the success on the extrapolation dataset against a near optimal solution (\textit{success rate}), the first iteration which the task is considered solved (\textit{speed of convergence}), and the extent of discretisation of the weights (\textit{sparsity error}). 
Sparsity error calculated by $\max\limits_{i,o}(\min(|W_{i,o}|, 1-|W_{i,o}|))$, measures the NALM weight element which is the furthest away from the acceptable discrete weights for a NALM. 
A success means the MSE of the trained model is lower than a threshold value (i.e. the MSE of a near optimal solution). 
For the Arithmetic Dataset Task, the threshold is a simulated MSE on 1,000,000 data samples using a model where each weight of the addition is off an optimal weight value by $\epsilon=$1e-5. 
The Single Module Task also uses a simulated threshold value with an $\epsilon=$1e-5 (see Appendix~\ref{app:sltr-eval} for details).  
95\% confidence intervals are calculated from a specific family of distributions dependant on the metric. 
The success rate uses a Binomial distribution because trials (i.e. run on a single seed) are pass/fail situations. 
The convergence metric uses a Gamma distribution and sparsity error uses a Beta distribution. 

\section{Robustness Issues with Multiplication Modules}\label{sec:single-layer-task}
Using the Single Module Task (Section~\ref{sec:exp-setup-and-eval}), we highlight robustness issues when training on different inputs and relate the cause of the problem to local optimas.
Results are displayed in Figure~\ref{fig:slt-range-mul}. 
No module has full success on all ranges, with all modules completely failing on small negative input ranges such as $\mathcal{U}$[-1.2,-1.1) and $\mathcal{U}$[-2,~-1). 
Though NPUs in theory can learn with negative inputs, empirical results suggest the module struggles. 
The $\mathrm{NAC}_{\bullet}$ outperforms the NALU on all ranges except $\mathcal{U}$[1,2), $\mathcal{U}$[10,20), and $\mathcal{U}$[-2,2) where both successes are equal. 
The failure of $\mathcal{U}$[-2,2) is expected due to the $\mathrm{NAC}_{\bullet}$'s inability to deal with mixed signed inputs, caused by the module using a log-exponential transformation in its architecture. 
Next, we reason as to why the NMU fails. We focus on the NMU because it has the best performance overall.

\paragraph{Problem: Inputs that Induce Local Optima.}%\label{sec:problem-local-optima} 
The NMU fails to get 100\% success for interpolation ranges $\mathcal{U}$[-1.2,-1.1) and $\mathcal{U}$[-2,-1).
We find weights converge towards other local optima with values outside the applied clipping range of [0, 1]. 
% Weights pass above the expected threshold to be pulled to a discrete value but not the desired value. 
Figure~\ref{fig:3d-loss} illustrates this issue. 
Such optima can be considered global for some interpolation cases (achieving a low enough train loss to be considered a solution), but are local optima for extrapolation cases (with high test errors). 
As the input range is an independent variable in our experiments, we conclude that the input data influences the weight learning towards local minimas which get wrongly enforced by the model's bias towards discrete weights (from regularisation and weight clipping). 

\section{Stochastic NMU (sNMU)}\label{sec:sNMU}
\begin{figure*}[t]
\centering
\includegraphics[width=0.55\textwidth]{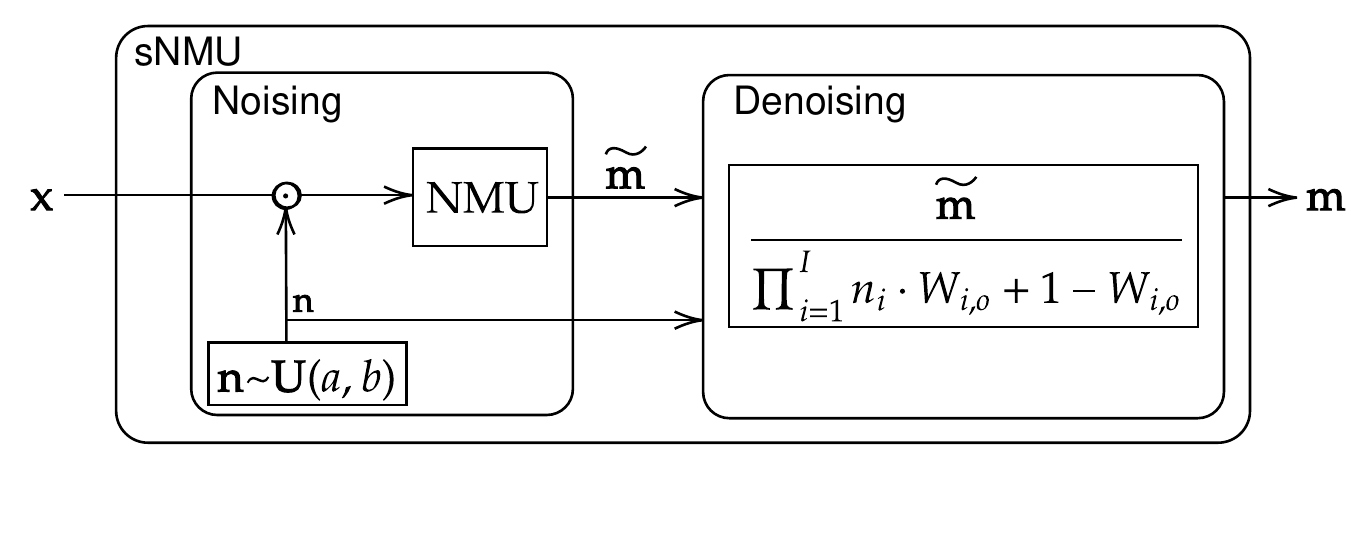}
\caption{Stochastic NMU architecture}
\label{fig:snmu}
\end{figure*}

We propose the stochastic NMU (sNMU), Figure~\ref{fig:snmu} to solve the above issue. 
This sNMU achieves full success on the Single Module Task on all ranges (Appendix~\ref{app:sltr-sNMU}) without compromising the existing advantages of the NMU i.e., the low parameter count, fast solving speeds and low sparsity errors. 

There are two stages to the sNMU: (1) \textit{Noising} to apply noise to the inputs of the NMU and (2) \textit{denoising} the output of the NMU to cancel the effect of the introduced noise. 
The resulting output value would be as though the input is applied to the NMU without noise. 
As gradients are influenced by the inputs, by manipulating the input with noise, the resulting gradients are more likely to favour the correct optimum and has a loss landscape with fewer optima near the subspace of desired weight values which only solve the interpolation range. 
The two stages are detailed below:

\textbf{Noising:} Noise $n_i$ is sampled from $\mathcal{U}[a,b]$ (where a and b are predetermined) and multiplied with each input, $x_i$:  
\begin{equation}
\textrm{NMU}_{\mathrm{noisy}}: \tilde{m}_o = \prod_{i=1}^{I}(n_{i}x_{i}W_{i,o} + 1 - W_{i,o}) \;. \label{eq:nmu-noise}
\end{equation}

\textbf{Denoising:} 
Only dividing by the cumulative noise would not fully reverse the effects if there are redundant inputs. To fully cancel the effect of the noise, the output is divided by a denoising factor which induces a bias in the weights forcing them towards being either 0 or 1:
\begin{equation}
\textrm{sNMU}: m_o = \frac{\tilde{m}_o}{\prod_{i=1}^{I}(n_{i}W_{i,o} + 1 - W_{i,o})} \;. \label{eq:nmu-denoise}
\end{equation}
The noising the denoising is only used during training; during inference, the module will act exactly like a NMU. 
To the best of our knowledge, our paper is the first in using reversible noise in a NALM. 
The sNMU weight values remain interpretable as 0 refers to ignoring an input and 1 refers to multiplying the input. 
Hence, we know with full confidence which inputs will be multiplied. 
In other words, if weights converge correctly, the sNMU acts as an extrapolative multiplication module which works on any valid input value. 
For additional discussion, refer to Appendix~\ref{app:sNMU-QA}. 

\section{Arithmetic Dataset Task}\label{sec:arithmetic-ds-task}
\begin{figure*}[t]
% \vskip 0.1in
\centering
\includegraphics[height=5cm, width=0.9\textwidth]{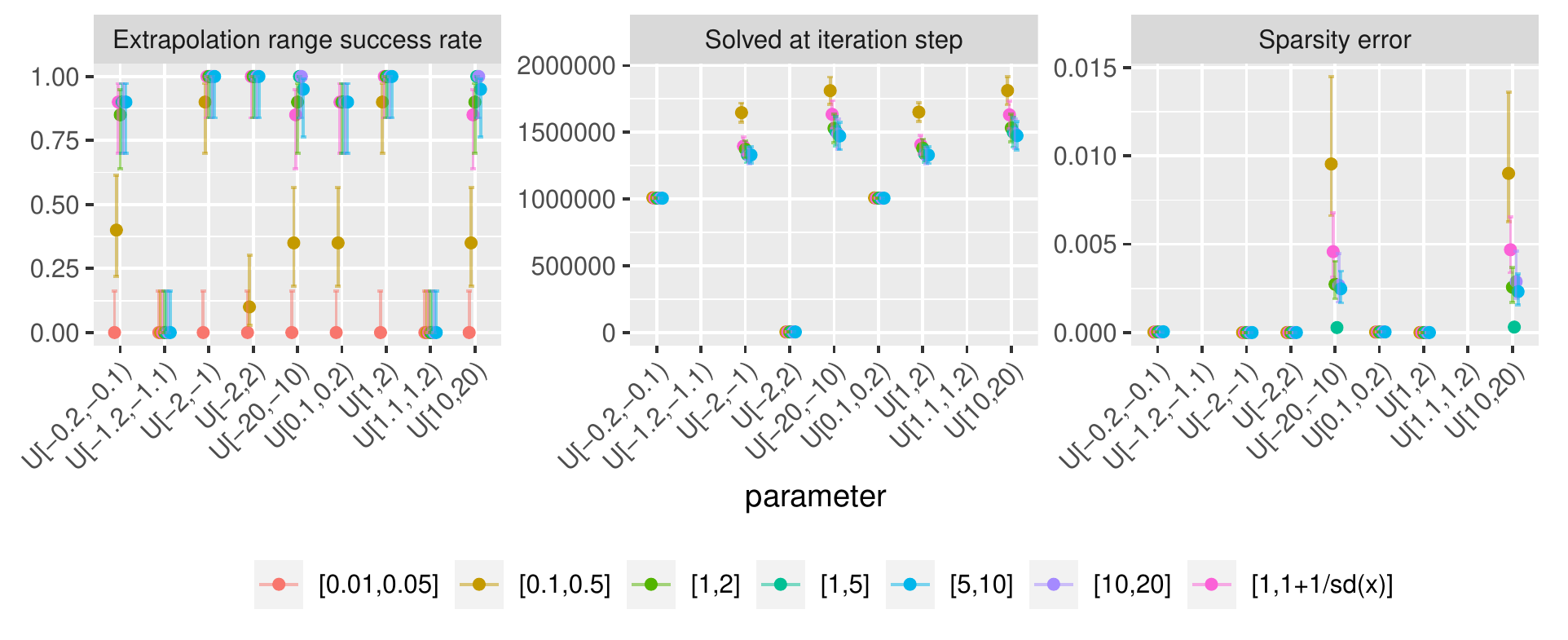}
\caption{Arithmetic Dataset Task for multiplication over various noise ranges for the NAU-sNMU.}
\label{fig:FTS-snmu-noise-ranges}
% \vskip -0.1in
\end{figure*}

We now evaluate against the more complex two-layer task (see Section~\ref{subsec:adt-setup}), presenting results for the stacked NAU-NMU and NAU-sNMU. 
We first compare the effects of different noise ranges and take the best performing noise to compare against the NMU based model. 

\textbf{Effect of noise range.}
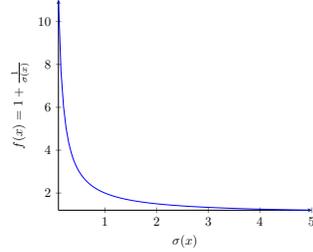
\begin{wrapfigure}{R}{0.3\columnwidth} %R for right
\vspace{-2.5em}
   \resizebox{0.3\columnwidth}{!}{%
    \begin{tikzpicture}
    \begin{axis}[
        axis lines = left,
        xlabel = $\sigma(x)$,
        ylabel = {$f(x)=1+\frac{1}{\sigma(x)}$},
    ]
    %Below the red parabola is defined
    \addplot [
        domain=0.1:5, 
        samples=100, 
        color=blue,
    ]
    {1 + (1/x)};
    \end{axis}
    \end{tikzpicture}
}
   \caption{Upper bounds for batch statistic noise}
   \label{fig:batch-snmu-upper}
  \vskip -0.1in
\end{wrapfigure}
Figure~\ref{fig:FTS-snmu-noise-ranges} shows the effect of using different noise ranges for the sNMU. 
Smaller ranges under one perform worse for data generated with a Uniform distribution and too large a range also shows degradation on the performance in success and sparsity on larger training ranges. 
Using \textit{batch noise}, which automates the noise range by using batch statistics to sample from $\mathcal{U}$[1, $1 + \frac{1}{\sigma(\bf{x})}$] (Figure~\ref{fig:batch-snmu-upper}), where $\sigma(x)$ is the standard deviation of the sNMU input $\mathbf{x}$, also achieves reasonable results, but the range $\mathcal{U}$[1,5] is found to be the best for this task in regards to the three metrics. 

\begin{figure*}[h]
% \vskip 0.1in
\centering
\includegraphics[height=5cm, width=0.85\textwidth]{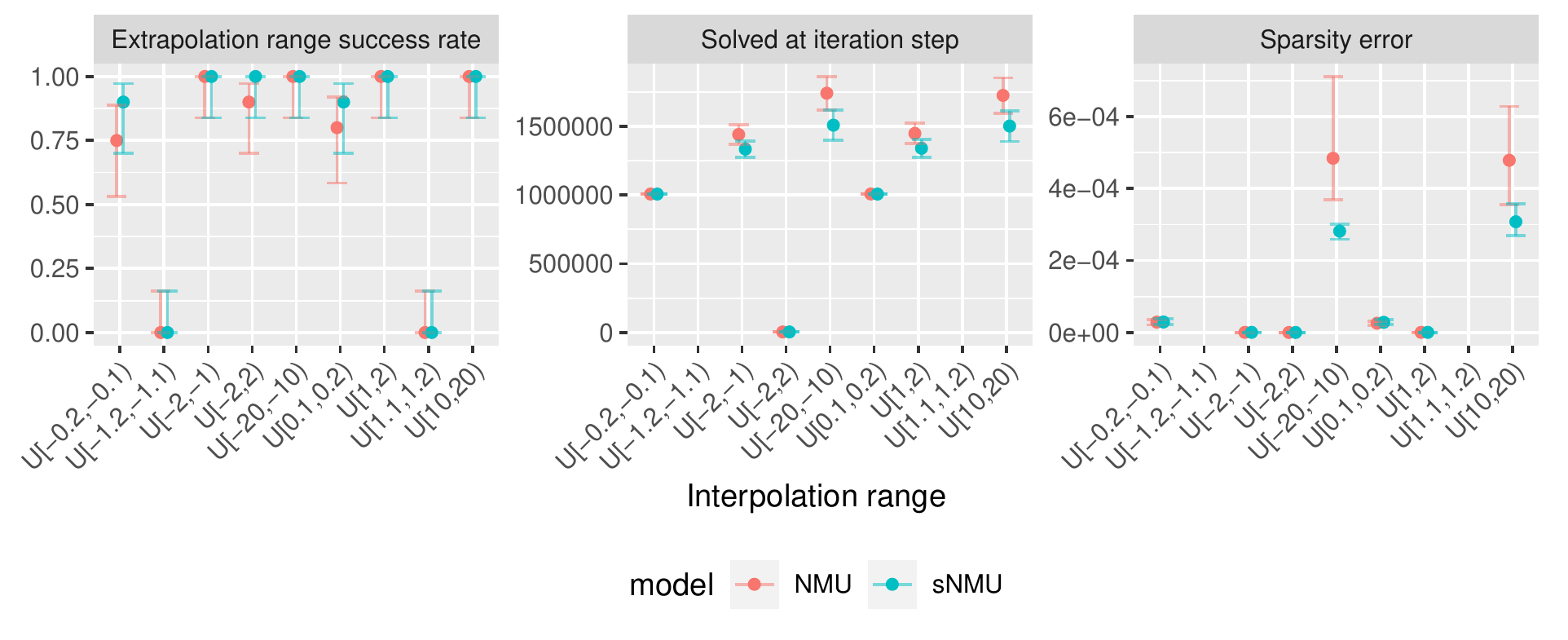}
\caption{Arithmetic Dataset Task for multiplication. Models include: NMU (stacked NAU-NMU) and sNMU (stacked NAU-sNMU) with a noise range of [1,5].}
\label{fig:2-layer-results}
% \vskip -0.1in
\end{figure*}

\textbf{NMU vs sNMU.} 
Figure~\ref{fig:2-layer-results} shows the stacked NAU-NMU fails on multiple ranges. 
Similar to the Single Module Arithmetic Task, the range $\mathcal{U}$[-2,2) is solved instantly. 
The sNMU shows improvement with faster solve speeds and lower sparsity errors compared to the NMU.
Furthermore, the sNMU fixes all failures in $\mathcal{U}$[-2,~2) and improves the success rate of $\mathcal{U}$[-0.2,-0.1) from 0.75 to 0.9 and $\mathcal{U}$[0.1,0.2) from 0.8 to 0.9. 
However, ranges $\mathcal{U}$[1.1,1.2) and $\mathcal{U}$[-1.2,-1.1) remain at a success rate of 0 for both models. 
The remaining failures can be explained due to the use of an MSE loss (which is out of the scope here, but is covered in Appendices~\ref{app:ADsT-failure} and \ref{app:ADT-grads}). 
\section{MNIST Arithmetic}\label{sec:mnist-arithmetic}
This section explores the effect of including specialist multiplication modules as a downstream layer for two image tasks: static MNIST product and sequential MNIST product. 
The static product task investigates learning to multiply images composed of two MNIST digits, and two variations of this task are explored: isolated digit classification and colour channel concatenated digit classification. 
The sequential product task investigates multiplying a sequence of MNIST images. 
Summaries of parameters and hardware/runtimes can be found in Appendices~\ref{app:exp-parameters} and \ref{app:hardware-and-timings}. 

\begin{figure}[t]
\centering
\subfloat[Isolated digits]{\includegraphics[height=4.25cm]{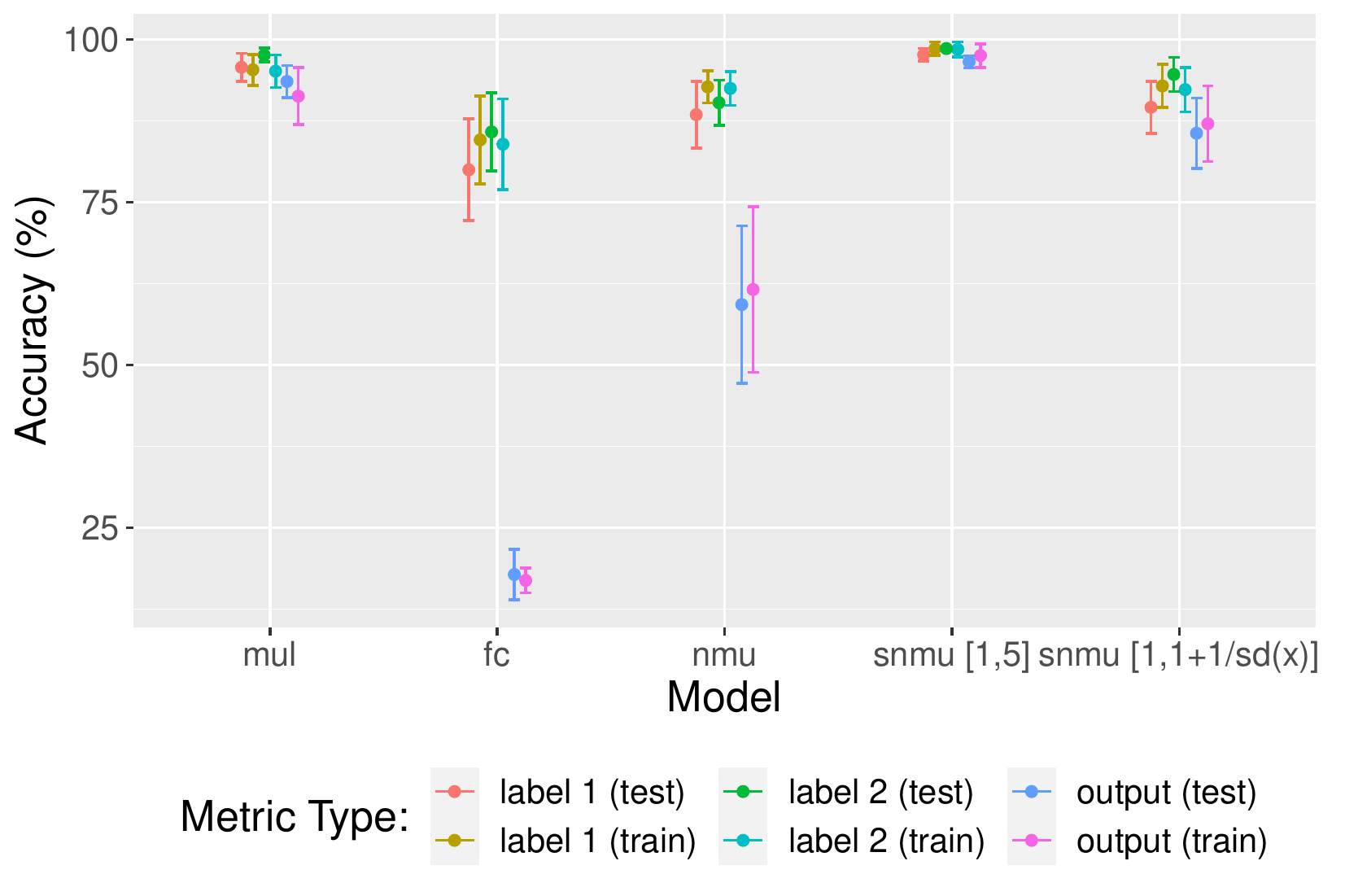}\label{subfig:1digit-mnist-conv-acc}} 
\hspace{0.8cm}%
\subfloat[Colour channel concatenated digits]{\includegraphics[height=4.25cm]{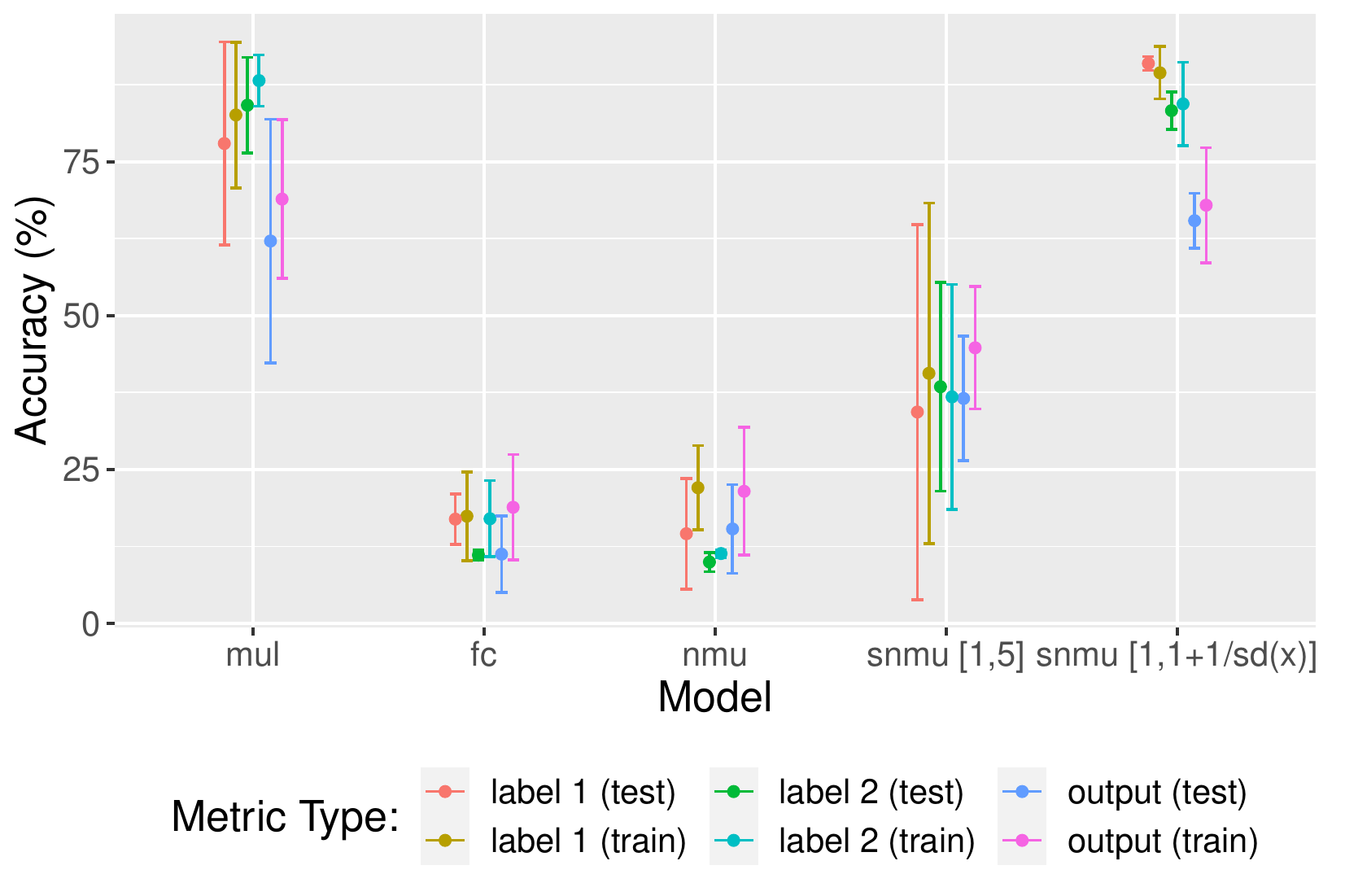}\label{subfig:mnist-st-tps-acc}}   
\caption{Isolated digit (left) and rounded colour channel concatenated (right) accuracies for classifying each of the two digits (label 1 and label 2) and the final product (output).}
\label{fig:2L-losses}
\end{figure}

\subsection{Isolated Digit Classification}\label{subsec:isolated-digit-clf}
\textbf{Motivation.} First, we determine if digit classification can be learnt in upstream layers in a simple setting where no image localisation is required. 

% How - exp setup
\textbf{Setup and network.} Following \citet{bloice2021Mnist}'s setup, the dataset contains permutation pairs of MNIST digits side-by-side with the target label being the product of the digits, e.g. input \includegraphics[height=2ex]{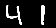} with output $4(=4\times 1)$. 
% A 90:10 train-test split is used in a 10-fold cross validation setting. 
Importantly, although there is no overlap between the permutation pairs in the train and test set, all individual digits (between 0-9) are seen during training. 
E.g., the pair `54' would exist in the test set and not the train set but the digits `5' and `4' would exist in other pairs of the train set such as `15' or `47'. 

% How - arch
The network learns a map from the input image to the labels of the two digits (digit classifier), followed by a map from the two labels to their product (multiplication layer). 
As the commutative property of multiplication can cause learning difficulties for the digit classifier, we separate the two digits to single digits, classify per digit and the recombine the two labels.  
The digit classifier is a convolutional network\footnote{From the PyTorch MNIST example \url{https://github.com/pytorch/examples/blob/master/mnist/main.py}}.\todo{which can classify single MNIST digits to an accuracy of XX} 
The multiplication layer is done in three different ways: (1) solved multiplication baseline model (MUL), (2) fully connected (FC) layer whose output is the product of the learnable weights, and (3) NALM: NMU/sNMU. 
For fair comparison the fully connected network uses the same initialisation scheme as the NMU. 
The FC layer uses weighted product accumulators rather than linear layers as the latter does not have the capacity to do multiplication.

% How - metrics 
\textbf{Metrics and results.} The MUL baseline only needs to learn to classify the images to their respective labels and therefore is considered a strong baseline. 
For a NALM to outperform the baseline would imply that the arithmetic inductive bias can aid learning of downstream layers. 
A strict criteria for measuring accuracy is used as the predictions are not processed in any way (e.g. rounded/truncated), hence a model must learn to apply the operation and classify the digits exactly. 
% results - which models. Key points
Figure~\ref{subfig:1digit-mnist-conv-acc} shows the results. 
% The sNMU [1,5] has the best results for individual label accuracies and the final output label. 
Both the sNMUs with U[1,5] noise and batch noise (96.6\% and 85.6\%) outperform the
NMU (59.3\%) for the test output metric, with no overlap in confidence bounds. 
The FC model learns to classify each label to a reasonable accuracy but does not learn the multiplication weights robustly, whereas the specialised modules, the NMU and sNMU do (see Figure~\ref{fig:1digit-mnist-weights-path} in Appendix~\ref{app:static-mnist-product}). 
The $\mathcal{U}$[1,5] sNMU outperforms both the NMU and MUL model, suggesting that the reversible stochasticity not only improves robustness but can aid with learning upstream layers. 

\subsection{Colour Channel Concatenated Digit Classification}\label{subsec:colour-channel-digit-clf}
\textbf{Motivation.} Confirming NALMs can be effective using simple digit classifiers, we now ask if this remains the case if the difficulty for the classification network is increased. 

\begin{figure*}[t]
% \vskip 0.1in
\centering
\includegraphics[height=4.5cm, width=\textwidth]{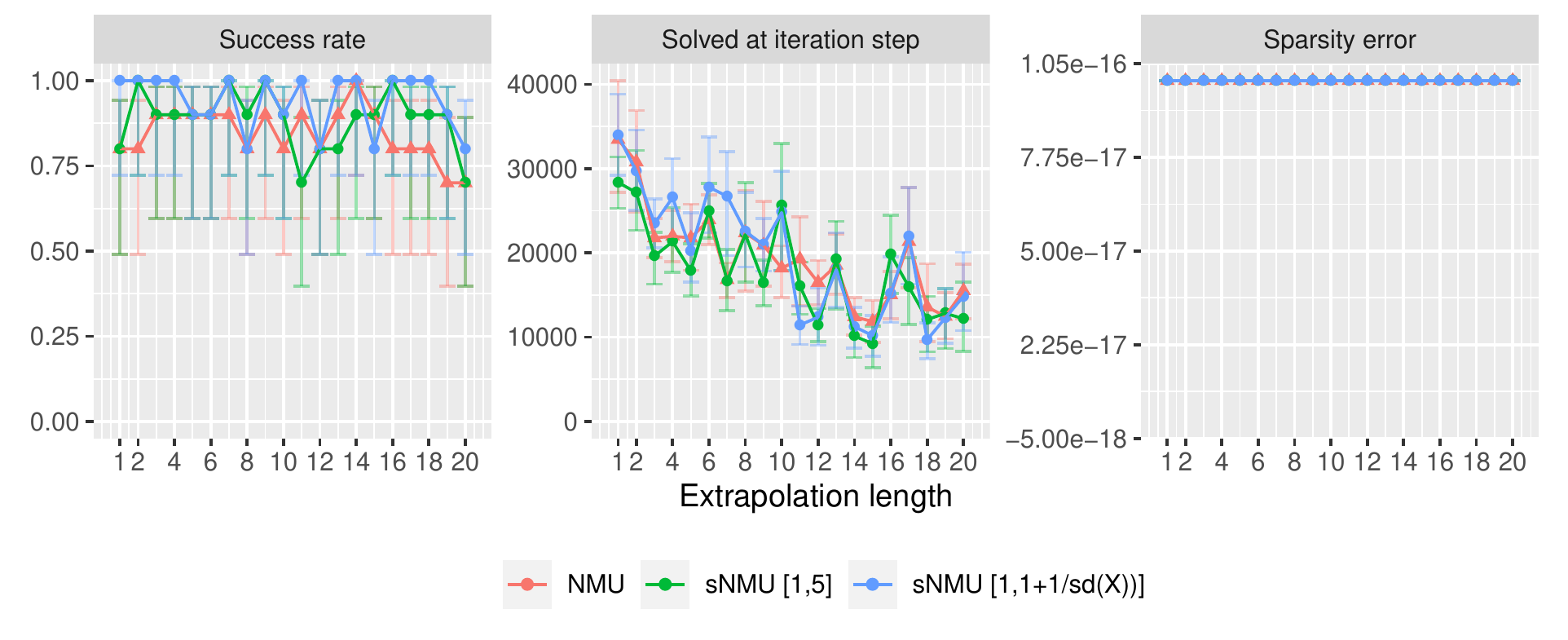}
\caption{Performance on Product of Sequential MNIST. Model names represent the type of multiplication cell used. All models use the same CNN architecture to do digit classification.}
\label{fig:seq-mnist-prod}
% \vskip -0.1in
\end{figure*}

\textbf{Setup and network.} Following~\citet{spatialTransformer2015}, random rotation, scaling and translation transforms are applied to the digits and the image classifier must learn to localise digits as images now contain both digits separated by the colour channel. 
The digit classifier uses a Spatial Transformer~\citep{spatialTransformer2015} with a Thin Plate Spline transformation for digit localisation~\citep{bookstein1989principal} which is bounded~\citep{shi2016robustTPS}. 
The multiplication layer uses the same options from Section~\ref{subsec:isolated-digit-clf} (i.e., MUL, FC, NMU and sNMU). 

\textbf{Metrics and results.} The accuracies of each digit label and the final output value are taken. Due to the increased classification difficulty, accuracies are rounded. 
Figure~\ref{subfig:mnist-st-tps-acc} shows that the sNMU with batch noise is able to get comparable test output accuracy to the solved baseline ($67.9\%$ vs $68.9\%$) with tighter confidence bounds, suggesting improved robustness for the digit classifier network. 
There is also an improvement (+13\%) in classifying the first digit. 
Using a noise range of $\mathcal{U}$[1,5] has a weaker performance in comparison to using the batch noise.
The NMU fails to converge towards the correct multiplication weights for a fold (see Figure~\ref{fig:mnist-st-tps-weights-path} in Appendix~\ref{app:static-mnist-product}), unlike its stochastic versions, achieving similar accuracies to the FC based network. 

\subsection{Sequential MNIST Product}\label{app:seq-mnist}
\textbf{Motivation.} To test the effect of the modules in a different extrapolative setting, a sequential task is adopted where the number of digits to multiply can be controlled. 

\textbf{Setup and network.} Following~\citet{madsen2020neural}, given a sequence of MNIST digits, process one image at a time using a classification network to convert an image to its label value, which gets passed into a recursive NALM cell to calculate the cumulative result. 
The NALM will take in two inputs: the predicted label of the image at the current timestep and the predicted accumulated value from the previous timesteps. 
A convolutional network is used to regress the images to digits. The multiplication layer is either solved (baseline), or requires learning a NALM (NMU or sNMU). 

\textbf{Metrics and results.} The success threshold is the $1\%$ one-sided upper confidence-interval using a student-t distribution over the MSE of solved NALM models. 
Training uses two-digit sequences while testing uses sequences up to 20 digits long.  
Figure~\ref{fig:seq-mnist-prod} shows the results with both sNMU networks outperforming the NMU over multiple extrapolation lengths while retaining fast convergence similar to the NMU. 
The batch sNMU underperforms in comparison to the NMU between sequence lengths 11-14, while the noise range $\mathcal{U}$[1,5] only underperforms on length 15. 

\section{Discussion}
We provide an explanation for the lack of robustness to training ranges of a numerical unit, leading us to propose the stochastic NMU (sNMU) resulting in improved extrapolation. 
As the process is fully reversible the injected noise does not hinder the ability of the module to learn the exact operation. 
The stochastic nature of the module provides gradients to avoid local optimas which a regular NMU falls into, while still propagating a useful signal to upstream layers as shown by performance on the arithmetic image tasks. 
Though we focus on providing a proof of concept for using reversible stochasticity to improve learning, this idea can be further explored. For example, using other distributions of noise, learnable noise ranges or other forms of batch statistics. Future investigation could also be done to learn why certain noise ranges work better than others. 

\begin{ack}
B.M. is supported by the EPSRC Doctoral Training Partnership
(EP/R513325/1). J.H. received funding from the EPSRC Centre for Spatial Computational
Learning (EP/S030069/1). The authors acknowledge the use of the IRIDIS High-
Performance Computing Facility, the ECS Alpha Cluster, and associated support services
at the University of Southampton in the completion of this work.
\end{ack}

\bibliographystyle{plainnat}
\bibliography{references}

\appendix
\newpage

\section{Regularisation used by the NAU and NMU}\label{app:nau-nmu-reg}
To enforce discretisation of weights both units have a regularisation penalty for a given period of training. The penalty is
\begin{equation}
\lambda \cdot \frac{1}{I \cdot O} \sum_{o=1}^{O} \sum_{i=1}^{I} \min\left(|W_{i,o}|, 1 - |W_{i,o}|\right) ,
\end{equation}
where $O$ is the number of outputs and $\lambda$ is defined as
\begin{equation}
\lambda = \hat{\lambda} \cdot \max\left(\min\left(\frac{iteration_i - \lambda_{start}}{\lambda_{end} - \lambda_{start}}, 1\right), 0\right) .
\label{eq:regualizer-scaling}
\end{equation}
Regularisation strength is scaled by a predefined $\hat{\lambda}$. The regularisation will grow from 0 to $\hat{\lambda}$ between iterations $\lambda_{start}$ and $\lambda_{end}$, after which it plateaus and remains at $\hat{\lambda}$.

\section{Experiment Parameters}\label{app:exp-parameters}
Refer to Tables~\ref{tab:exp-params}, \ref{tab:params-nmu-and-nau},  \ref{tab:sltr-npu-params} and \ref{tab:sltr-inalu-params} for the breakdown of parameters used in the Single Module Task and Arithmetic Dataset Task. 
Regularisation is only applied for a given period
of time to avoid forcing weights to converge too early/late. 
For example, if started to early then the weights will not have a chance to converge to the correct value in time for discretisation. 
If started too late there is a danger the discretisation will not complete before the maximum number of training iterations is reached. 

Refer to Table~\ref{tab:mnist-two-digit-exp-params} for the breakdown of parameters used in the Static MNIST Product experiments. 

Refer to~\citet{madsen2020neural} for the hyperparameters, setup, and regularisation used in the original experiment used in the Sequential MNIST Product experiments.

\begin{table}[h]
\caption{Parameters which are applied to all modules. Parameters have been split based on the experiment. $^*$Validation and test datasets generate one batch of samples at the start which gets used for evaluation for all iterations.}
\label{tab:exp-params}
\vskip 0.1in
\begin{center}
\begin{tabular}{p{3cm}p{4.5cm}p{4.5cm}}
\toprule
\textbf{Parameter}          & \textbf{Arithmetic Dataset Task} & \textbf{Single Module Task}     \\ \midrule
\textbf{Layers}             & 2                                & 1                              \\ 
\textbf{Input size}         & 100                              & 2                              \\ 
\textbf{Subset ratio}       & 0.25                             & 0.5                            \\ 
\textbf{Overlap ratio}      & 0.5                              & 0                              \\ 
\textbf{Total iterations}   & 5 million         & 50000                          \\ 
\textbf{Train samples}      & 128 per batch                    & 128 per batch                  \\ 
\textbf{Validation samples$^*$} & 10000                            & 10000                          \\ 
\textbf{Test samples$^*$}       & 10000                            & 10000                          \\ 
\textbf{Seeds}              & 20                               & 25                             \\ 
\textbf{Optimiser}          & Adam (with default parameters)   & Adam (with default parameters) \\ \bottomrule
\end{tabular}
\end{center}
\vskip -0.1in
\end{table}

\begin{table}[h]
\caption{Additional parameters for the NMU (and NAU) for the Single Module and Arithmetic Dataset Task. The $\hat{\lambda}$, $\lambda_{start}$, $\lambda_{end}$.} 
\label{tab:params-nmu-and-nau}
\vskip 0.1in
\begin{center}
\begin{tabular}{lp{1.75cm}p{1.75cm}}
\toprule
\textbf{Parameter}          & \textbf{Arithmetic Dataset Task} & \textbf{Single Module Task} \\ \midrule
$\hat{\lambda}$   & 10                               & 10                         \\
$\lambda_{start}$  & 1 million                        & 20000                      \\
$\lambda_{end}$   & 2 million                        & 35000                      \\
Learning rate     & 1.00E-03                         & 1.00E-03                   \\ \bottomrule
\end{tabular}
\end{center}
\vskip -0.1in
\end{table}

\begin{table}[h]
\centering
\caption{Parameters specific to the NPU and RealNPU modules for the Single Module Task.}
\label{tab:sltr-npu-params}
\vskip 0.1in
\begin{tabular}{ll}
\toprule
\textbf{Parameter}                & \textbf{Value} \\ \midrule
($\beta_{start}$,$\beta_{end}$) & (1e-7,1e-5)    \\ 
$\beta_{growth}$                & 10             \\  
$\beta_{step}$                  & 10000          \\  
Learning rate                   & 5.00E-03       \\ \bottomrule
\end{tabular}
\vskip -0.1in
\end{table}

\begin{table}[h]
\caption{Parameters specific to the iNALU for the Single Module Task.}
\label{tab:sltr-inalu-params}
\vskip 0.1in
\begin{center}
\begin{tabular}{ll}
\toprule
\textbf{Parameter}  & \textbf{Single Module Task} \\ \midrule
$\omega$     & 20\\ 
$t$          & 20\\ 
Gradient clip range     & [-0.1,0.1]\\     
Max stored losses (for reinitalisation check) & 5000\\
Minimum number of epochs before regularisation starts & 10000\\\bottomrule
\end{tabular}
\end{center}
\vskip -0.1in
\end{table}

\begin{table}[h!]
\caption{Static MNIST Product experiment parameters.}
\label{tab:mnist-two-digit-exp-params}
\vskip 0.1in
\begin{center}
\begin{tabular}{p{4cm}p{4.5cm}p{4.5cm}}
\toprule
\textbf{Parameter}          & \textbf{Two digit independent} & \textbf{Localisation with Spatial Transformers}     \\ \midrule
\textbf{Epochs}             & 1000                              & 1000                          \\ 
\textbf{Samples per permutation}    & 1000                      & 1100                          \\ 
\textbf{Train:Val:Test}       & 90:-:10                         & 51:15:34                      \\ 
\textbf{Batch Size}         & 128                              & 256                            \\ 
\textbf{Train samples}      & 90,000                           & 61,710                         \\ 
% \textbf{Validation samples} & -                            & 10,890                       \\ \hline
\textbf{Test samples}   & 10,000                            & 37,400                        \\ 
\textbf{Folds/Seeds}        & 10                               & 3                              \\ 
\textbf{Optimiser}          & Adam (with default parameters)   & Adam (with default parameters) \\ 
\textbf{Criterion}          & MSE                               & MSE                           \\ 
\textbf{Learning rate}          & 1e-3   & 1e-3 \\ 
\textbf{$\lambda_{start}-\lambda_{end}$ epochs}          & 30-40   & 30-40 \\ 
\textbf{$\hat{\lambda}$}          & 100   & 100 \\ \bottomrule
\end{tabular}
\end{center}
\vskip -0.1in
\end{table}

% \newpage
\section{Hardware and Time to Run Experiments}\label{app:hardware-and-timings}
All experiments for the Single Module Task and the Arithmetic Dataset Tasks were trained on the CPU, as training on GPUs takes considerably longer, using a 16 core CPU server with 125 GB memory 1.2 GHz processors. 
For any model, a single seed for a single training range can be completed within 5 minutes for the Single Module Task and within 4.5 hours for the Arithmetic Dataset Task. 
Timings are based on a single run rather than the runtime of a script execution because the queuing time from jobs when executing scripts is not relevant to the experiment timings. 

All experiments for the Arithmetic Dataset Tasks were trained using a single GeForce GTX 1080 GPU. 
For the Static MNIST experiments, a single fold can be completed in approximately  5 hours for the Isolated Digit setup experiment and 10.5 hours for the Colour Channel Concatenated Digit setup. 
The Sequential MNIST experiments runtimes and memory usage are found in Table~\ref{tab:mnist-timings}. 
\begin{table*}[t]
\centering
\caption{Time taken and GPU memory required to run Sequential MNIST experiments. Experiments are run over 10 seeds.}
\label{tab:mnist-timings}
\scalebox{0.8}{
\begin{tabular}{lllllp{3cm}p{1.5cm}}
\toprule
\textbf{Experiment} &
  \textbf{Model} &
  \textbf{Criterion} &
  \textbf{Device} &
  \textbf{Epochs} &
  \textbf{Approximate time for completing 1 seed/fold (hh:mm:ss)} &
  \textbf{GPU memory (MiB)} \\ \midrule
\multirow{4}{2cm}{Sequential MNIST Product} &
  Reference &
  \multirow{4}{*}{MSE} &
  \multirow{4}{*}{GPU} &
  \multirow{4}{*}{1000} &
  02:00:00 &
  \multirow{4}{*}{679} \\ %\cline{2-2} \cline{6-6}
 & NMU                     &    &  &  & 02:55:00 &  \\ %\cline{2-2} \cline{6-6}
 & sNMU $\mathcal{U}$[1,5] &    &  &  & 03:00:00 &  \\ %\cline{2-3} \cline{6-6} 
 & sNMU $\mathcal{U}$[1,1+1/sd(x)] &  &  &  & 03:10:00 &  \\ %\cline{2-3} \cline{6-6}
 \bottomrule
\end{tabular}
}
\vskip -0.1in
\end{table*}

\newpage
\section{Single Module Task Evaluation:}\label{app:sltr-eval}
We adopt the \citet{maep-madsen-johansen-2019}'s evaluation scheme used for the Arithmetic Dataset Task but adapt the expression used to generate the predictions of an $\epsilon$-perfect model ($y_o^\epsilon$). For multiplication, the expression used is: 

\begin{align}
y_o^\epsilon &= (x_1x_2)(1-\epsilon)^2 \times \prod_{i\in X_{irr}}(1 - |x_i|\epsilon) 
\end{align}

Assume $x_1$ and $x_2$ are the operands to apply the operation to and any remaining features ($x_3,...,x_n$) be irrelevant to the calculation and part of the set $X_{irr}$. We use $I$ to denote the total number of input features. 
The $\epsilon$ for each feature will contribute some error towards the prediction. 
A simulated MSE is then generated with an $\epsilon=1e-5$ and used as the threshold value to determine if a NALM converges successfully for a particular range by comparing the NALMs extrapolation error against the threshold value.

\section{Single Module Task for sNMU on Multiplication}\label{app:sltr-sNMU}
\begin{figure*}[h!]
\vskip 0.1in
\centering
\includegraphics[width=0.8\textwidth]{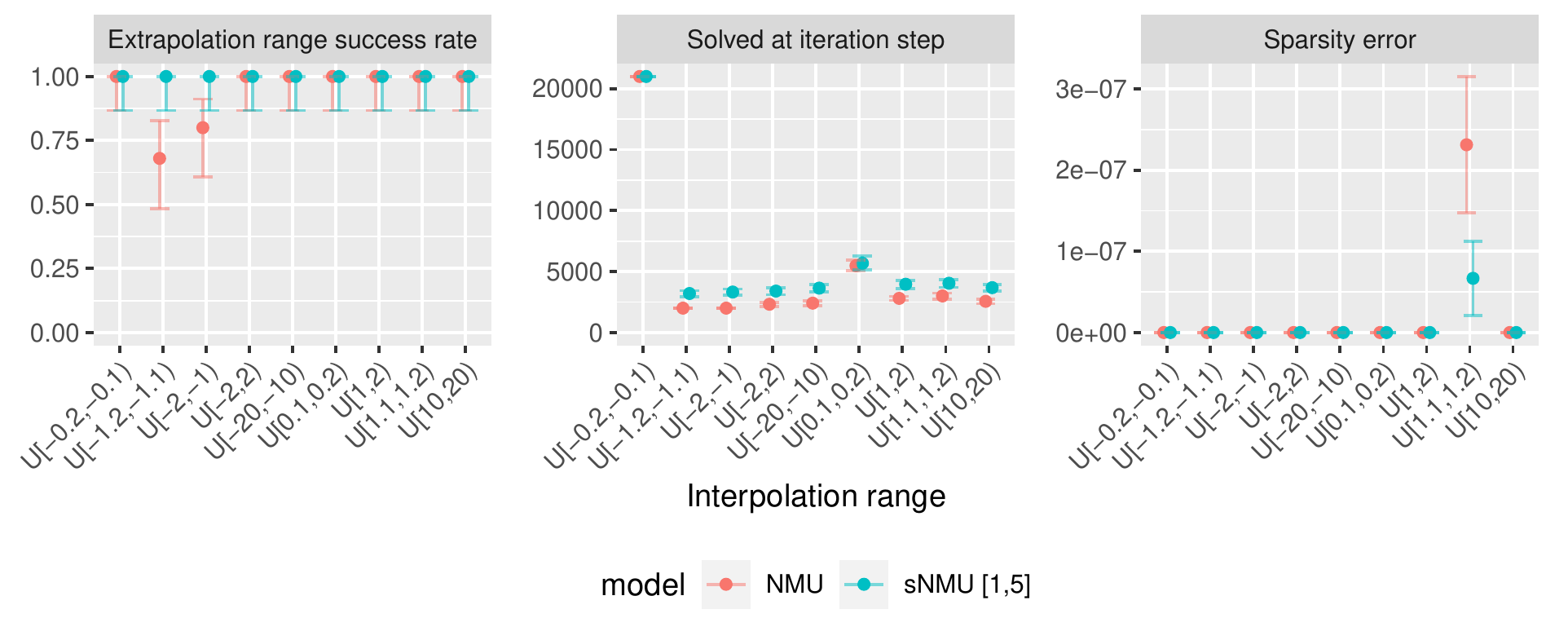}
\caption{Single Module Task for multiplication comparing the NMU to a stochastic NMU (sNMU) with noise sampled from U[1,5].}
\label{fig:slt-mul-snmu}
\vskip -0.1in
\end{figure*}

Figure~\ref{fig:slt-mul-snmu} shows the robustness of the sNMU on the Single Module Task. The sNMU achieves full success on all training ranges and solves the ranges within a reasonable number of iterations.

\section{Stochastic NMU - Additional Thoughts}\label{app:sNMU-QA}
\textbf{Why not just divide by the noise values?} Denoising using only the product of the noise values ($\prod_{i=1}^{I}n_i$) is not valid for cases with redundant inputs (i.e. $w_i=0$), where not all inputs are selected for multiplication. 
To alleviate the redundancy issue we multiply the noise with the weight values ($n_iW_{i,o}$). 
However, this causes a division by 0 if weight values are 0 or weight/noise values are close to 0 where numeric precision errors occur. 
Therefore, we include the $+1-W_{i,o}$ term, resulting in lowest possible value being the product of the noise. 
As the noise distribution is predefined, the lower bound of the noise can be controlled, avoiding issues with very small noise values. 
% Though less interpretable combinations of weights do exist, the discrete values would also solve the problem which is less likely with the non-discrete weights. 

\textbf{What about other denoising factors?} The linearity of the denominator in the chosen denoising factor is required for robust convergence. For example, an alternate mathematically valid denoising factor could be 
\begin{align}
\frac{m_o^{(n)}}{\prod_{i=1}^{I} n_{i}^{w_{i}}} , \label{eq:nmu-denoise-expo}
\end{align}
where each partial calculation of the product in the denominator is a value between [1, $n_i$]. 
This form of denoising was found to encourage the model to converge to optima which could solve interpolation data but not the extrapolation data. 
Furthermore, issues regarding stability occur for noise values close to 0, and negative noise values are not allowed due to causing complex-valued solutions if weight values are between 0 and 1.  

\textbf{Are the inputs/outputs normalised?} We do not normalise, rather part of our novel step is to completely reverse the effect of the stochasticity. 
Generic normalisation cannot work for such modules as multiplying normalised inputs gives the incorrect output and unnormalising to the correct output would require prior knowledge of the task. 

\textbf{Why apply stochasticity at the module level, and not the full network level?} When the modules are used as part of a larger end-to-end network the resulting feed-forward expression of the network can become quite complex making it difficult to denoise. At module level, this complexity is vastly reduced. 

\textbf{Motivations to real-world tasks:} 
Our proposed module, the sNMU, are part of Neural Arithmetic Logic Modules (NALMs) which are independent reusable modules. Such modules can be integrated into networks for specific applications, such as \citet{Zhang2019LivesmartAQ}’s deep reinforcement network which schedules content-delivery-networks using a NALM’s extrapolative ability to reduce failures outside the training range. Their work uses a module, whose multiplication unit could be replaced with the sNMU allowing for improved robustness. 

\textbf{How are gradients affected by the stochasticity?} For the two layer arithmetic dataset task, compare the derivations for gradient updates for the NAU-NMU and NAU-sNMU models (see Appendix~\ref{app:ADT-grads}), finding that the noise results in additional scaling of the gradients which does not exist in the NMU version. 

\textbf{Intuition behind the automatic batch statistic based noise range.} 
The automatic noise range samples noise from a uniform distribution in which the lower bound is 1 and the upper bound is $1 + 1/\sigma(x)$. 
The lower bound is set to 1 so the noise cannot scale down the gradient magnitude. 
The upper bound function (see Figure 6 of the main body) models an exponential decay curve where data batches with larger standard deviations result in smaller noise values while smaller standard deviations have much larger values. 
The intuition behind this is when the data distribution has similar magnitude samples (i.e., small standard deviations) it is easier for the module to confuse the different input features therefore having more noise makes it easier to differentiate between them. 

\section{Arithmetic Dataset Task Cause of Failure: Uninformative MSE loss}\label{app:ADsT-failure}
From Section~\ref{sec:arithmetic-ds-task} we saw certain training ranges resulted in no success for any model used. 
These failures are due to the input range causing difficulty for the NAU weights to select the relevant inputs. 
This is explained by the following scenario. 
Imagine the same task but with input size 4 and overlap and subset ratio of 0.5. 
Like before the aim is to select and add two different subsets of the input and multiply them together. 
For this specific case, we want the first subset to sum the 2nd and 3rd elements and the second subset to sum the 3rd and 4th elements. 
Consider two inputs, one sampled from $\mathcal{U}$[1,5] ($\bm{i}_{1}=$[1, 2, 3, 4]) and another from $\mathcal{U}$[1.1,1.2] ($\bm{i}_{2}=$[1.11, 1.12, 1.13, 1.14]). Using $\bm{i}_1$ as input and assuming weights select the correct inputs and converge as expected, we get the following: 
\begin{equation*}
\bm{i}_1\bm{W}^{(\text{NAU})}\bm{W}^{(\text{NMU})} =
\begin{bmatrix}
1 & 2 & 3 & 4
\end{bmatrix}\begin{bmatrix}
0 & 0\\
1 & 0\\
1 & 1\\
0 & 1
\end{bmatrix}\begin{bmatrix}
1\\
1
\end{bmatrix} = 35\\
% &= (2+3) \times (3+4) \\
% &= 35
\label{expr:i1-output}
\end{equation*}
Now consider cases: 1) NAU selection is correct but one weight did not converge; and 2) NAU selection is incorrect for one element and that weight did not converge. 
For each case a valid example of the NAU weight matrix ($\bm{W}^{(\text{NAU})}$) is:  
\begin{center}
Case 1: $\displaystyle \begin{bmatrix}
0 & 0\\
0.9 & 0\\
1 & 1\\
0 & 1
\end{bmatrix}$ \quad Case 2: $\displaystyle \begin{bmatrix}
0 & 0\\
0 & 0\\
1 & 1\\
0.9 & 1
\end{bmatrix}$ 
\end{center}

\begin{table}[t]
\caption{Output values and absolute errors for simplified 2-layer task inputs $\bm{i}_1$ [1, 2, 3, 4] and $\bm{i}_2$ [1.11, 1.12, 1.13, 1.14]. Selection is if the NAU module selected the correct inputs. Weights is if the weights for the NAU are converged to the correct values.}
\label{tab:weight-selection}
\vskip 0.1in
\begin{center}
\begin{small}
\begin{sc}
    \begin{tabular}{llllll}
    \hline
    \textbf{Case}                                                                            & \textbf{$\bm{i}_{1}$ Out} & \textbf{$\bm{i}_{1}$ AE} & \textbf{$\bm{i}_{2}$ Out} & \textbf{$\bm{i}_{2}$ AE} \\ \hline
    \textbf{\begin{tabular}[c]{@{}l@{}}Selection \cmark \\ Weights \cmark\end{tabular}}    & 35          & 0              & 5.1075      & 0              \\ \hline
    \textbf{\begin{tabular}[c]{@{}l@{}}Selection \cmark\\ Weights \xmark\end{tabular}}   & 33.6        & 1.4            & 4.85326     & 0.25424        \\ \hline
    \textbf{\begin{tabular}[c]{@{}l@{}}Selection \xmark\\ Weights \xmark\end{tabular}} & 46.2        & 11.2           & 4.89412     & 0.21338        \\ \hline
    \end{tabular}
\end{sc}
\end{small}
\end{center}
\vskip -0.1in
\end{table}

Calculating the output and the absolute error (from the ideal solution) of these cases for the both inputs (Table~\ref{tab:weight-selection}) shows that [1.11, 1.12, 1.13, 1.14] has a much smaller difference in error than [1, 2, 3, 4]. 
The model will struggle to differentiate between correct and incorrect selection of weights for the input drawn from the distribution with a smaller range. 
This specific case also shows that the selection of an incorrect input and non-converged weight gives lower error than the case with the correct selection and non-converged weight, suggesting that the MSE calculation cannot differentiate between a better and worse solution. 

The gradients of the loss also contribute to the invalid NAU weights (see Appendix~\ref{app:ADT-grads} for derivations). 
Considering the stacked NAU-NMU, the gradients of both the NAU and NMU weight matrix is scaled by a residual factor $(\bm{y}-\bm{\hat{y}})$. 
We know that input ranges with little variance can be trained to small training errors which give the illusion of a solved model. 
Therefore, by multiplying the small residual, the gradient gets scaled to very small values. 
When using the NAU-sNMU the gradients get scaled by a noise term (whose magnitude can be predefined to $>1$). However, the residual term still remains, creating small gradients. 

\section{Arithmetic Dataset Task Partial Derivatives}\label{app:ADT-grads}
This section derives the generalised partial derivatives for the stacked NAU-NMU and NAU-sNMU. 
To keep derivations as simple as possible, we assume formulations of models without the use of regularisation or clipping. 

Throughout this section, we assume the following notations:
\begin{itemize}
    \item Superscript A ($\bm{W}^A$) = Weight matrix of a summative module (i.e. NAU)
    \item Superscript M ($\bm{W}^M$) = Weight matrix of a multiplicative module (i.e. NMU or sNMU) 
    \item Weight matrix indexing $W_{r,c}$ where $r$ = row index and $c$ = column index (starting at 1)
    \item $I$ = total number of input elements for the respective module
    \item $O$ = output size of the NAU weight matrix (or number of elements in the intermediate vector)
    \item $l$ = index for an output element 
    \item $i$ = index for an input element
\end{itemize}

\subsection{MSE Loss for the Arithmetic Dataset Task} 
We define the MSE loss specific to the two-layer task below. 
$N$ is the number of batch items. 
$\bm{X}_n$ is the input vector for batch item $n$ with target scalar $y_n$ and predicted scalar $\hat{y}_n$. 
$I$ is the number of input elements (=100) in the input vector. 
\begin{equation*}
\begin{aligned}
L &= \frac{1}{N}\sum_{n}^{N}(y_n-\hat{y}_n)^2\\
% \frac{\partial L}{\partial \hat{y_n}} &= 2\sum_{n}^{N}(y_n-\hat{y}_n)(y_n-\hat{y}_n)^\prime\\
L &= \frac{1}{N}\sum_{n}^{N}(y_n-\text{NMU}(\text{NAU}(\bm{X_{n}})))^2\\
z_{1} &= \sum_{i}^{I}(X_{n,i} \cdot W_{i,1}^A)\\
z_{2} &= \sum_{i}^{I}(X_{n,i} \cdot W_{i,2}^A)\\
L &= \frac{1}{N}\sum_{n}^{N}[y_n-((1+W_{1,1}^M \cdot z_1 - W_{1,1}^M) \\ 
% &\phantom{=} \hphantom{\frac{1}{N}\sum_{n}^{N}[y_n-(} \;\; 
& \hspace{2cm} (1+W_{2,1}^M \cdot z_2 - W_{2,1}^M))]^2
\end{aligned}
\end{equation*}

\subsection{Explicit Gradients}
\begin{figure*}[t]
\vskip 0.2in
\centering
\includegraphics[width=0.8\textwidth]{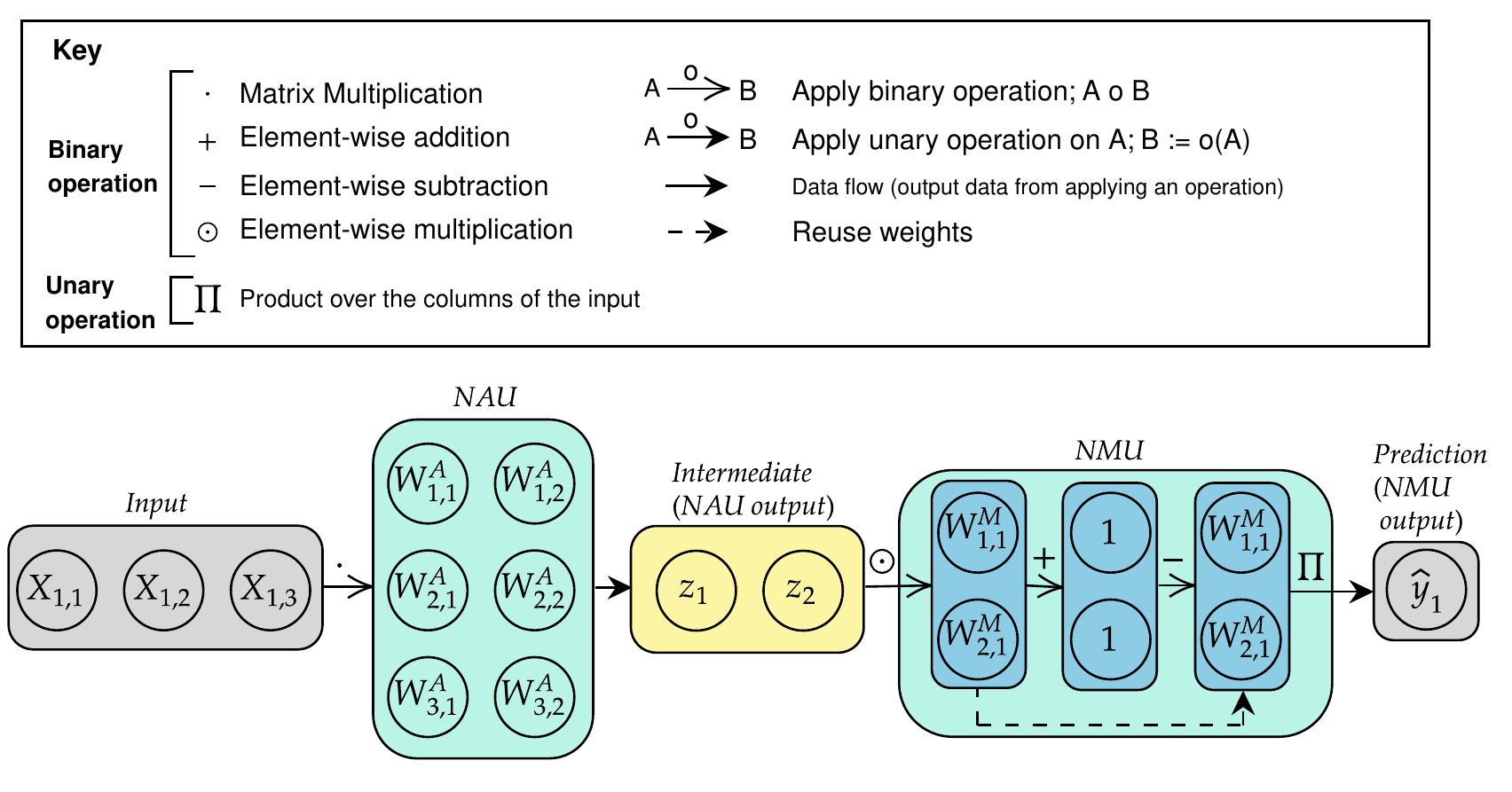}
\caption{Illustration of the data flow of a NAU-NMU module. The annotation of weights will be consistent with the gradient calculations.}
\label{fig:NAU-NMU-illustration}
% \vskip -0.2in
\end{figure*}

To help improve familiarity with notation, we first work through an example using predefined network sizes. 
For the following (simplified) two module example using the baseline stacked NAU-NMU, we assume an input vector size 3, intermediate size 2, and output size 1. As a further simplification, we only consider the loss for a single data-label pair ($\bm{X_1}$, $y_1$).
We annotate the weights and components explicitly in Figure~\ref{fig:NAU-NMU-illustration}. Matrix indexing follows the standard (row, column) convention, with indexing starting from 1. 

\subsubsection{MSE Loss Partial Derivatives:}
Derivation~\ref{der:mse-2LT} calculates the loss derivative with respect to (wrt) three different weights values: $W^A_{1,1}$, $W^A_{1,2}$, and $W^M_{1,1}$ indicated by colours (yellow, purple, and teal). 
Colours red and blue are used to identify the parts of the derivative which are derived from the original equation (i.e. the predicted value $\hat{y_1}$). 
Using the chain rule on the loss requires calculating the partial derivative of the predicted value wrt the weight, which is calculated via the product rule using the underline red and blue terms as the parts. To differentiate each part further, also requires another application of the product rule.   

\begin{derivation*}[!ht]
\begin{equation*}
\begin{aligned}
L &= (y_1 - \hat{y_1})^2\\
\hat{y_1} &= \Cline[blue]{
            (\textcolor{orange}{W_{1,1}^M} \cdot (X_{1,1}\cdot \textcolor{violet}{W^A_{1,1}} + X_{1,2}\cdot W_{2,1}^{A} + X_{1,3}\cdot W_{3,1}^{A}) + 1 - \textcolor{orange}{W_{1,1}^M})
        } \cdot \\
       &\Cline[red]{
            (W_{2,1}^M \cdot (X_{1,1}\cdot \textcolor{teal}{W_{1,2}^{A}} + X_{1,2}\cdot W_{2,2}^{A} + X_{1,3}\cdot W_{3,2}^{A}) + 1 -  W_{2,1}^M)
        }\\
\frac{\partial L}{\partial W^A_{1,1}} &= -2 \cdot\frac{\partial \hat{y_1}}{\partial W^A_{1,1}} \cdot (y_1-\hat{y_1})\\
\frac{\partial \hat{y_1}}{\partial \textcolor{violet}{W^A_{1,1}}} &= 
        \textcolor{blue}{
            W_{1,1}^M\cdot X_{1,1}
        }\cdot 
        \textcolor{red}{
            (W_{2,1}^M \cdot (X_{1,1}\cdot W_{1,2}^{A} + X_{1,2}\cdot W_{2,2}^{A} + X_{1,3}\cdot W_{3,2}^{A}) + 1 - W_{2,1}^M)
        }\\
        &=  \textcolor{blue}{
            W_{1,1}^M\cdot X_{1,1}
        }\cdot 
        \textcolor{red}{
            (W_{2,1}^M \cdot (\sum_{i}^{I}X_{1,i} \cdot W^A_{i,2}) + 1- W_{2,1}^M)
        }\\
\frac{\partial L}{\partial W^A_{1,2}} &= -2 \cdot \frac{\partial \hat{y_1}}{\partial W^A_{1,2}} \cdot (y_1-\hat{y_1})\\
\frac{\partial \hat{y_1}}{\partial \textcolor{teal}{W^A_{1,2}}} &= 
        \textcolor{blue}{
            (W_{1,1}^M \cdot (X_{1,1}\cdot W^A_{1,1} + X_{1,2}\cdot W_{2,1}^{A} + X_{1,3}\cdot W_{3,1}^{A}) +1 - W_{1,1}^M)
        } \cdot
        \textcolor{red}{
            W_{2,1}^M\cdot X_{1,2}
        }\\
        &= \textcolor{blue}{
            (W_{1,1}^M \cdot (\sum_{i}^{I}X_{1,i} \cdot W^A_{i,1}) + 1 - W_{1,1}^M)
        } \cdot
        \textcolor{red}{
            W_{2,1}^M\cdot X_{1,2}
        }
        \\
\frac{\partial L}{\partial W^M_{1,1}} &= -2 \cdot \frac{\partial \hat{y_1}}{\partial W^M_{1,1}}\cdot (y_1-\hat{y_1})\\
\frac{\partial \hat{y_1}}{\partial \textcolor{orange}{W^M_{1,1}}} &= 
        \textcolor{blue}{
            ((X_{1,1}\cdot W^A_{1,1} + X_{1,2}\cdot W_{2,1}^{A} + X_{1,3}\cdot W_{3,1}^{A}) - 1)
        }\cdot \\
        &\textcolor{red}{
            (W_{2,1}^M \cdot (X_{1,1}\cdot W_{1,2}^{A} + X_{1,2}\cdot W_{2,2}^{A} + X_{1,3}\cdot W_{3,2}^{A}) + 1 - W_{2,1}^M)
        }\\
        &= \textcolor{blue}{
            (\sum_{i}^{I}(X_{1,i}\cdot W^A_{i,1}) - 1)
        }\cdot \textcolor{red}{
            (W_{2,1}^M \cdot (\sum_{i}^{I}(X_{1,i}\cdot W^A_{i,2}) + 1 - W_{2,1}^M)
        }
\end{aligned}
\end{equation*}
\caption{Partial derivatives on the MSE Loss of the NAU-NMU wrt weight elements $W^A_{1,1}$, $W^A_{1,2}$, and $W^M_{1,1}$.}
\label{der:mse-2LT}
\end{derivation*}

\textbf{The partial derivative of the prediction wrt to a NAU weight is the product of two terms:} 
One term is the NMU weight (whose row index matches the column index of the target NAU weight) multiplied with the input (whose column index corresponds to the target NAU weight's row index). 
The other term is the result of what would be the output of the NMU if it is only applied to intermediate $z_i$ where $i$ is the value which is not the value of the column of the target NAU weight. 
E.g. $W^A_{1,2}$ considers $z_1$. 

\textbf{The partial derivative of the prediction wrt to a NMU weight is the product of two terms:} 
One term is the intermediate element which corresponds to the row value of the target NMU weight minus 1, e.g. $W^M_{1,1}$ would have $z_1-1$. 
The other term is the result of what would be the output of the NMU if only applied to the intermediate $z_i$ where $i$ is the value which is not the value of the column of the target NAU weight. 
E.g. $W^M_{1,1}$ considers $z_2$. 
This term also occurs in the partial derivative when target weight being derived to is a NAU weight.

\subsection{Generalised NAU and NMU Partial Derivatives of the loss for a NAU-NMU}
The derivative of the loss wrt either a NAU or NMU weight can the be derived using the chain rule. 
We formulate these gradients for the generalised case. 
The expression is generalised such that it can be applied to any element in the NAU weight matrix regardless of the matrix's size, and the NMU weight matrix regardless of the matrix's row size.
Like before, we assume derivatives for a single data-label pair ($\bm{X_1}$, $y_1$). 

To reiterate, the NAU weight matrix is denoted as $W_{l,i}^A$ where the A represents a summative module (for adding/subtracting), $l$ is the output element index for the output applying the NAU, and $i$ is the index to select an element from the input. 

\begin{equation*}
\begin{aligned}
\frac{\partial L}{\partial W^A_{i,l}} = 
-2(y_1 - \hat{y_1}) \cdot W^M_{l,1}X_{1,i} \\\cdot \prod_{j}^{\{O\backslash l\}}(W^M_{j,1}(\sum_{k=1}^{I}{X_{1,k}W^A_{k,j}}) + 1 - W^M_{j,1})
\end{aligned}
\end{equation*}
\begin{equation*}
\begin{aligned}
\frac{\partial L}{\partial W^M_{l,1}} = -2(y_1 - \hat{y_1}) \cdot (\sum_{i=1}^{I}X_{1,i}W_{i,l}^{A}-1) \\\cdot \prod_{j}^{\{O\backslash l\}}(W_{j,1}^{M}(\sum_{k=1}^{I}X_{1,k}W_{k,j}^{A})+1-W_{j,1}^{M})
\end{aligned}
\end{equation*}

$\{O\backslash l\}$ represents the indices of all output elements from applying the module excluding the index corresponding to the output for the weight element you are calculating the partial derivative of.  

\subsection{Generalised NAU and NMU Partial Derivatives for a NAU-sNMU}
We derive the generalised gradients as before but now using a sNMU rather than a NMU. Gradients are derived using the quotient rule.  
Let $\bm{N}$ be the noise matrix (same shape as input $\bm{X}$). 

\subsubsection{MSE Loss Definition}
\begin{equation*}
\begin{aligned}
L &= (y_1-\hat{y_1})^2\\
  &= (y_1-\text{sNMU}(\text{NAU}(\bm{X_{1}})))^2\\
\end{aligned}
\end{equation*}

\subsubsection{Loss derivatives wrt NAU and sNMU weights}
\begin{align*}
\shortintertext{Let}
A &=\prod_{j}^{\{O\backslash l\}}N_{1,j}W_{j,1}^{M}\cdot (\sum_{k=1}^{I}X_{1,k}W_{k,j}^{A})+1-W_{j,1}^{M}, \\
\intertext{be the result of the sNMU applied only to the output values of the NAU whose index is not the value of the column of the target NAU weight. Let }
D &= \prod_{i}^{O} N_{1,i}W^M_{i,1} + 1 - W^M_{i,1}, \\
\shortintertext{be the denoising term. Therefore, }
\frac{\partial L}{\partial W^A_{i,l}} &= 
    -2(y_1 - \hat{y_1}) \cdot \frac{A}{D} \cdot W_{i,1}^{M}N_{1,i}X_{1,i}\\
\frac{\partial L}{\partial W^M_{i,1}} &= 
    -2(y_1-\hat{y_1})\cdot \frac{A}{D^2} \cdot [D(N_{1,i}z_{i}-1)
    \\&-(W_{i,1}^{M}N_{1,i}z_{i} +1 - W^M_{i,1})(N_{1,i}-1)
    \\&(\prod_{j}^{\{O\backslash l\}}(N_{1,j}W^M_{j,1}+1-W^M_{j,1}))]
\end{align*}

Although the residual term $(y_1-\hat{y_1})$ remains, the sNMU will scale the gradients by some noise factor $\geqslant 1$ which magnifies the magnitudes of the gradients. 
Hence, the sNMU’s noise amplifies its gradients helping to alleviate the gradient dampening caused by the residual term. 

\section{Static MNIST Product}\label{app:static-mnist-product}
This section details the architectures used and further explores the learnt models from the static MNIST tasks. 

\subsection{Architecture Details.}
\begin{figure}[h]
    \centering
    \subfloat[]{{\includegraphics[width=0.58\textwidth]{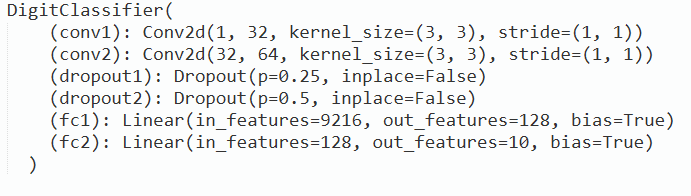} }}
    % \qquad
    \subfloat[]{{\includegraphics[width=0.4\textwidth]{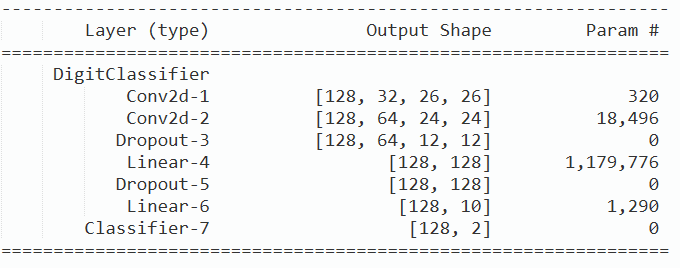}}}
    \caption{Digit classification network structure and summary used in the Isolated Digits MNIST task}
    \label{fig:isolated-digitClf}
\end{figure}

\textbf{Isolated Digits} 
The digit classification network can be found in Figure~\ref{fig:isolated-digitClf}. The output of the digit classifier is passed to a multiplication module which returns the final output predictions. 

\textbf{Colour Channel Concatenated Digits.} 
We use the rotated, translated, and scaled (RTS) dataset described in~\citet[Appendix A.4]{spatialTransformer2015}. 
The RTS dataset is generated by randomly rotating an MNIST digit by +45 and -45 degrees, randomly scaling the digit by a factor of between 0.7 and 1.2, and placing the digit in a random location in a 42×42 image. 

\begin{figure}[h]
    \centering
    \subfloat[]{{\includegraphics[width=0.58\textwidth]{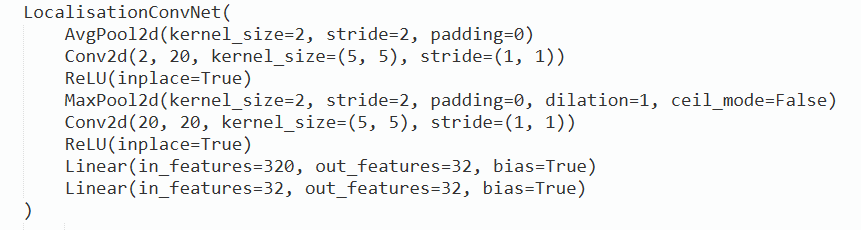} }}
    % \qquad
    \subfloat[]{{\includegraphics[width=0.4\textwidth]{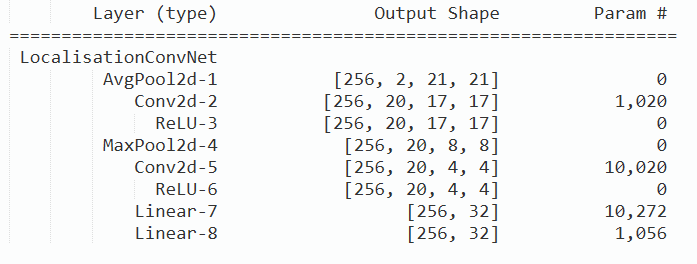}}}
    \caption{Localisation network structure and summary used in the Colour Channel Concatenated Digits MNIST task}
    \label{fig:concat-locNet}
\end{figure}

\begin{figure}[h]
    \centering
    \subfloat[]{{\includegraphics[width=0.58\textwidth]{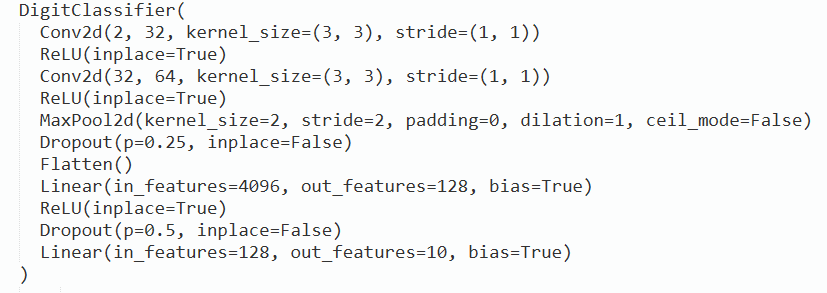} }}
    % \qquad
    \subfloat[]{{\includegraphics[width=0.4\textwidth]{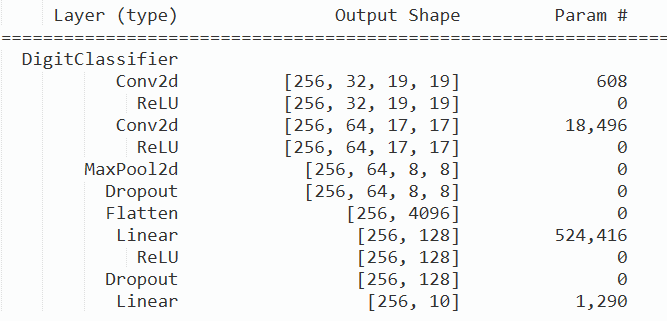}}}
    \caption{Digit classification network structure and summary used in the Colour Channel Concatenated Digits MNIST task}
    \label{fig:concat-digitClf}
\end{figure}

Given an image which is distorted via random scaling, rotation and translation, the spatial transformer network can learn to locate the digit of interest and transform the source image to produce a version of the digit more like its non-distorted form. 
First a localisation network learns a set of K control coordinates which are normalised between [-1,1]. 
These control points learn a grid around the point of interest which ub this case is the digit. 
As there are two digits, two localisations are learnt. 
A localisation network consists of a convolutional network (see Figure~\ref{fig:concat-locNet}) with a tanh transformation at the end for normalisation to [-1,1]. 
The Thin Plate Spline (TPS) transformation parameters are calculated using the control points from a localisation network. 
The TPS transformation will transform the target image's pixel coordinates to the source image's pixel coordinates. 
To generate the TPS transformation matrix, we follow~\citet[Section 3.1.2]{shi2016robustTPS}. 
Finally, a sampling grid will take the source image and its pixel locations to produce the transformed image. 
The transformed image is then passed to a classification network (see Figure~\ref{fig:concat-digitClf}) to produce logits for digit classification. 
The classified digits are then passed to the relevant multiplication network. 

\newpage
\subsection{Multiplication Weight Trajectories}
This section plots the trajectories for the two learnable weights used when calculating the multiplication operation for each fold. 
The baseline, which uses a solved multiplier, will not have any learnable weights for multiplication hence has no trajectory plot. 
The remaining models i.e., FC, NMU, sNMU do have multiplication weights to learn and for all cases the true solution of the weights is $[1, 1]$. 

\begin{figure*}[h]
    \vskip 0.1in
    \centering
    \subfloat[\centering FC]{{\includegraphics[ width=0.24\textwidth]{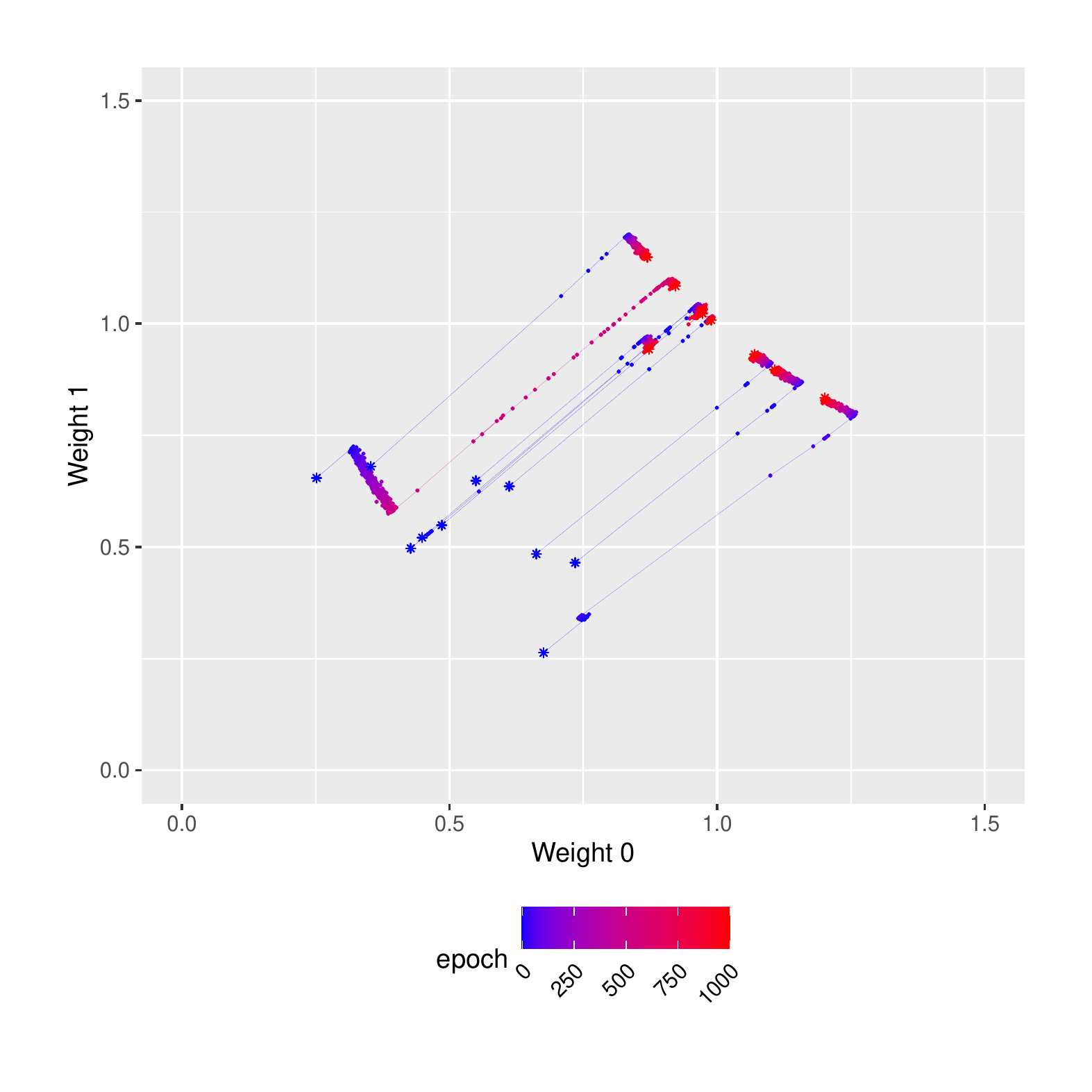}}}
    % \qquad
    \subfloat[\centering NMU]{{\includegraphics[ width=0.24\textwidth]{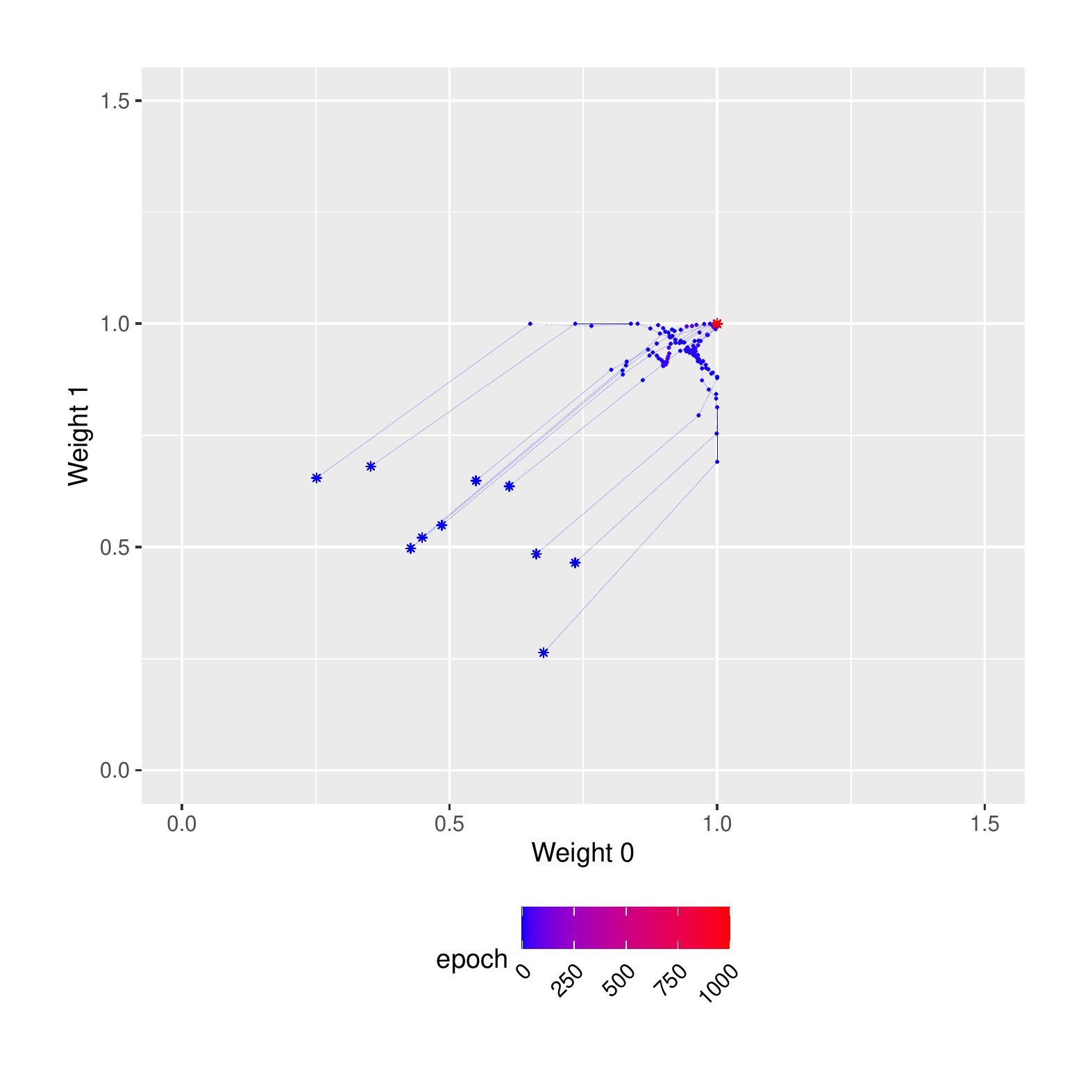}}}
    % \qquad
    \subfloat[\centering {sNMU $\mathcal{U}$[1,5]}]{{\includegraphics[ width=0.24\textwidth]{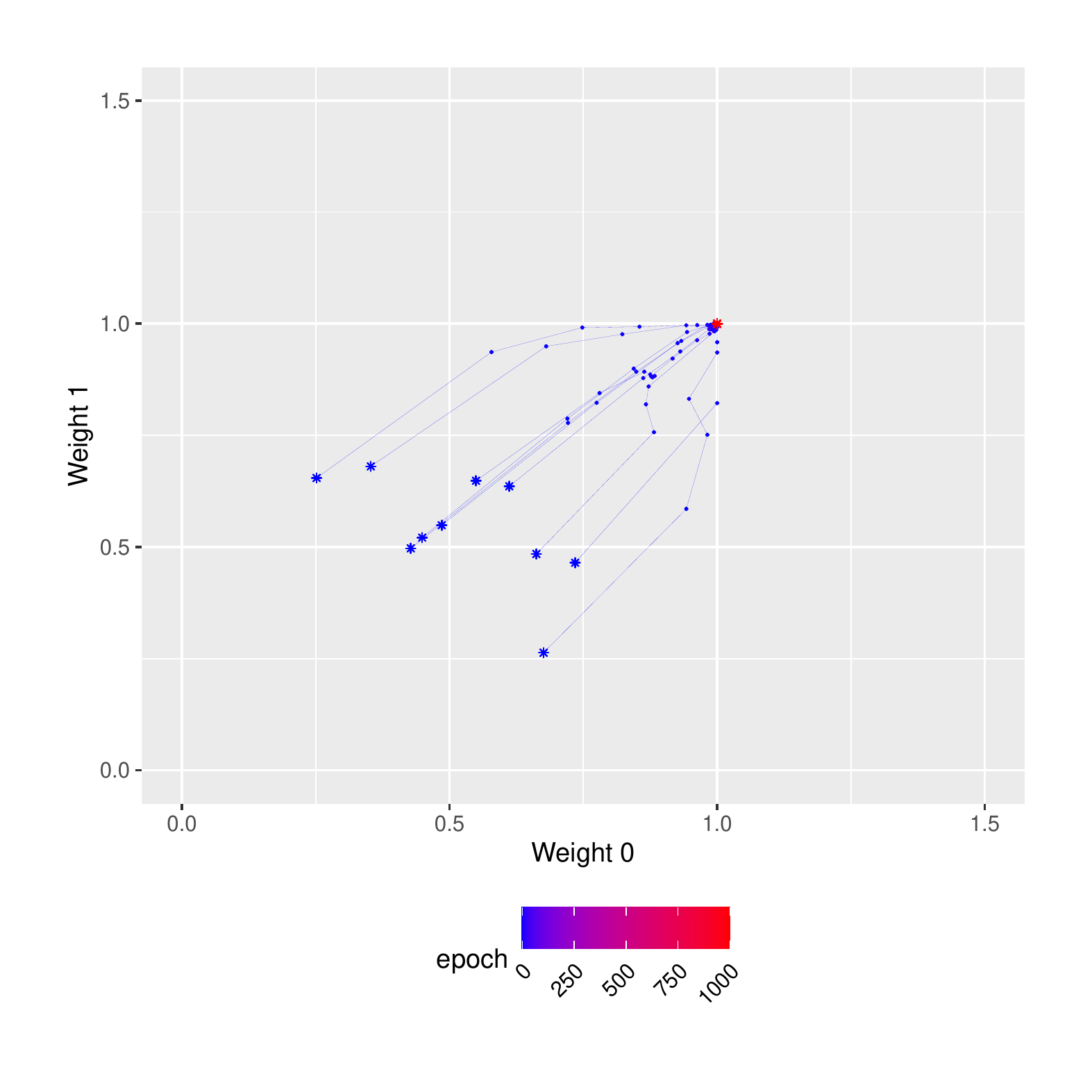} }}
    % \qquad
    \subfloat[\centering {sNMU $\mathcal{U}$[1,1+1/sd(x)]}]{{\includegraphics[ width=0.24\textwidth]{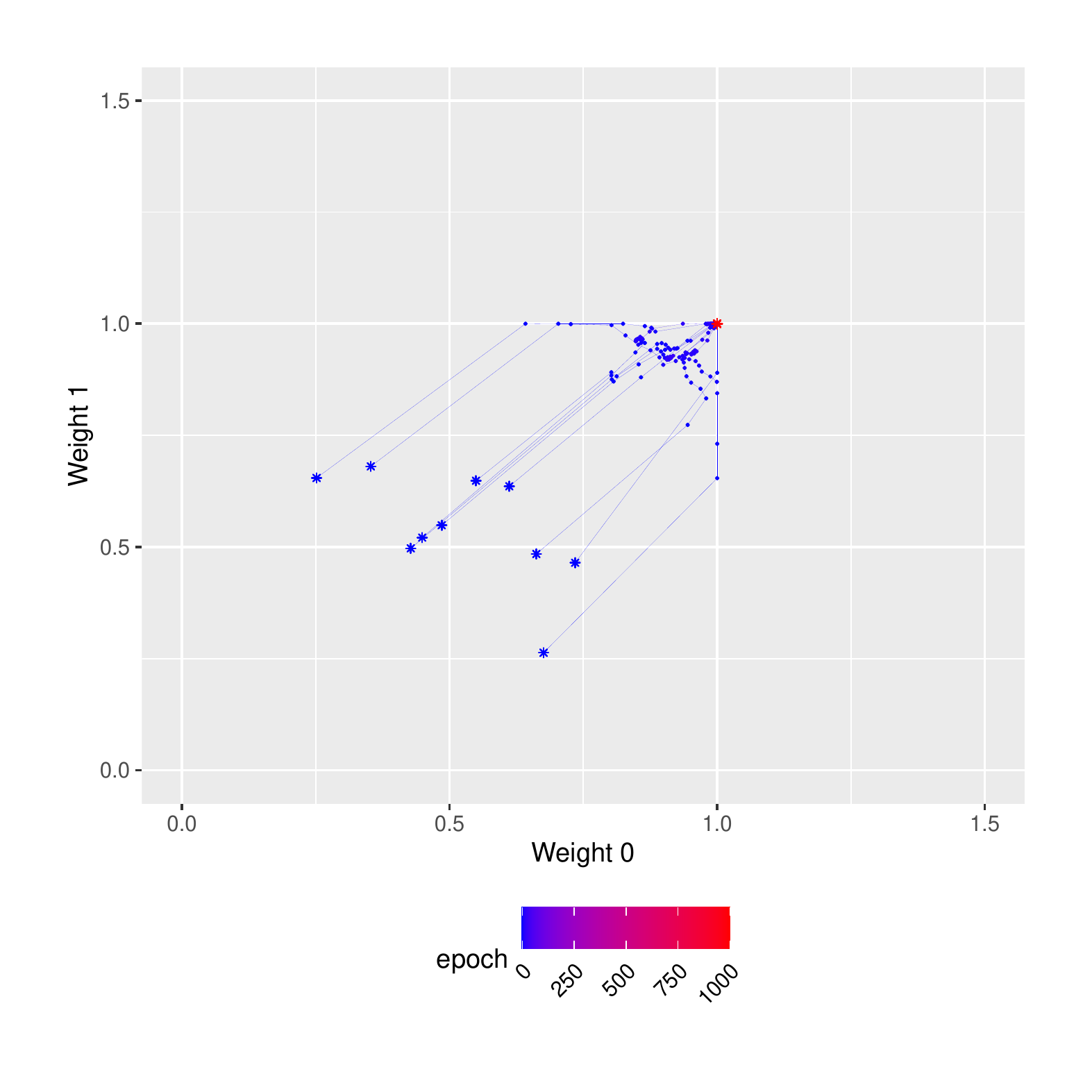} }}
    \caption{Path learnt by the weights for the multiplication network. Each path represents a different seed. Blue and red asterisks represent the starting and ending points respectively.}
    \label{fig:1digit-mnist-weights-path}
    \vskip -0.1in
\end{figure*}

\textbf{Isolated digits.} Figure~\ref{fig:1digit-mnist-weights-path}. 
The FC models are unable to reliably converge to the true solution on any run. 
The NMU gets close to the solution but only 70\% of runs converge to weights of 1 exactly, while 100\% of the sNMU models converge. 

\begin{figure*}[h]
    \centering
    \subfloat[\centering FC]{{\includegraphics[ width=0.24\textwidth]{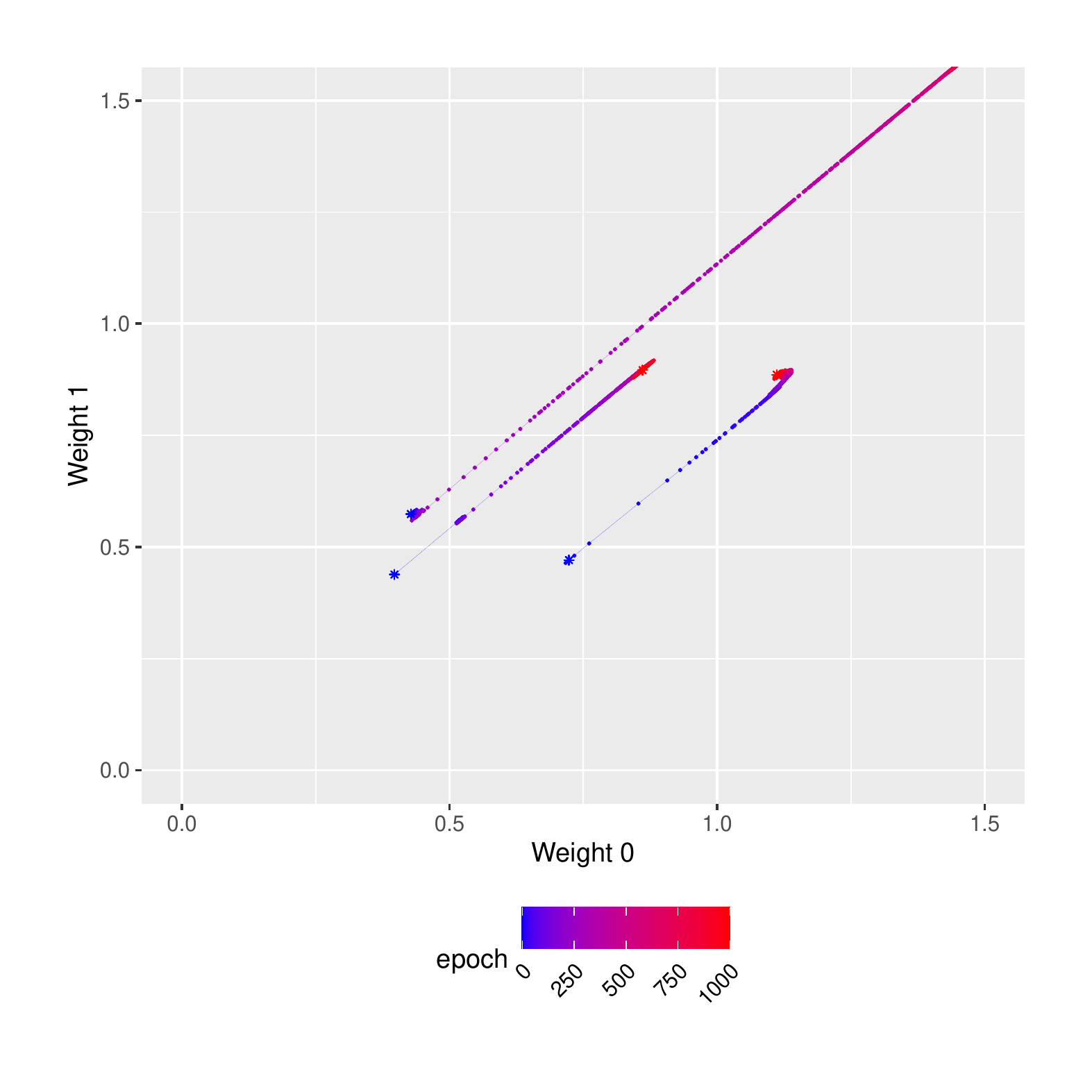}}}
    % \qquad
    \subfloat[\centering NMU]{{\includegraphics[ width=0.24\textwidth]{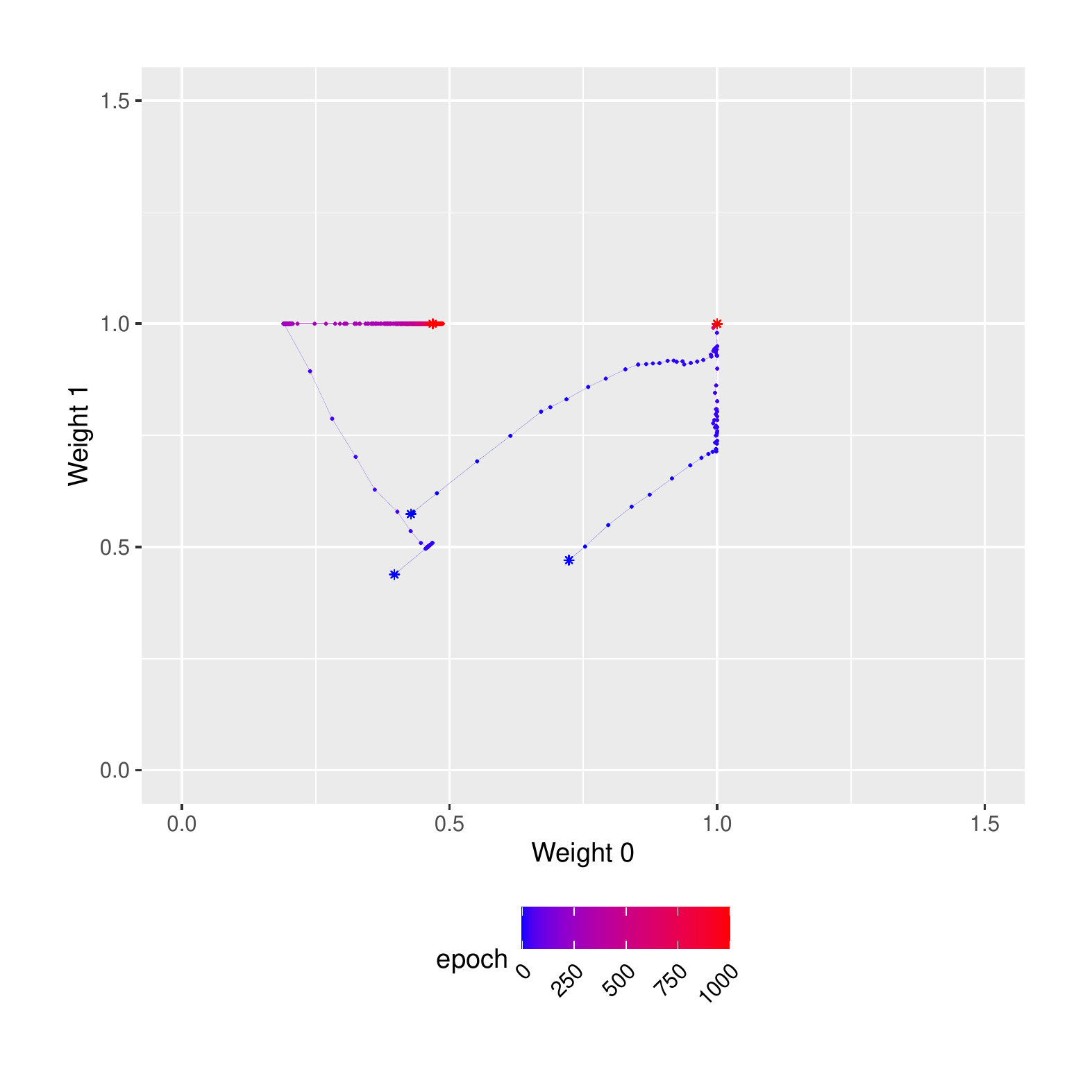}}}
    % \qquad
    \subfloat[\centering {sNMU $\mathcal{U}$[1,5]}]{{\includegraphics[ width=0.24\textwidth]{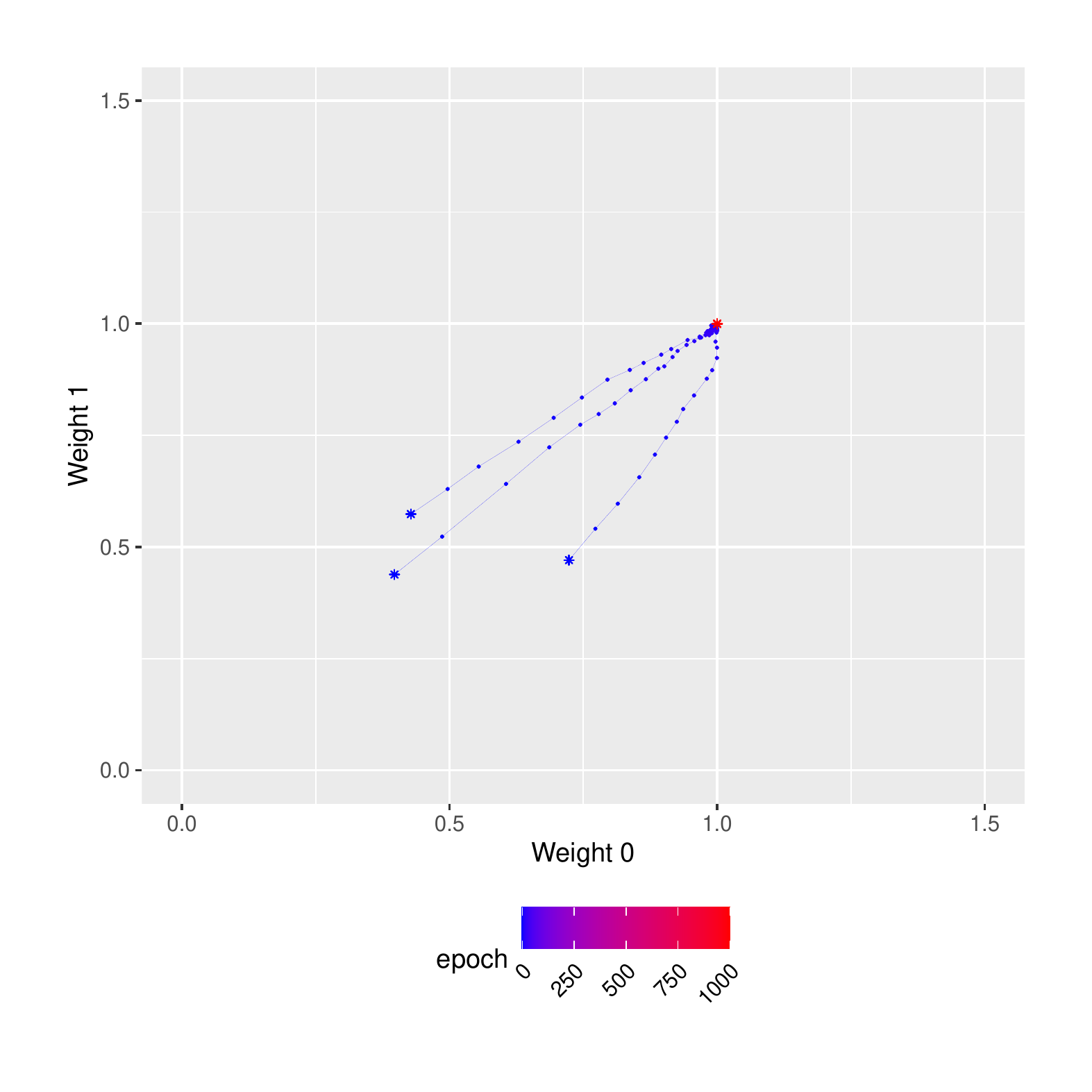} }}
    % \qquad
    \subfloat[\centering {sNMU $\mathcal{U}$[1,1+1/sd(x)]}]{{\includegraphics[ width=0.24\textwidth]{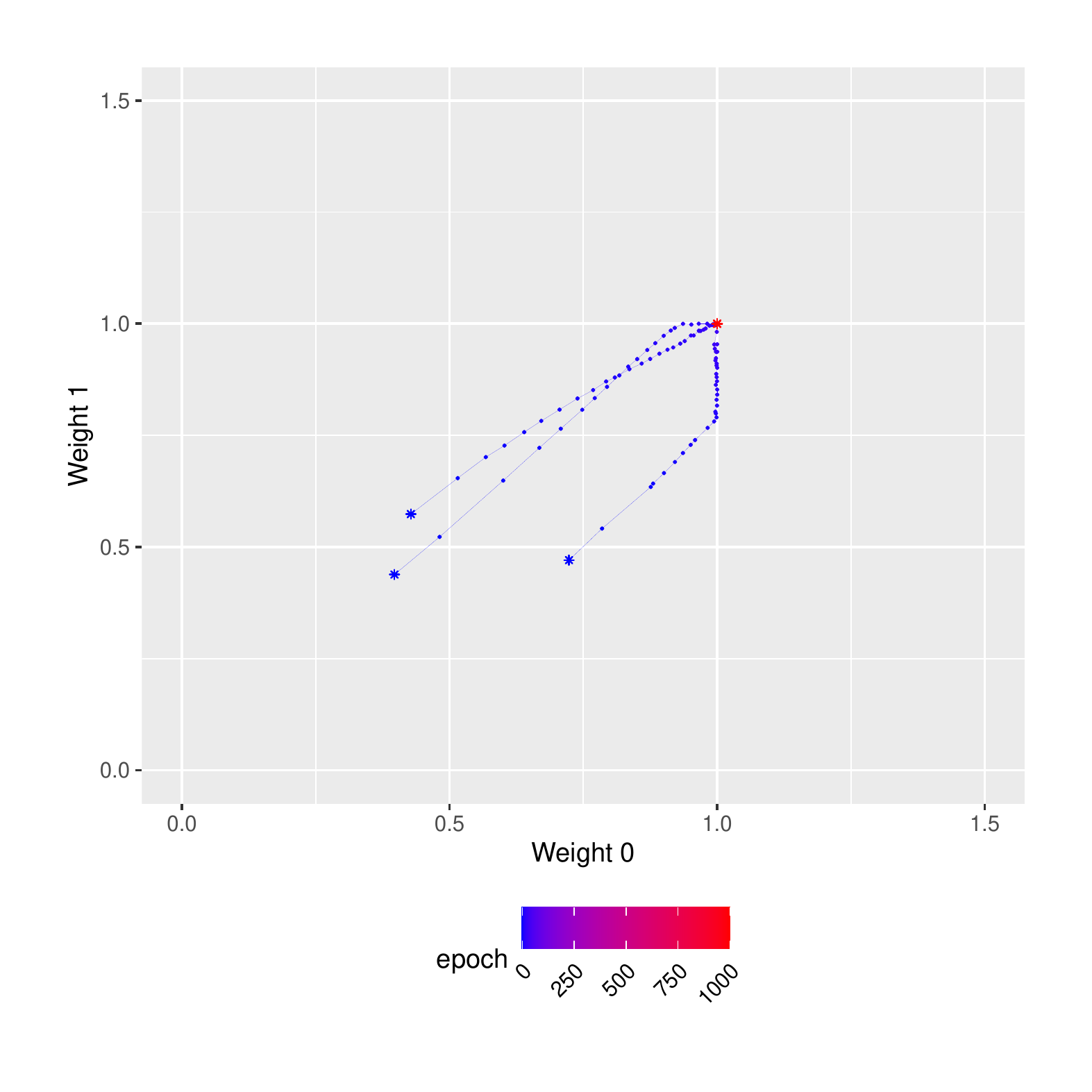} }}
    \caption{Path learnt by the weights for the multiplication network. Each path represents a different seed. Blue and red asterisks represent the starting and ending points respectively.}
    \label{fig:mnist-st-tps-weights-path}
\end{figure*}

\textbf{Colour Channel Concatenated Digits.} Figure~\ref{fig:mnist-st-tps-weights-path}. When the task difficulty of classifying the digits is increased, the NMU is also unable to converge, while the sNMUs still can converge for all folds. 
The FC models are again unable to reliably converge to the true solution.
% snmu [1,5] 1 seed nearly there, snmu 1 completely fails

\subsection{Class Accuracies}
This section plots the class accuracies of the models for a fold, evaluated on the test dataset. 
Doing so helps assess the learnt representations of the digit classifier network. 
The accuracy for classifying each digit over the test set is plotted with a further breakdown of the decisions over each digit using a unnormalised confusion matrix. 
The distribution of digit labels will be non-uniform. 

\begin{figure*}[h!]
\vskip 0.1in
\centering
\includegraphics[width=\textwidth]{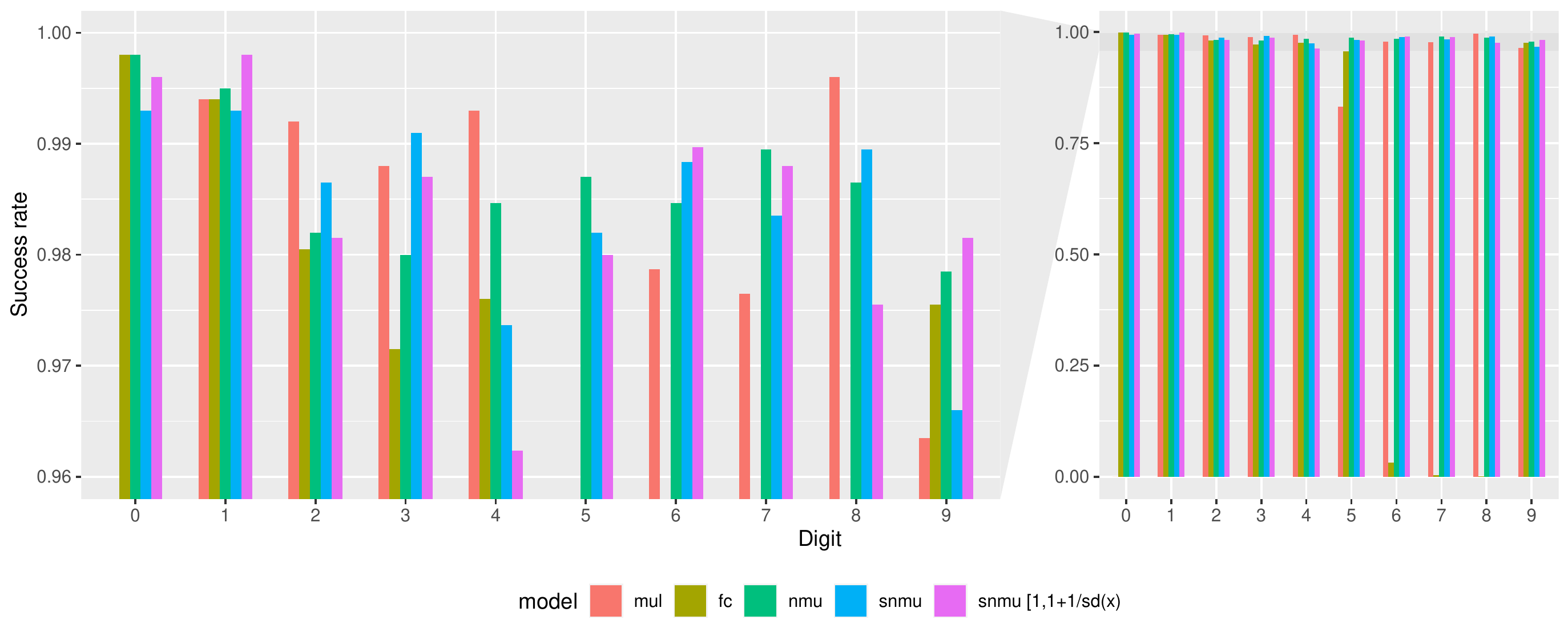}
\caption{Success rates for classifying each digit in the test dataset for a single seed. (Left) Zoom-in for success in range 0.95-1. (Right) Full plot from success rate 0-1.}
\label{fig:mnist-indep-digit-class-accs}
\vskip -0.1in
\end{figure*}

\begin{figure}[h!]
    \vskip 0.1in
    \centering 
\begin{subfigure}{0.32\textwidth}
  \includegraphics[trim=0.5cm 1cm 7cm 1cm, width=\linewidth]{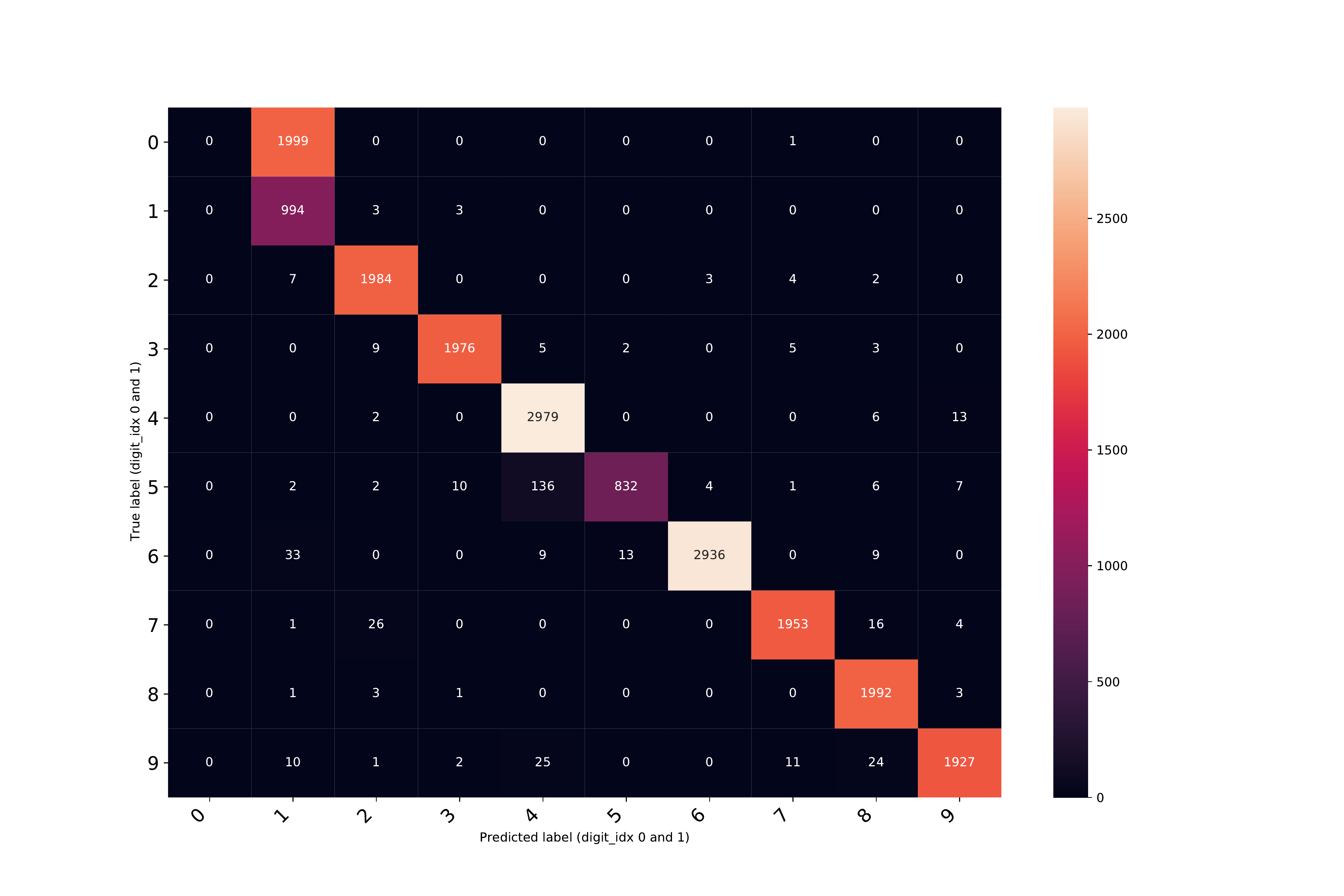}
  \caption{Baseline (MUL)}
\end{subfigure}%\hfil 
\begin{subfigure}{0.32\textwidth}
  \includegraphics[trim=0.5cm 1cm 7cm 1cm, width=\linewidth]{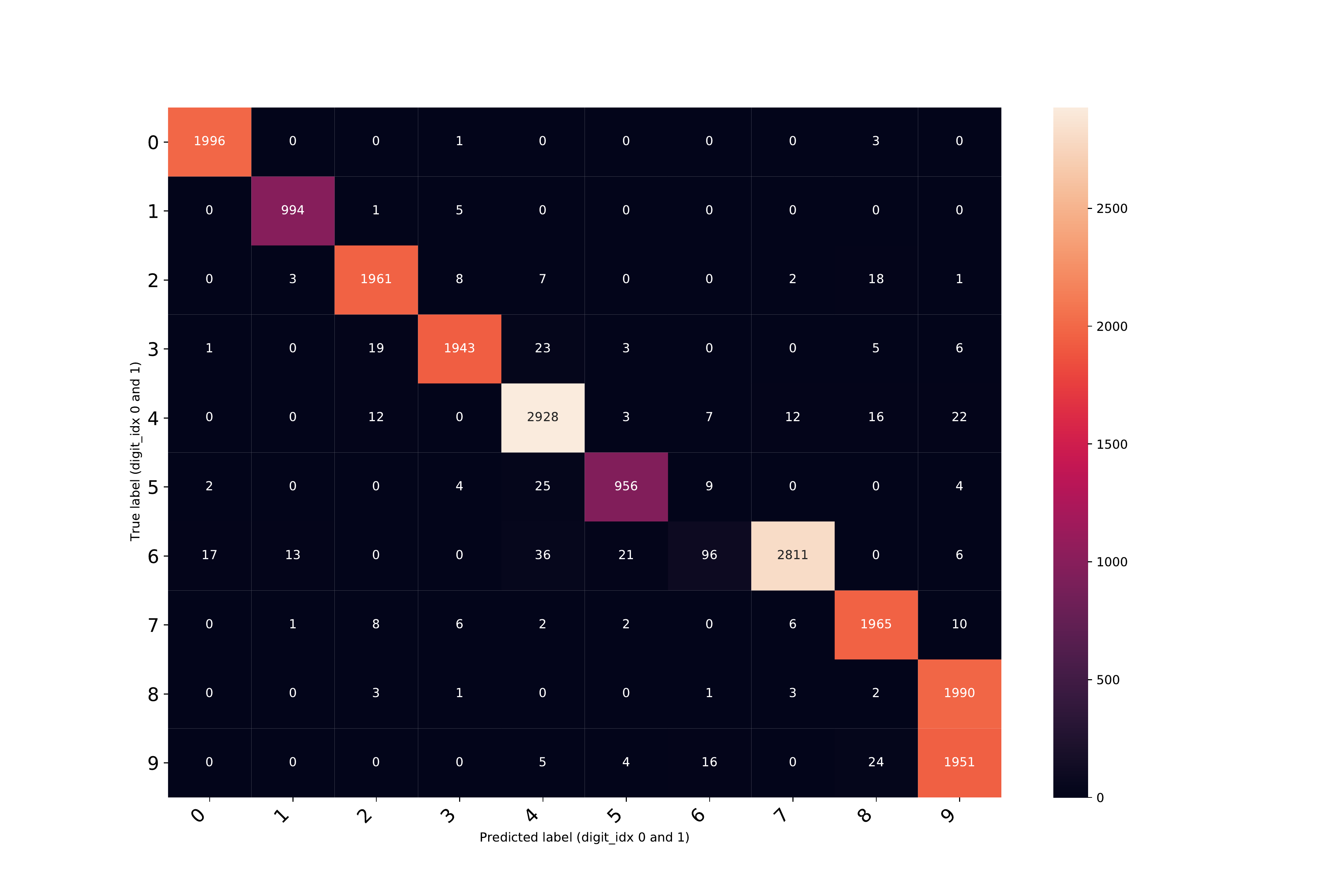}
  \caption{FC}
\end{subfigure}%\hfil
\begin{subfigure}{0.32\textwidth}
  \includegraphics[trim=0.5cm 1cm 7cm 1cm, width=\linewidth]{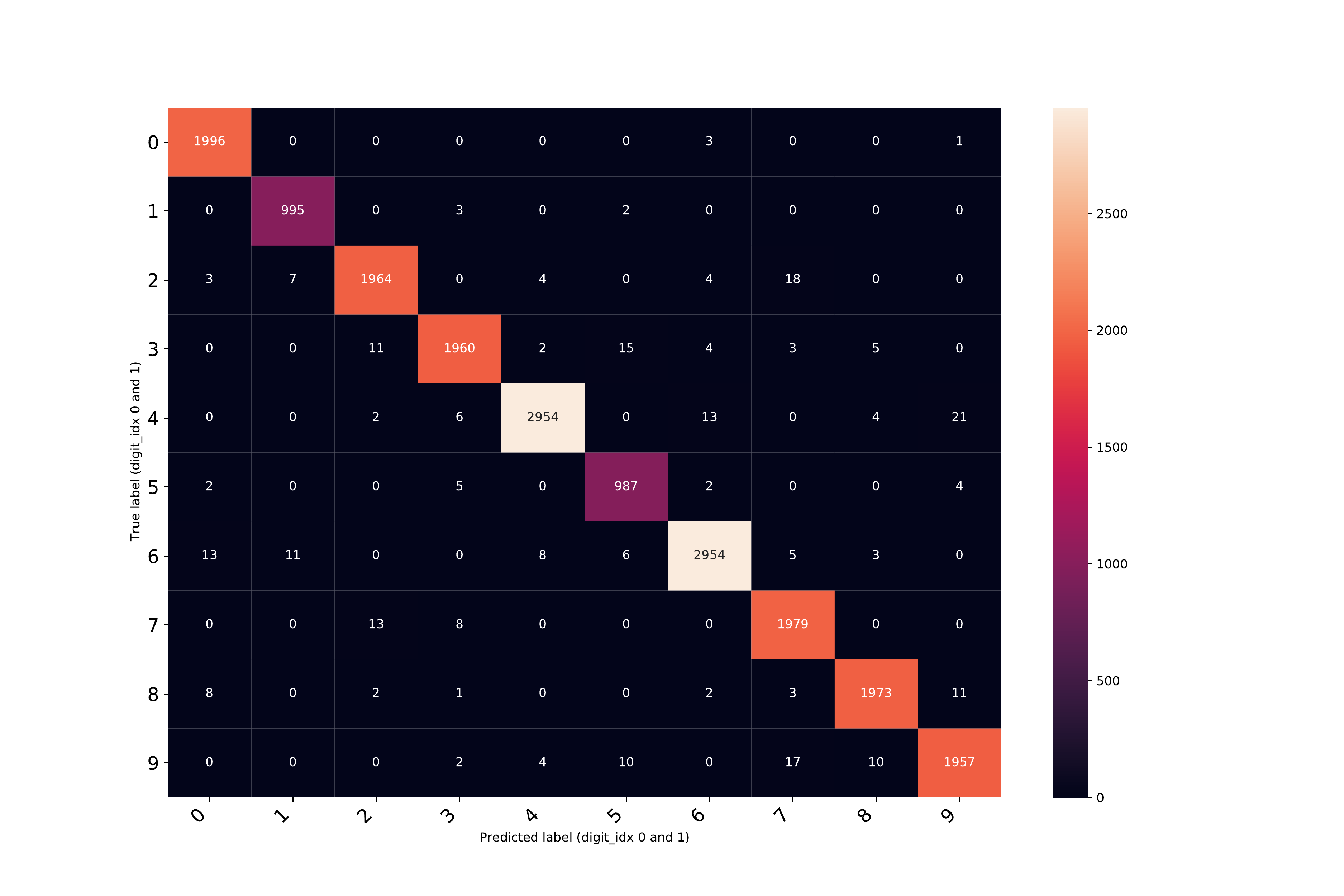}
  \caption{NMU}
\end{subfigure}

\medskip
\begin{subfigure}{0.32\textwidth}
  \includegraphics[trim=0.5cm 1cm 7cm 1cm, width=\linewidth]{imgs/mnist/two-digit/1digit_conv-Adam/confusion-matricies/24_f2_op-mul_nalmT_learnLT_s2_confusion_matrix.pdf}
  \caption{sNMU $\mathcal{U}$[1,5]}
\end{subfigure}%\hfil 
\begin{subfigure}{0.32\textwidth}
  \includegraphics[trim=0.5cm 1cm 7cm 1cm, width=\linewidth]{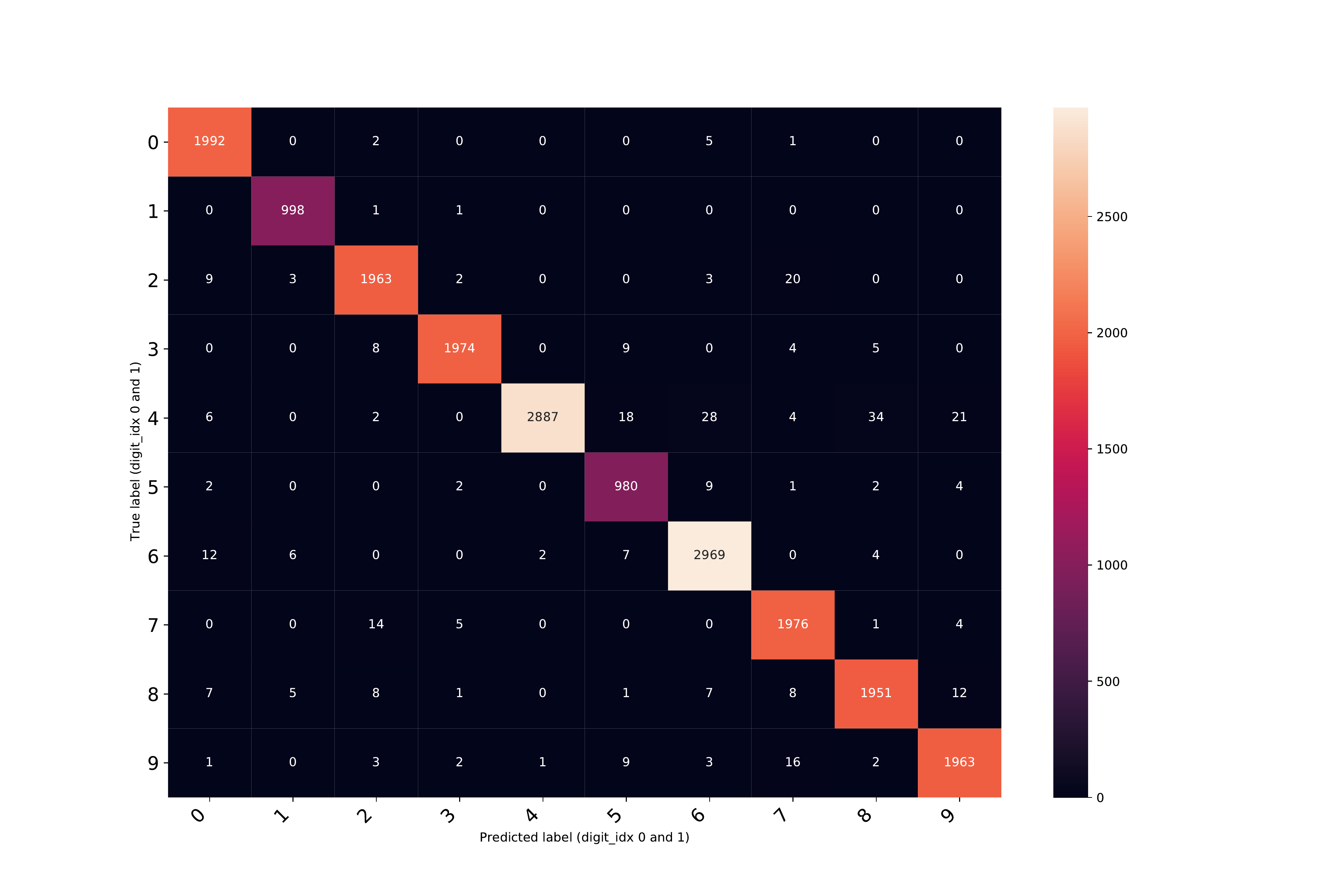}
  \caption{sNMU $\mathcal{U}$[1,1+1/sd(x)]}
\end{subfigure}%\hfil
\caption{Confusion matrices for intermediate label classification}
\label{fig:1digit-mnist-label-confusion-matricies}
\vskip -0.1in
\end{figure}

\textbf{Isolated digits.} 
Figures~\ref{fig:mnist-indep-digit-class-accs} and \ref{fig:1digit-mnist-label-confusion-matricies}. 
The baseline is unable to classify zeros, mistaking all occurrences except 1 as the number 1. 
The FC model completely misclassifies digits 6,7 and 8 as 7,8 and 9 respectively. 
Both NMU and sNMU variants have strong classifiers with each digit getting at least 96\% success in classification. 

\begin{figure*}[h!]
\vskip 0.1in
\centering
\includegraphics[width=\textwidth]{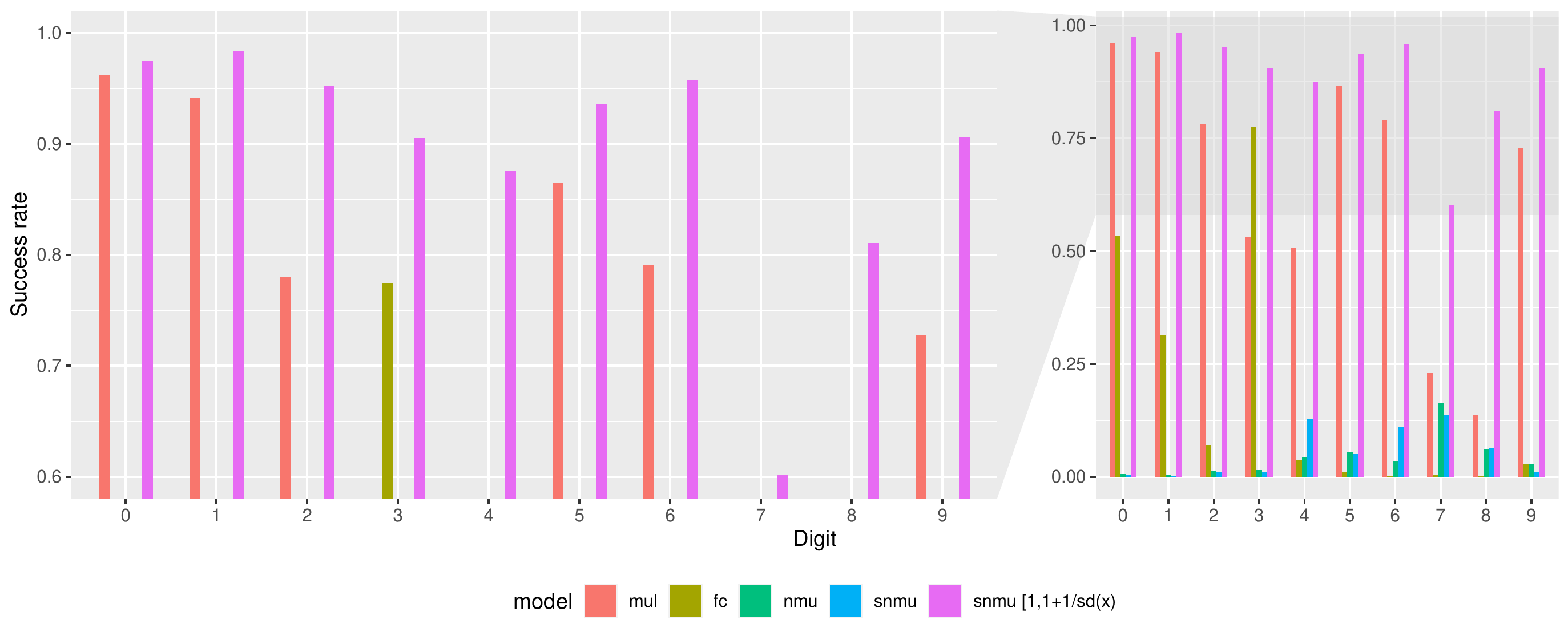}
\caption{Success rates for classifying each digit (with rounding) in the test dataset for a single seed. (Left) Zoom-in for success in range 0.95-1. (Right) Full plot from success rate 0-1.}
\label{fig:mnist-st-tps-label-class-accs}
\vskip -0.1in
\end{figure*}

\begin{figure}[h!]
    \vskip 0.1in
    \centering 
\begin{subfigure}{0.32\textwidth}
  \includegraphics[trim=0.5cm 1cm 7cm 1cm, width=\linewidth]{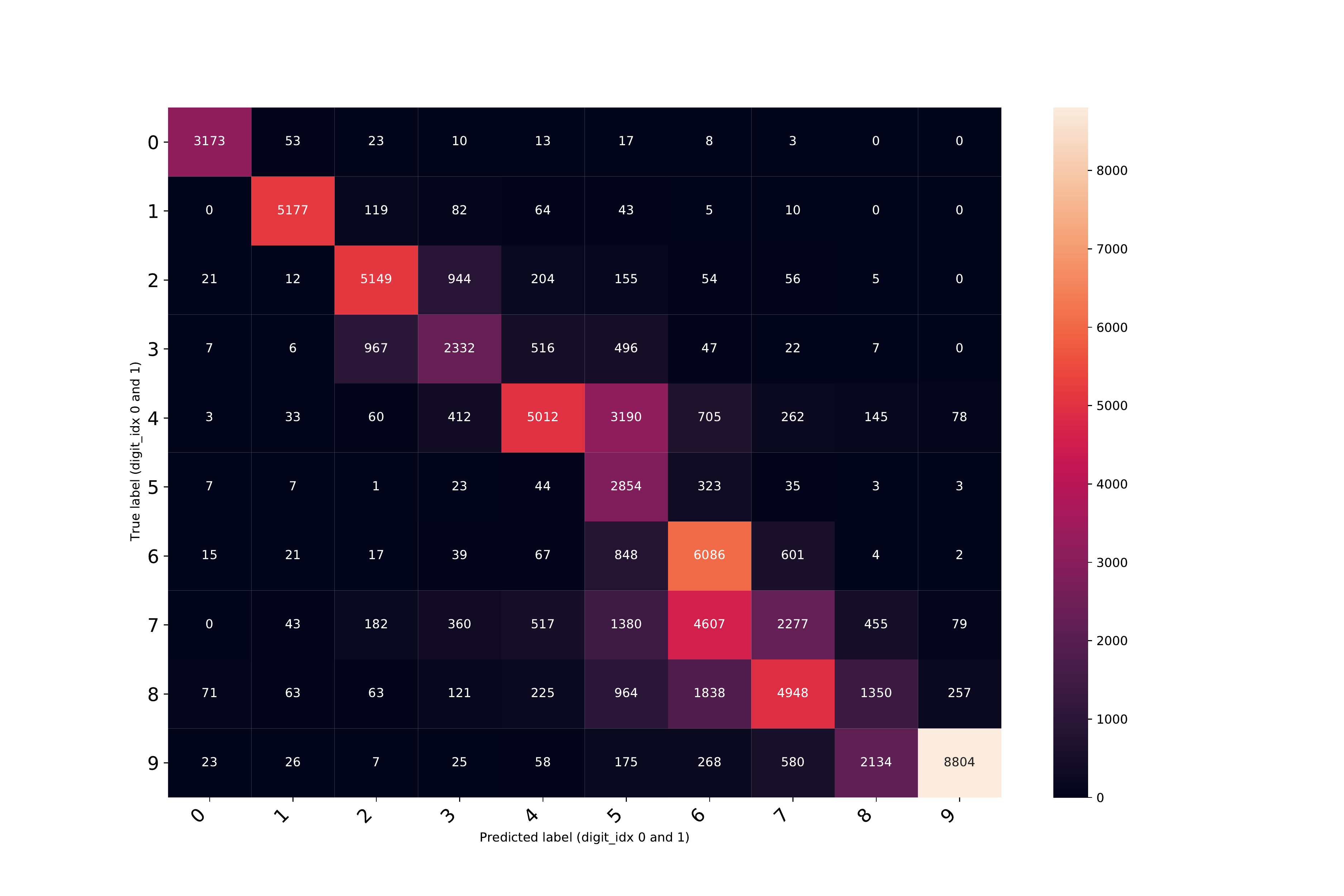}
  \caption{Baseline (MUL)}
\end{subfigure}%\hfil 
\begin{subfigure}{0.32\textwidth}
  \includegraphics[trim=0.5cm 1cm 7cm 1cm, width=\linewidth]{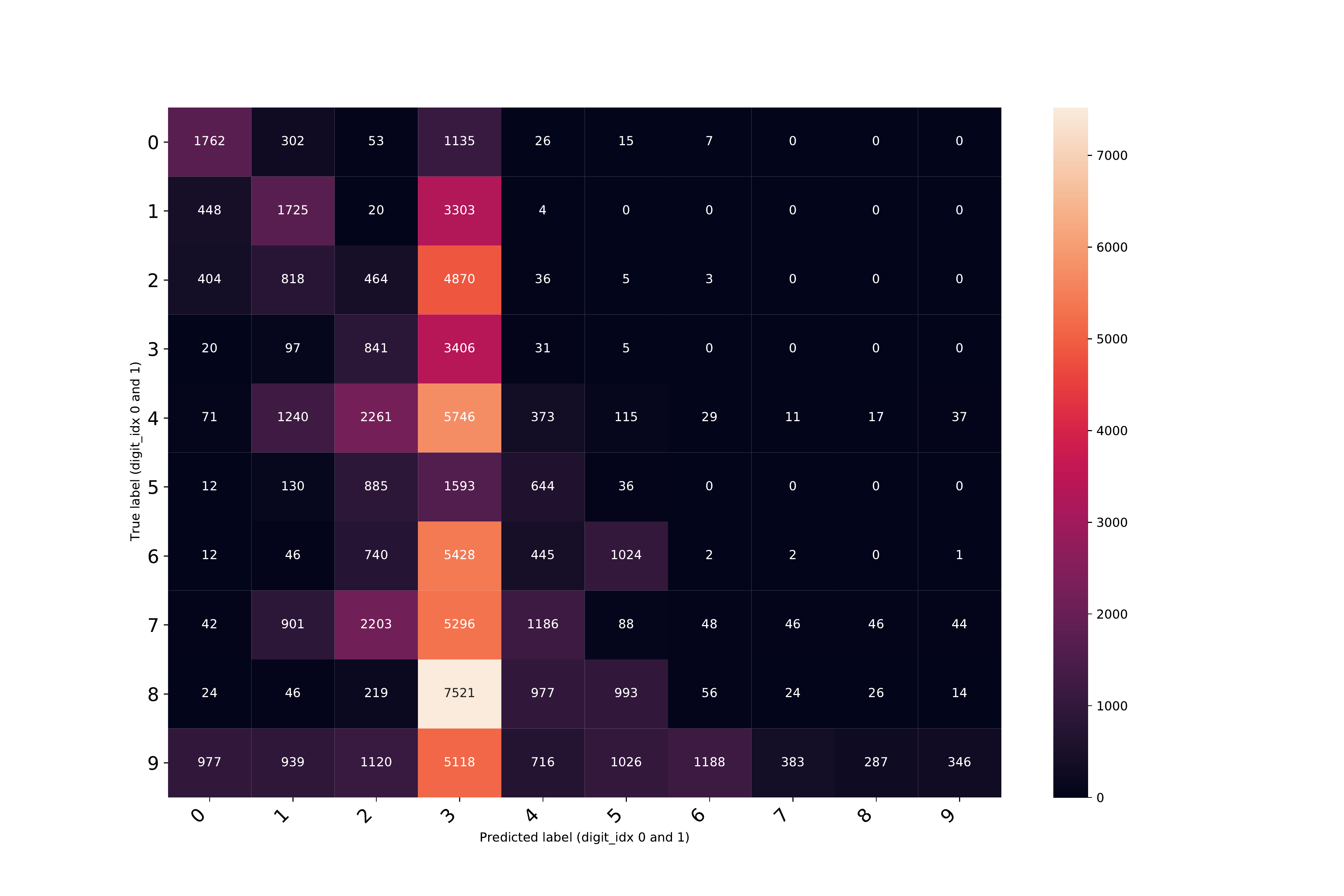}
  \caption{FC}
\end{subfigure}%\hfil
\begin{subfigure}{0.32\textwidth}
  \includegraphics[trim=0.5cm 1cm 7cm 1cm, width=\linewidth]{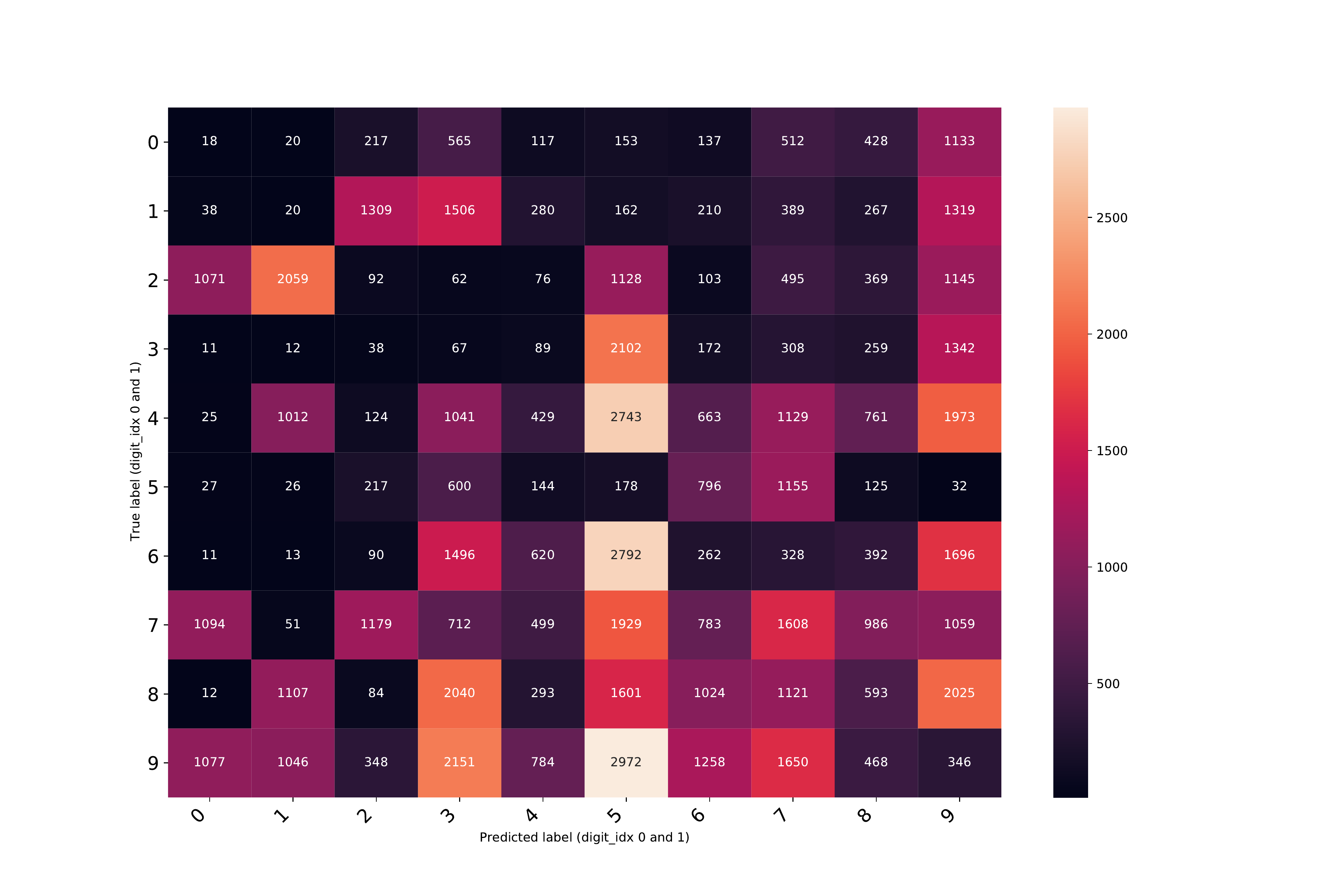}
  \caption{NMU}
\end{subfigure}

\medskip
\begin{subfigure}{0.32\textwidth}
  \includegraphics[trim=0.5cm 1cm 7cm 1cm, width=\linewidth]{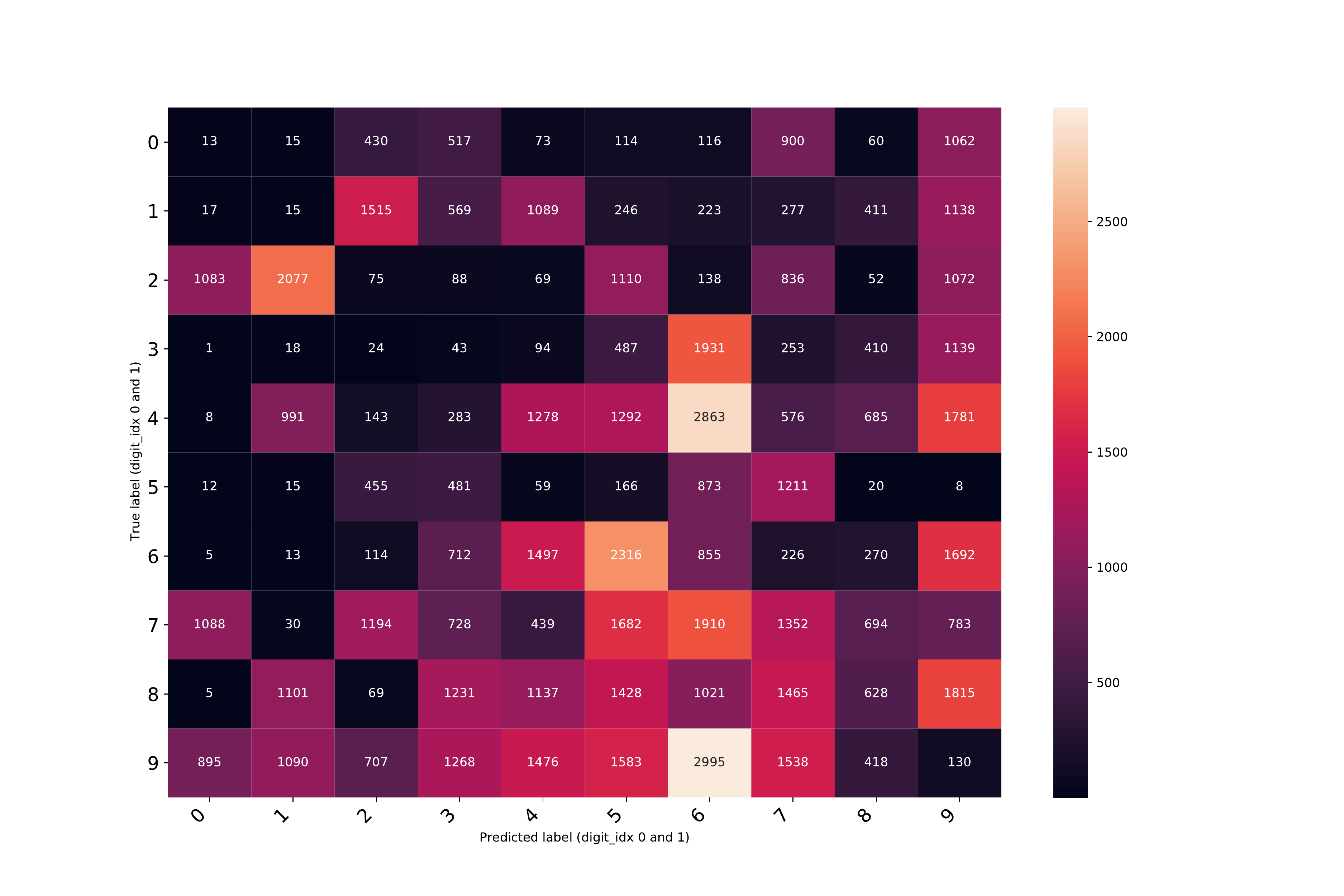}
  \caption{sNMU $\mathcal{U}$[1,5]}
\end{subfigure}%\hfil 
\begin{subfigure}{0.32\textwidth}
  \includegraphics[trim=0.5cm 1cm 7cm 1cm, width=\linewidth]{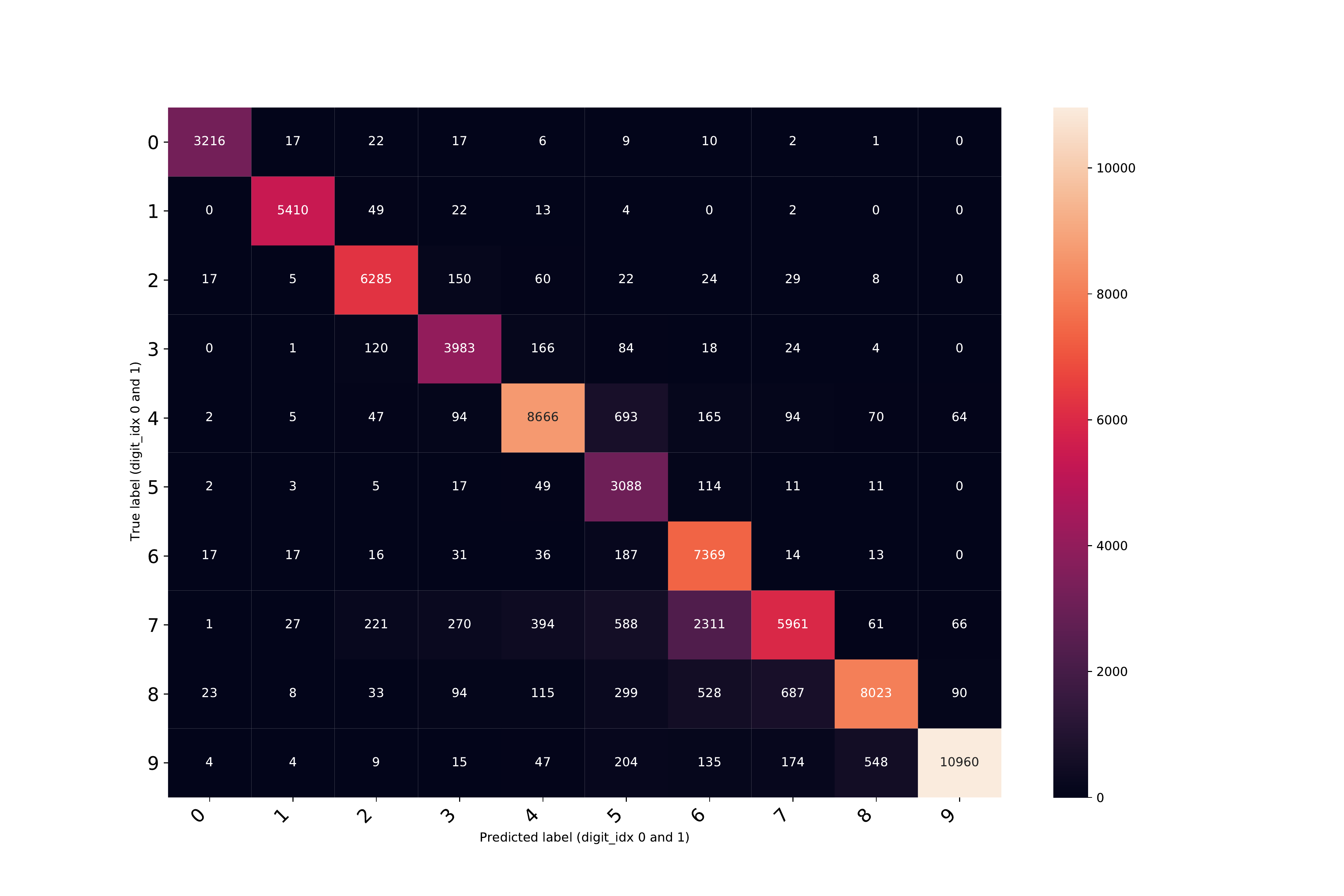}
  \caption{sNMU $\mathcal{U}$[1,1+1/sd(x)]}
\end{subfigure}%\hfil
\caption{Confusion matrices for intermediate label classification.}
\label{fig:mnist-st-tps-label-confusion-matricies}
\vskip -0.1in
\end{figure}

\textbf{Colour Channel Concatenated Digits.} %uses rounded accs
Figures~\ref{fig:mnist-st-tps-label-class-accs} and \ref{fig:mnist-st-tps-label-confusion-matricies}. 
Results are shown for a fold which models found especially challenging. 
Only the baseline and the sNMU using batch statistics are able to learn a classifiers which can provide a distinct diagonal over the confusion matrix. 
For this fold, the batch sNMU can outperform the baseline's classifier for every digit, implying the learnable multiplication layer provides a better optimisation landscape. 
The remaining multiplication models have no sign of convergence, with the FC model learning to have a high bias towards classifying the digit 3. 

\newpage
\subsection{Digit Classification Accuracy over Epochs}
This section shows how accuracies for classifying each digits evolves over the epochs. The average values over all folds are shown (with 95\% confidence intervals). 

\begin{figure*}[t!]
    \vskip 0.1in
    \centering
    \subfloat[\centering First digit]{{\includegraphics[ width=0.48\textwidth]{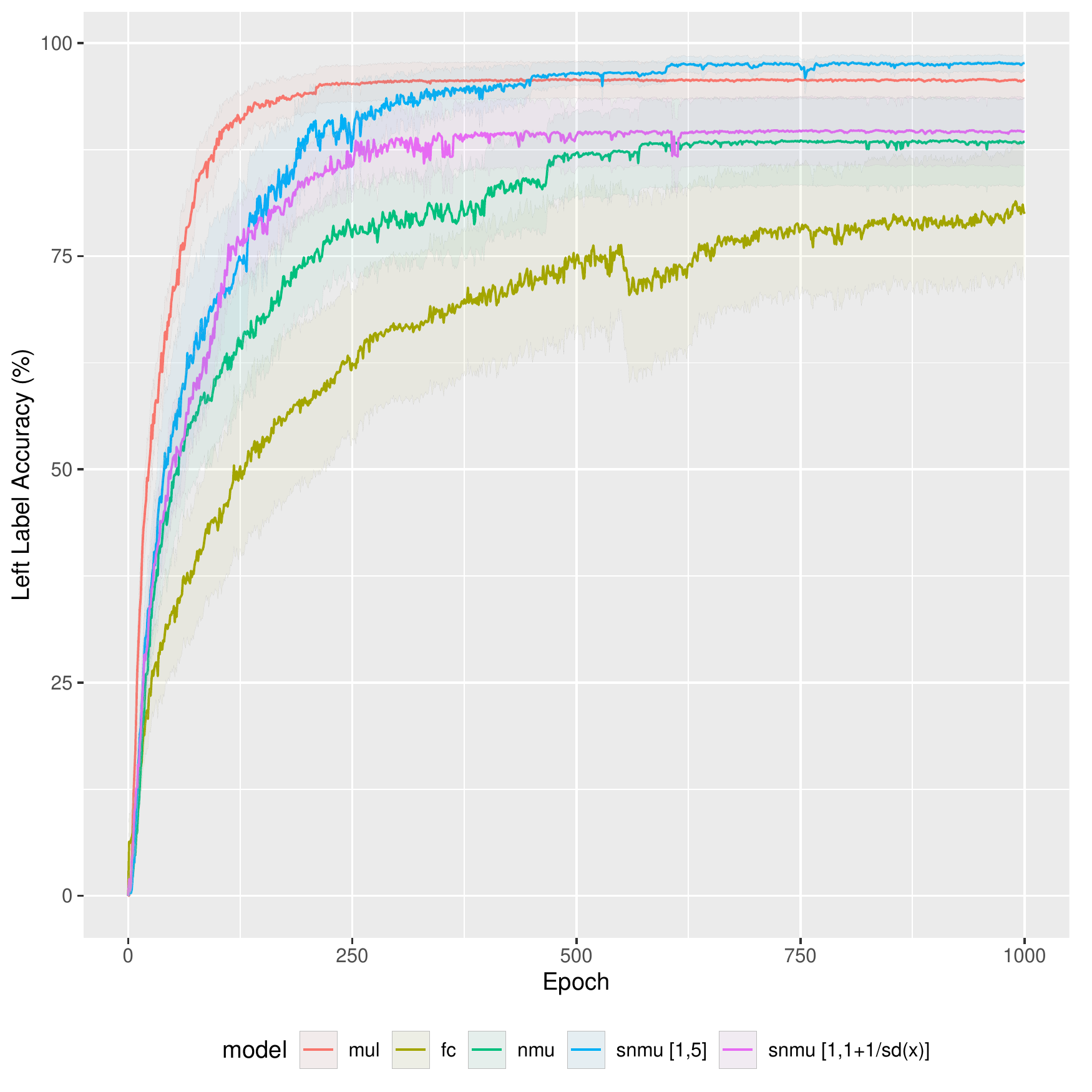}}}
    % \qquad
    \subfloat[\centering Second digit]{{\includegraphics[ width=0.48\textwidth]{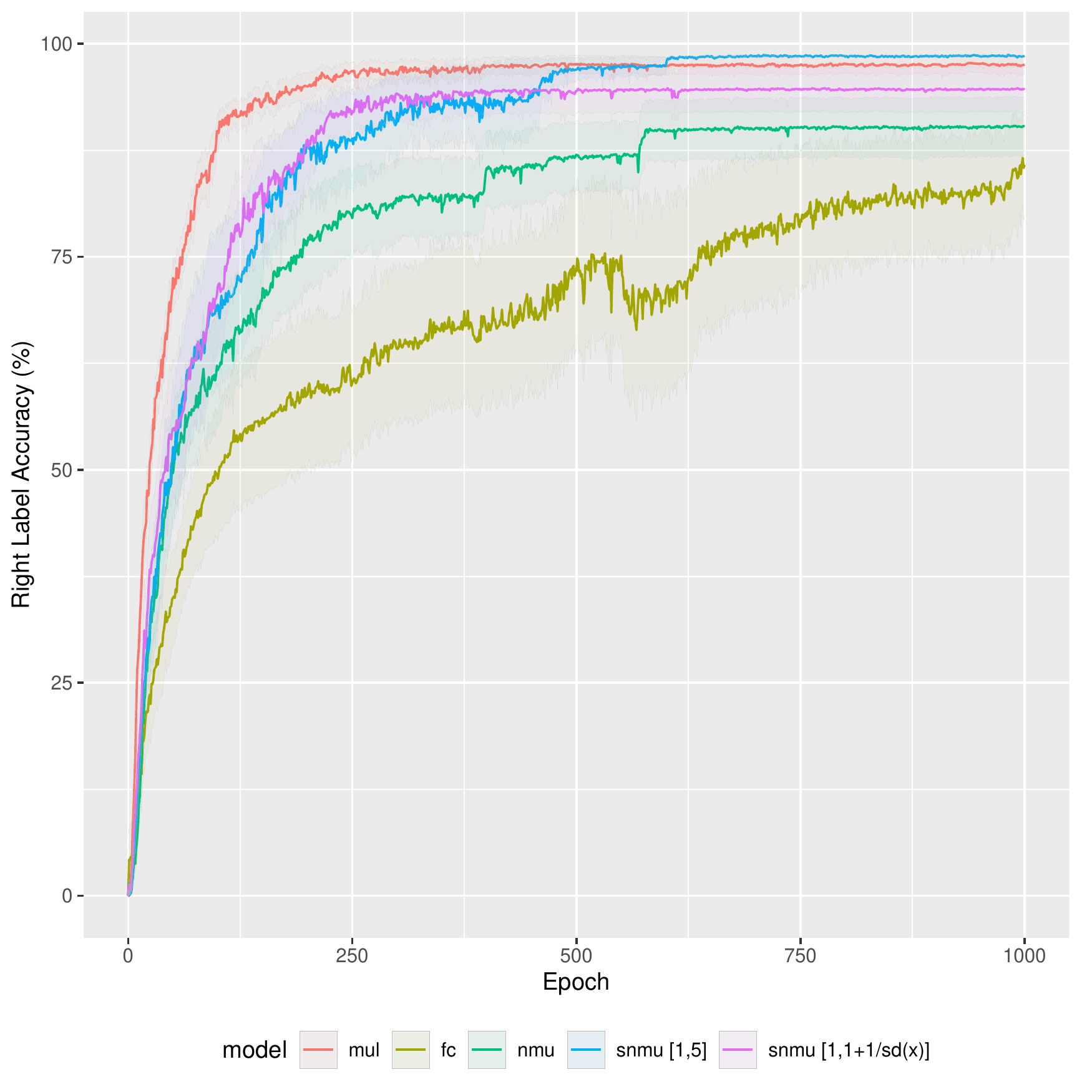}}}
    \caption{Label accuracy vs epoch of the two digits for the Isolated digit variant of the Static MNIST Product task.}
    \label{fig:1digit-mnist-labels-vs-epochs}
    \vskip -0.1in
\end{figure*}

\textbf{Isolated digits.}  
Figure~\ref{fig:1digit-mnist-labels-vs-epochs} shows similar leaning for both digits, with sNMU modules providing small confidence bounds. 
The sNMU with range $\mathcal{U}$[1,5] also shows better accuracy over the baseline after approximately 600 epochs. 

\begin{figure*}[th!]
    \vskip 0.1in
    \centering
    \subfloat[\centering First digit]{{\includegraphics[ width=0.47\textwidth]{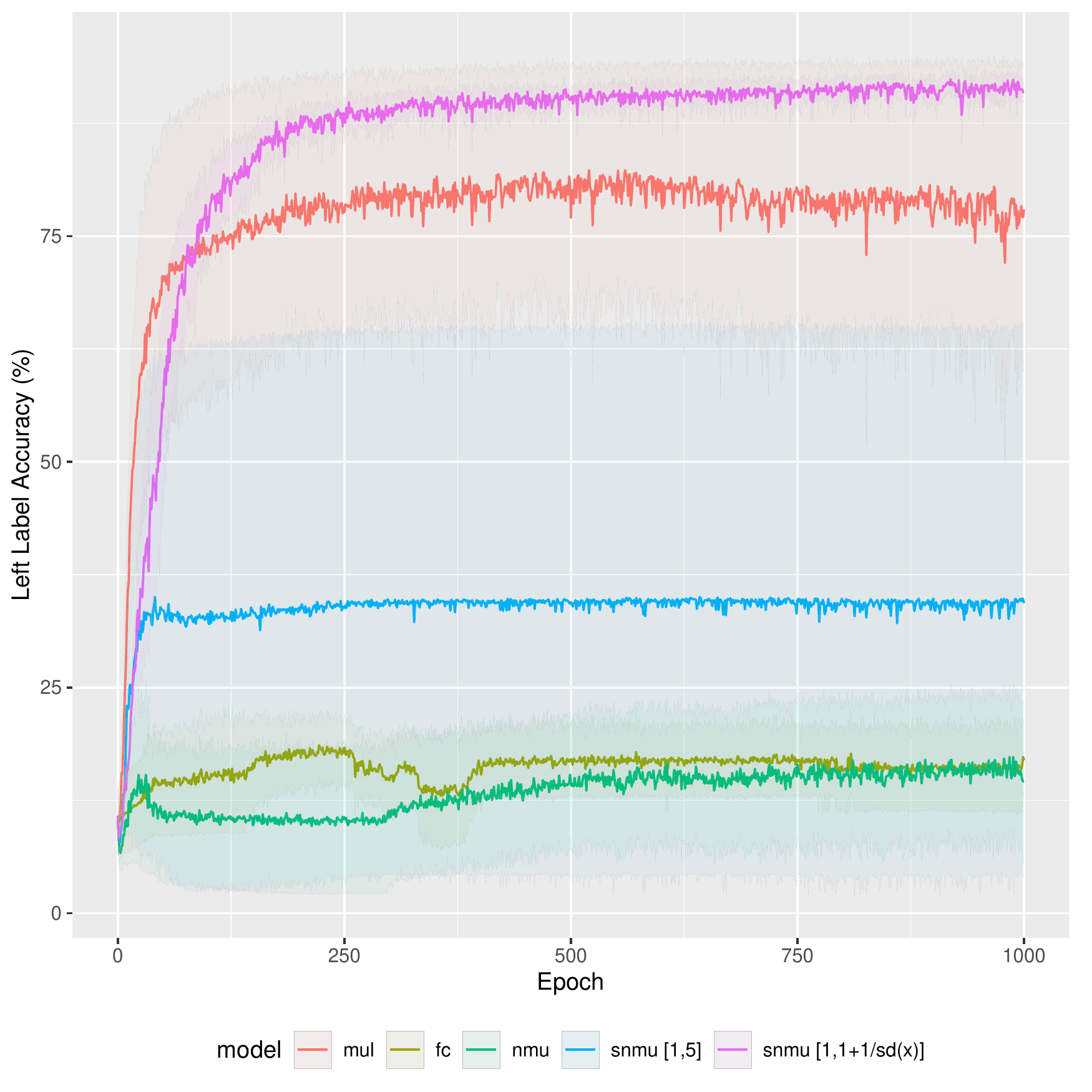}}}
    % \qquad
    \subfloat[\centering Second digit]{{\includegraphics[ width=0.47\textwidth]{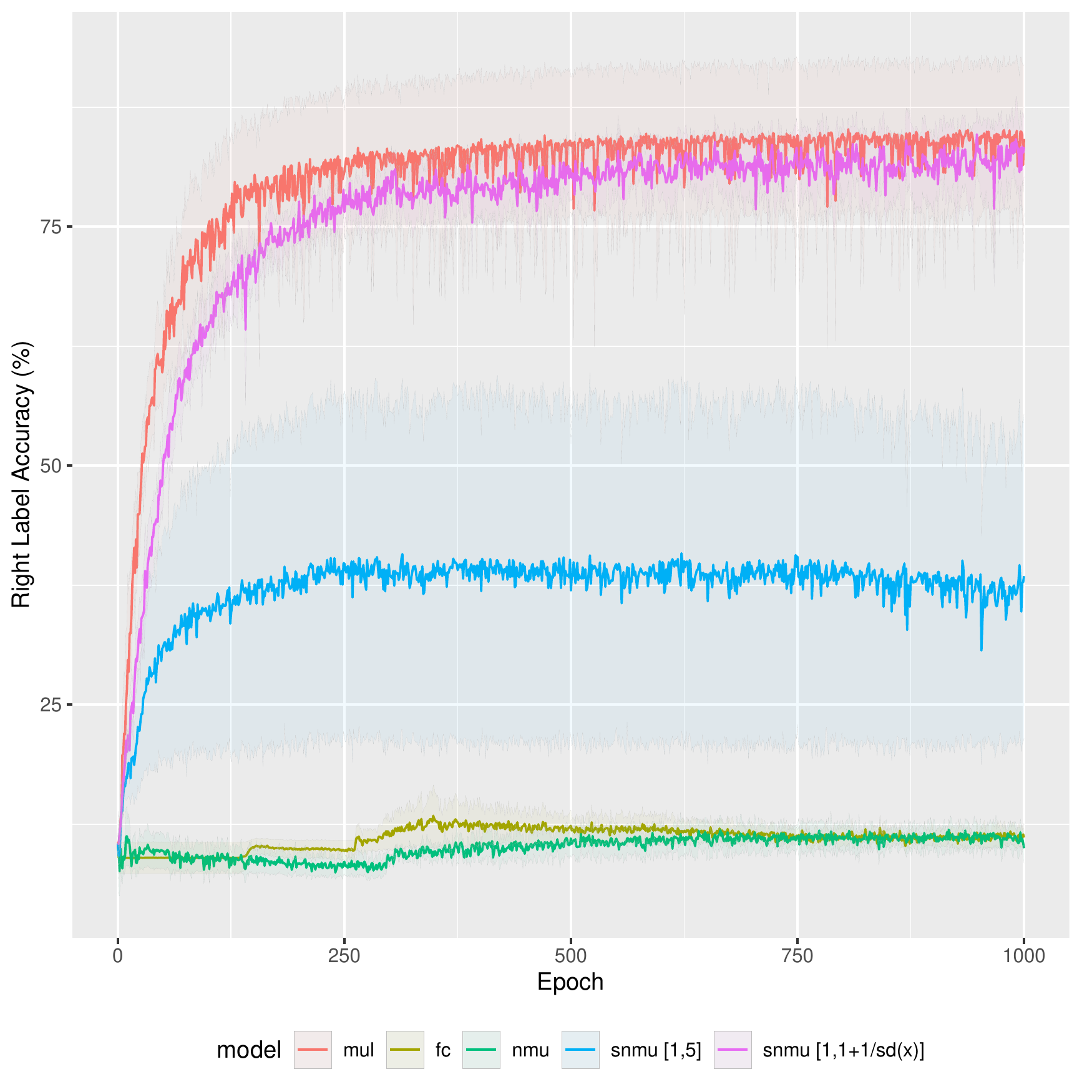}}}
    \caption{Label accuracy vs epoch of the two digits for the Colour channel concatenated digit variant of the Static MNIST Product task.}
    \label{fig:mnist-st-tps-labels-vs-epochs}
    \vskip -0.1in
\end{figure*}

\textbf{Colour Channel Concatenated Digits.} 
Figure~\ref{fig:mnist-st-tps-labels-vs-epochs} shows a greater variation of performance over the different models and the different folds in comparison the the isolated digits' results. 
The difference can be explained by the increased difficulty for this task, where localisation for each digit must be completed by the image classifier network. 
The solved baseline model shows challenges in robustness from the large confidence bounds while the batch sNMU provides much tighter bounds. 
The importance of having a reasonable noise range is also reflected in this task, with the sNMU using $\mathcal{U}$[1,5] noise unable to learn any reasonable image classifiers. 
It is also clear even with a bad noise interval, using stochasticity is better than not using stochasticity (i.e. sNMU vs NMU). 

\newpage
\section{Alternate Stochastic Methods}

We explore how other forms of stochasticity can effect learning of a NALM. 
The two methods are gradient noise and stochastic gating. 

\subsection{Gradient Noise}
Rather than implicitly altering the gradients by using reversible stochasticity, we see if the altering the gradients of a NALM explicitly can improve learning. 
Following ~\cite{neelakantan2015adding}, we add noise sampled from $N(0, \sigma^2_t)$ to the gradients every training step. 
The noise is annealed over epochs, therefore $\sigma^2_t=\frac{\eta}{(1+e)^{0.55}}$ where e is the epoch and $\eta$ is a scaling factor. 

\subsection{Stochastic Gating}
Rather than using stochasticity to modify the gradients of the weights, we reformulate the NMU weights such that each weight learns using directly injected noise. 
To do this, we represent a NMU weight as a learnable stochastic gate~\cite{yamada20_stg}. 
A gate represents a continuous relaxation of a Bernoulli distribution by modelling a mean shifted Gaussian random variable clipped between [0,1]. A learnable mean, $\mu_i$, is initialised to 0.5. 
The NMU weight is obtained by transforming the gate weight using a hard sigmoid. 
A NMU weight is defined as $w_i= \max(0, \min(1, \mu_i + \epsilon_i))$ where $\epsilon_i$ is noise sampled from $N(0, \sigma^2)$ and $\sigma=0.5$. 
During training noise is used but during inference no noise is added. 
L0 regularisation is applied by taking the probability that the gates are active, which can be calculated using a standard Gaussian CDF i.e., $\Sigma_{i=1}^I\Phi(\frac{\mu_i}{\sigma})$.   
To balance the regularisation with the main loss objective, the regularisation is scaled by a pre-defined lambda.

\subsection{Single Layer Task}
\begin{figure*}[h]
\centering
\includegraphics[height=5cm, width=0.9\textwidth]{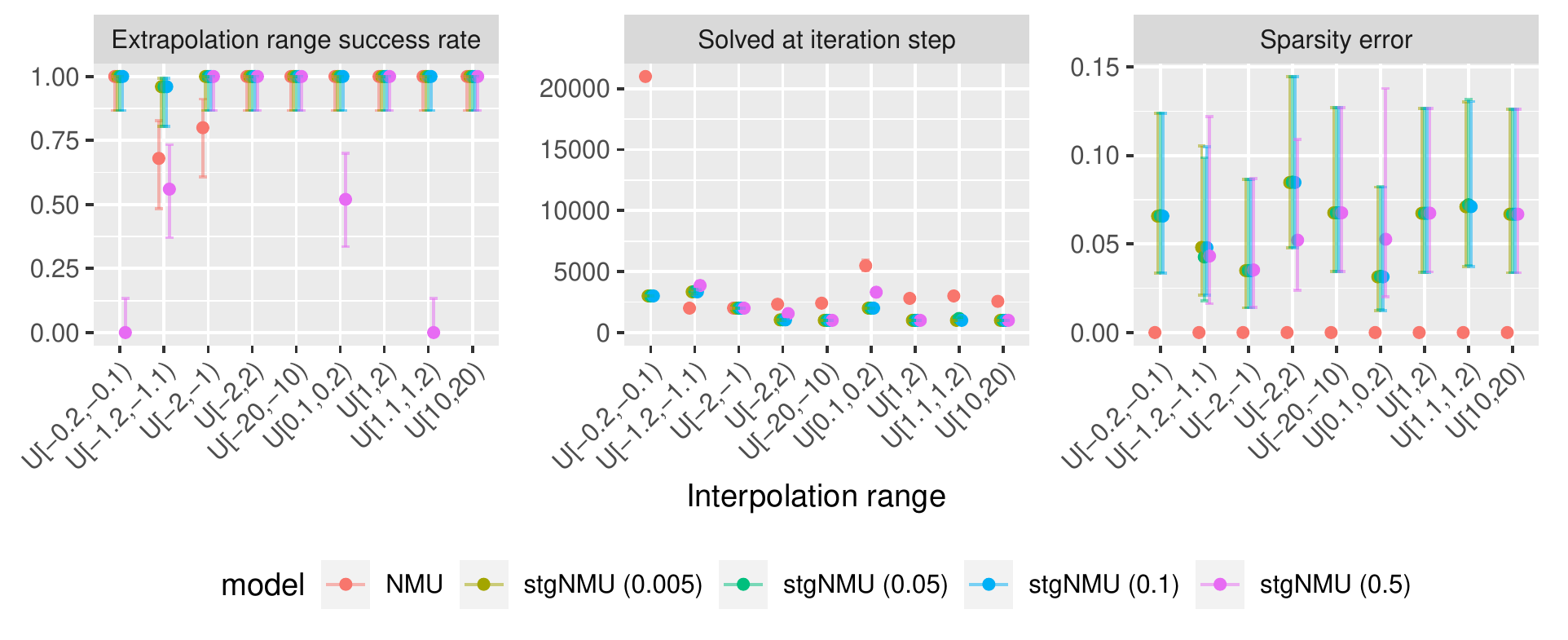}
\caption{Single Layer Task results for using stochastic gating to learn the NMU weights over different $\lambda$. }
\label{fig:in2-stg-sweep}
\end{figure*}

Figure~\ref{fig:in2-stg-sweep} shows that using stochastic gates for the NMU weighs can be beneficial if the lambda is sufficiently small (0.1 or under). When small lambdas are used, the stgNMU can improve on the remaining two ranges which the NMU struggles to obtain full success. 

\begin{figure*}[h]
\centering
\includegraphics[height=5cm, width=0.9\textwidth]{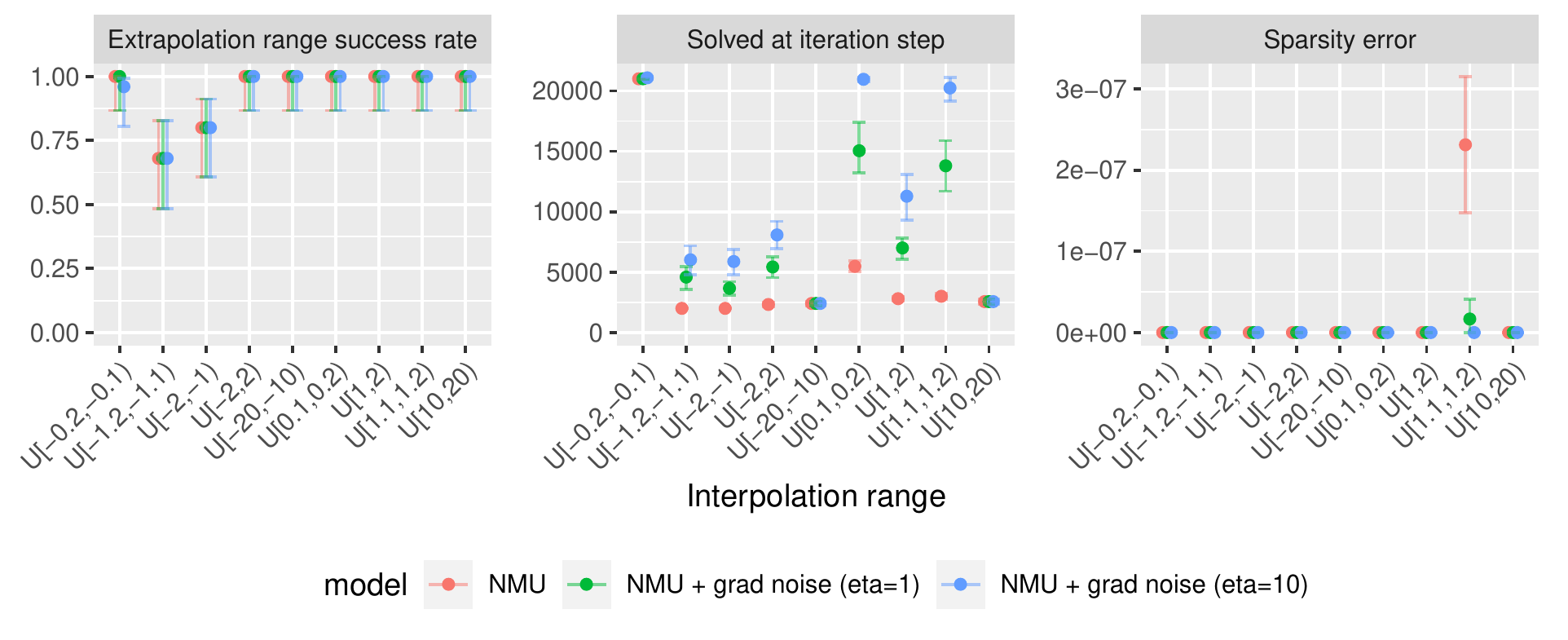}
\caption{Single Layer Task results for using stochastic gating to learn the NMU weights over different $\eta$.}
\label{fig:in2-gradNoise-sweep}
\end{figure*}

Figure~\ref{fig:in2-gradNoise-sweep} shows that adding gradient noise does not effect the success rate in comparison to the NMU. 
If larger gradient noise is used (i.e. eta=10) then the success can degrade (see U[-0.2, -0.1]). 
Furthermore, the convergence speeds get slower with gradient noise. 

\begin{figure*}[h]
\centering
\includegraphics[height=5cm, width=0.9\textwidth]{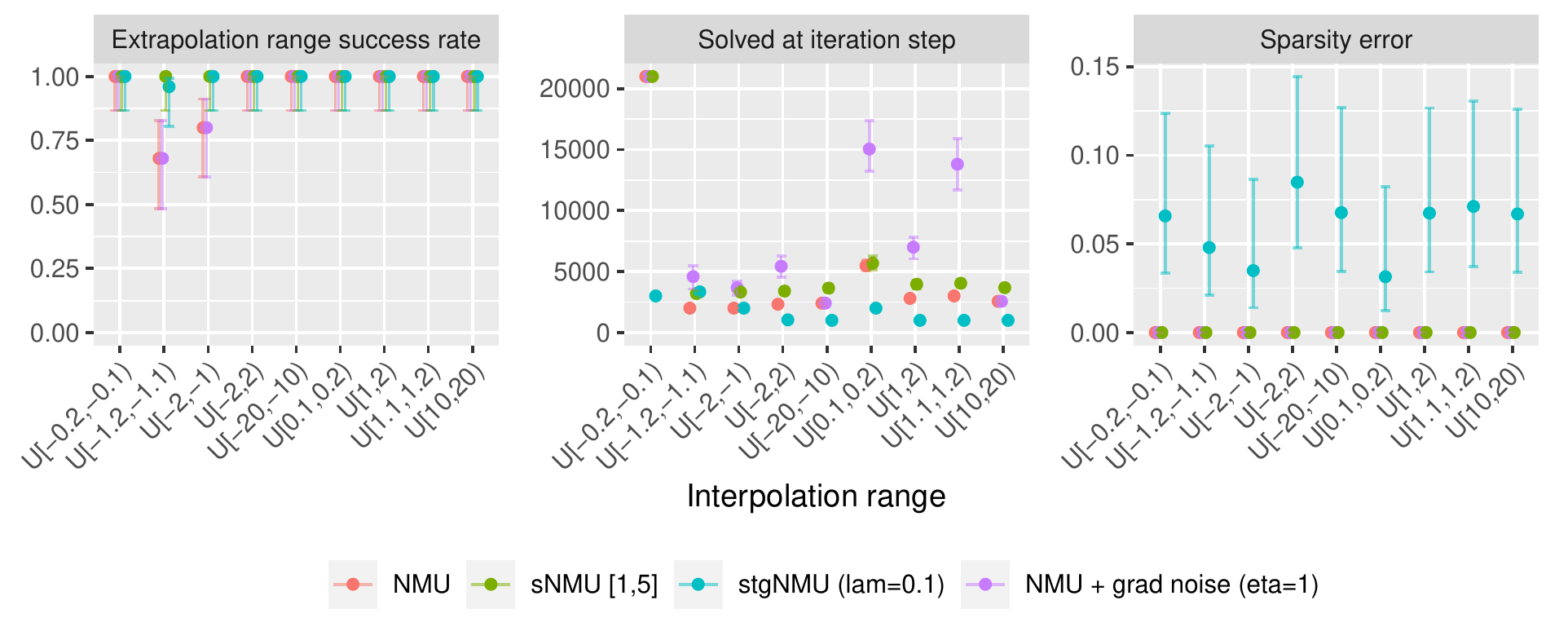}
\caption{Single Layer Task results comparing the NMU to NMUs trained with the following types of stochastic methods: reversible stochasticity (sNMU), stochastic gate weights (stgNMU) and gradient noise (NMU + grad noise).}
\label{fig:in2-noise-methods}
\end{figure*}

Figure~\ref{fig:in2-noise-methods} shows the three different noise methods compared against the baseline NMU. 

\subsection{Arithmetic Dataset Task}
\begin{figure*}[h]
\centering
\includegraphics[height=5cm, width=0.9\textwidth]{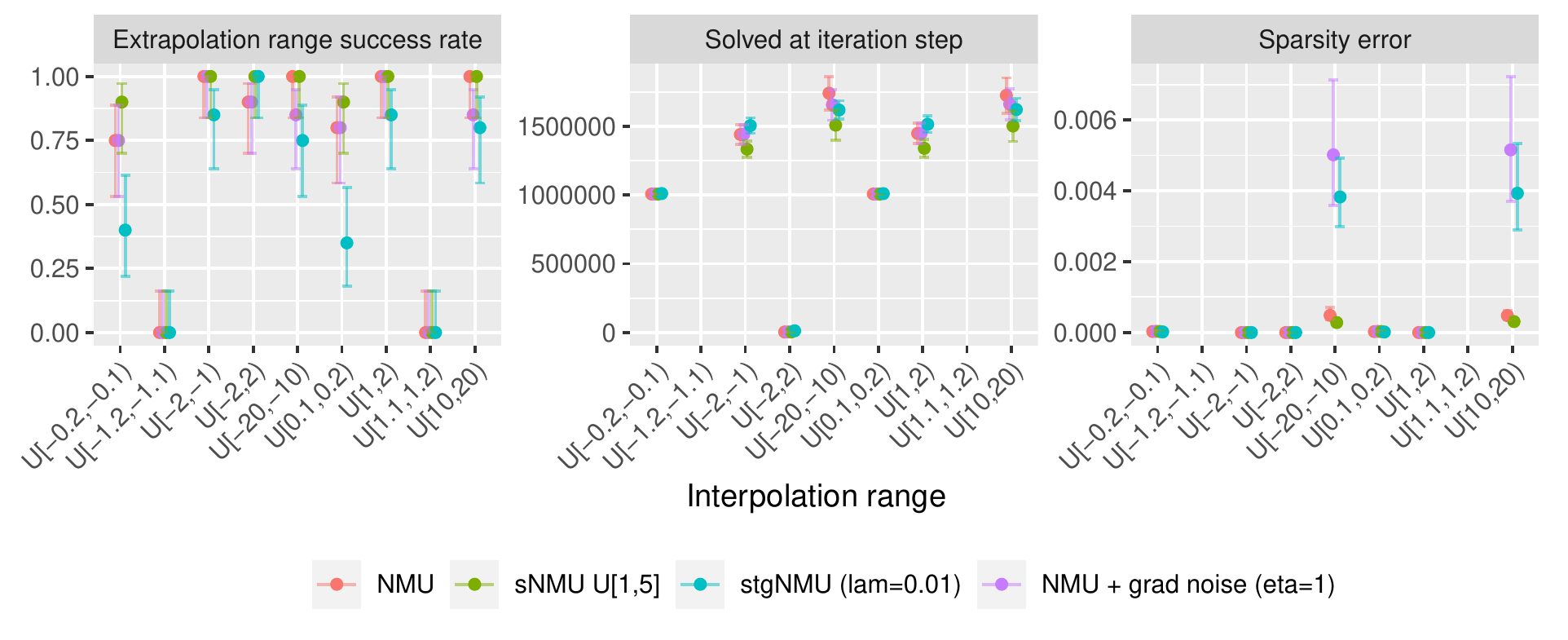}
\caption{Arithmetic Dataset Task comparing the NMU to NMUs trained with the following types of stochastic methods: reversible stochasticity (sNMU), stochastic gate weights (stgNMU) and gradient noise (NMU + grad noise). }
\label{fig:FTS-noise-methods}
\end{figure*}

Taking the best hyperparameters from the single layer task and training on the two layer Arithmetic dataset task results in Figure~\ref{fig:FTS-noise-methods} which shows that both the stgNMU and gradient noise hurt performance.  

% \bibliographystyle{plainnat}
% \bibliography{references}

\end{document}